\documentclass[preprint,authoryear,review,3p,12pt]{elsarticle}

\usepackage{amssymb}
\usepackage{amsmath}

\usepackage{lineno}

\journal{arXiv}

\usepackage{booktabs}
\usepackage{longtable}
\usepackage{url}
\usepackage[colorlinks]{hyperref}
\usepackage{multirow}
\usepackage{graphicx}
\usepackage{amsmath}
\usepackage{subcaption}
\usepackage[dvipsnames]{xcolor}
\usepackage{bbm}
\usepackage{bm}
\usepackage{coffeestains}

\usepackage{lscape}
\usepackage{array}
\usepackage{setspace}
\usepackage{adjustbox}

\newcommand{\botrule}{\bottomrule}

\newdefinition{definition}{Definition}

\DeclareMathOperator*{\argmin}{arg\,min}

\usepackage[most]{tcolorbox}

\newtcolorbox{findingbox}[1][]{
  enhanced,
  breakable,
  colback=white,      
  colframe=black!60,  
  boxrule=0.4pt,
  arc=1mm,
  fonttitle=\bfseries,
  title=#1
}

\begin{document}

\begin{frontmatter}



\title{\LARGE{{\it TabAttackBench}: A Benchmark for Adversarial Attacks on Tabular Data}} 


\author[inst1,inst2]{Zhipeng He\corref{cor1}}
\ead{zhipeng.he@hdr.qut.edu.au}
\author[inst1,inst2]{Chun Ouyang}
\ead{c.ouyang@qut.edu.au}
\author[inst3]{Lijie Wen}
\ead{wenlj@tsinghua.edu.cn}
\author[inst4]{Cong Liu}
\ead{cliu@novaims.unl.pt}
\author[inst5,inst2,inst6]{Catarina Moreira}
\ead{catarina.pintomoreira@uts.edu.au}

\affiliation[inst1]{organization={School of Information Systems, Queensland University of Technology},
            city={Brisbane},
            country={Australia}}

\affiliation[inst2]{organization={Center for Data Science, Queensland University of Technology},
            city={Brisbane},
            country={Australia}}

\affiliation[inst3]{organization={School of Software, Tsinghua University},
            city={Beijing},
            country={China}}

\affiliation[inst4]{organization={NOVA Information Management School, NOVA University of Lisbon},
            city={Lisboa},
            country={Portugal}}
            
\affiliation[inst5]{organization={Data Science Institute, University of Technology Sydney},
            city={Sydney},
            country={Australia}}

\affiliation[inst6]{organization={INESC-ID/Instituto Superior Técnico, University of Lisboa},
            city={Lisboa},
            country={Portugal}}

\cortext[cor1]{Corresponding author}

\begin{abstract}
Adversarial attacks pose a significant threat to machine learning models by inducing incorrect predictions through imperceptible perturbations to input data. While these attacks are well studied in unstructured domains such as images, their behaviour on tabular data remains underexplored due to mixed feature types and complex inter-feature dependencies. This study introduces a comprehensive benchmark that evaluates adversarial attacks on tabular datasets with respect to both effectiveness and imperceptibility. We assess five white-box attack algorithms (FGSM, BIM, PGD, DeepFool, and C\&W) across four representative models (LR, MLP, TabTransformer and FT-Transformer) using eleven datasets spanning finance, energy, and healthcare domains. The benchmark employs four quantitative imperceptibility metrics (proximity, sparsity, deviation, and sensitivity) to characterise perturbation realism. The analysis quantifies the trade-off between these two aspects and reveals consistent differences between attack types, with $\ell_\infty$-based attacks achieving higher success but lower subtlety, and $\ell_2$-based attacks offering more realistic perturbations. The benchmark findings offer actionable insights for designing more imperceptible adversarial attacks, advancing the understanding of adversarial vulnerability in tabular machine learning.
\end{abstract}




\end{frontmatter}





\section{Introduction}\label{sec:intro}

In recent years, the field of machine learning has seen substantial advancements, leading to the deployment of models across a wide range of applications. However, with these advancements comes increasing concern about the robustness and security of models, particularly in the context of adversarial attacks. Adversarial attacks involve the intentional manipulation of input data to deceive machine learning models, causing incorrect or misleading outputs \citep{szegedy2014intriguing}. This area of research has drawn significant attention as researchers strive to understand and mitigate the vulnerabilities in various types of data and models. 
Adversarial perturbations to images involve pixel intensity modifications \citep{weng2024comparative}, spatial transformations \citep{aydin2023adversarial}, texture perturbations \citep{geirhos2018imagenet}, and localised patches~\citep{wang2025unified} that cause dramatic misclassifications while remaining visually imperceptible in Computer Vision (CV). Similarly, in Natural Language Processing \citep{zhang2020adversarial}, attacks typically involve word substitutions \citep{yang2023quantifying}, character-level modifications \citep{rocamora2024revisiting}, or syntactic transformations \citep{asl2024semantic} that preserve semantic meaning while fooling text classifiers \citep{gao2024semantic}. Adversarial vulnerabilities have also been demonstrated in audio processing \citep{noureddine2023adversarial} through amplitude modifications \citep{ko2023multi}, frequency perturbations \citep{abdullah2019practical}, and psychoacoustic masking \citep{qin2019imperceptible} that cause speech recognition systems to misinterpret commands.
By addressing the vulnerabilities in these types of data, researchers aim to develop more robust and secure machine learning systems across various domains.

\subsection{Challenges in Adversarial Attacks on Tabular Data}

Tabular data, structured yet rich in semantics, heterogeneity, and interdependencies, is prevalent in domains such as finance, healthcare, and e-commerce \citep{borisov2022deep}. These datasets often contain vital information used for decision-making processes such as credit scoring \citep{mushava2024flexible}, medical diagnosis \citep{khosravi2023demystifying}, and fraud detection \citep{yi2023fraud}. Despite their significance, machine learning models trained on tabular data remain comparatively underexplored regarding their vulnerabilities to to adversarial manipulation. Even though small, carefully crafted perturbations can significantly alter predictions \citep{ballet2019imperceptible}, as illustrated in Table~\ref{tab:german_adv_example}.

Adversarial perturbations in tabular data arise through subtle yet realistic modifications of feature values --- such as adjusting a numerical attribute by a small increment, or flipping a categorical feature within its valid set --- to shift a model's decision boundary while maintaining a realistic record. For instance, in electronic health records, marginal changes to clinical indicators might lead to misclassification of disease risk \citep{an2019longitudinal}; in fraud detection, slightly altering transaction frequency or amount could conceal illicit behaviour \citep{lunghi2023adversarial}; and in loan approval systems, minor modifications to income or credit utilisation ratios could influence acceptance outcomes \citep{gunasekaran2023generating}.

These examples illustrate the broader issue: adversarial vulnerabilities in tabular data are not confined to a specific domain but are inherent to the data's structural and semantic complexity. Unlike image or text modalities, tabular datasets combine heterogeneous feature types (categorical and numerical) often with varying scales, missing values, and intricate interdependencies \citep{borisov2022deep}. These complexities make applying adversarial attacks to tabular data more intricate compared to image or text data.

\begin{table}[t]
\centering
\caption{Illustrative example of adversarial perturbations in the \textit{German Credit} dataset. 
Minor yet realistic feature changes can alter the model prediction while preserving record plausibility.}
\label{tab:german_adv_example}
\small
\begin{tabular}{lcc}
\toprule
\textbf{Feature} & \textbf{Original Record} & \textbf{Perturbed Record} \\
\midrule
Duration (months) & 24 & \textbf{31.3}~($\uparrow$) \\
Credit amount (DM) & 2,384 & \textbf{4,606}~($\uparrow$) \\
Age (years) & 64 & \textbf{58.6}~($\downarrow$) \\
Account check status & $<$ 0 DM & $<$ 0 DM \\
Credit history & existing credits paid back duly & same \\
Purpose & domestic appliances & same \\
Property & real estate & real estate \\
Housing & rent & rent \\
\midrule
\textbf{Model prediction} &
\textcolor{blue!70!black}{No Default (\checkmark)} &
\textcolor{red!70!black}{Default (\(\times\))} \\
\bottomrule
\end{tabular}
\end{table}

\subsection{Benchmarking Adversarial Attacks on Tabular Data}

To systematically study above challenges for tabular data, it is essential to establish \emph{benchmarks}—standardised frameworks that evaluate how different attack algorithms perform under controlled and comparable conditions on tabular data. While considerable progress has been made in understanding adversarial attacks on different modalities such as images, graphs, and time series (see Table~\ref{tab:benchmark}), there remains a relatively underexplored area: adversarial attacks on tabular data. Existing benchmarks have primarily focused on image data, employing metrics such as attack success rate, robust accuracy, and norm-based metrics (e.g., $\ell_\infty$, $\ell_2$) to quantify the strength of the adversarial perturbations. 
While these metrics are well-suited for image data, where imperceptibility is measured by slight pixel changes that remain ``indistinguishable to the human eye''~\citep{goodfellow2015explaining}, they do not translate directly to tabular data. 

\begin{table}[!htp]
\centering
\caption{Overview of existing benchmarks on adversarial attacks across different data types, attack scopes, and evaluation metrics. 
Our work uniquely focuses on benchmarking multiple \emph{attack algorithms} and their \emph{imperceptibility} on tabular data.}
\label{tab:benchmark}
\resizebox{\textwidth}{!}{%
\begin{tabular}{@{}lcll@{}}
\toprule%
\textbf{Benchmark}                       & \textbf{Data Type}                       & \textbf{Attack Type}                               & \textbf{Evaluation Metric}              \\
\midrule
\citet{jin2024benchmarking}      & Image                           & Transferable Attacks          & Attack Transferability Score   \\  
\citet{dong2020benchmarking}     & Image                           & White-box, Black-box Attacks              & $\ell_\infty$ Norm, Robust Accuracy \\
\citet{zheng2023blackboxbench}   & Image                           & Black-box Attacks             & Attack Success Rate, Query Count  \\ 
\citet{croce2020robustbench}     & Image                           & $\ell_\infty$, $\ell_2$ Norm-based Attacks & $\ell_\infty$, $\ell_2$ Norm, Robust Accuracy \\ 
\citet{hingun2023reap}           & Image                           & Patch-based Attacks           & Attack Success Rate, Realism Score  \\ 
\citet{cina2024attackbench}      & Image                           & Gradient-based Attacks                    & Attack Success Rate  \\ 
\citet{zheng2021graph}           & Graph                           & Graph Modification             & Robust Accuracy   \\ 
\citet{li2021adversarial}        & VQA & Visual–Textual Attacks                & Robust Accuracy  \\ 
\citet{siddiqui2020benchmarking} & Time-series                     & Temporal Attacks & Attack Success Rate  \\ 
\citet{simonetto2024tabularbench} & Tabular &  Single Tabular Attack & Robust Accuracy \\
\textit{Our paper} & \textit{Tabular} & \textit{Mutiple Tabular Attacks} & \textit{Attack Success Rate, Imperceptibility Metrics} \\ 
\botrule

\end{tabular}
}%
\end{table}


A recent effort on tabular data, \citet{simonetto2024tabularbench}, provides a valuable benchmark focusing on dataset and model-level evaluation under a single attack setup, emphasising data augmentation and robust accuracy. In contrast, our benchmark is designed to systematically evaluate and compare \emph{multiple attack algorithms} across various tabular datasets. 
Crucially, it integrates the concept of imperceptibility for tabular data by focusing on four key quantitative properties: \textbf{Proximity}, \textbf{Sparsity}, \textbf{Deviation}, and \textbf{Sensitivity} \citep{he2025investigating}. These properties ensure that adversarial examples closely resemble original data, minimise feature alterations, and respect the statistical distribution of the data.

By integrating measures of \emph{attack effectiveness} and \emph{imperceptibility}, we propose a benchmark that jointly evaluates how successful and how realistic adversarial attacks are on tabular data.

\subsection{Contribution}


\noindent This study seeks to address three open research questions:
\begin{enumerate}
    \item \textit{How effective are existing adversarial attack algorithms on tabular data?} 
    \item \textit{How imperceptible are existing adversarial attacks when evaluated against the imperceptibility properties?}  
    \item \textit{To what extent can existing adversarial attacks achieve effectiveness while remaining imperceptible?}  
\end{enumerate}

\noindent To answer these questions, this paper makes the following key contributions:

\begin{itemize}
    \item \textbf{Comprehensive Benchmark for Tabular Data.}  
    We introduce a benchmark that systematically evaluates representative \emph{white-box} adversarial attacks on diverse tabular datasets.  
    The benchmark jointly assesses \emph{attack effectiveness} (e.g., Attack Success Rate) and \emph{imperceptibility}, providing the first unified evaluation framework for structured data.

    \item \textbf{Imperceptibility Metrics Grounded in Prior Theory.}  
    Building upon our earlier formal framework~\citep{he2025investigating}, we operationalise four quantitative imperceptibility metrics, including \emph{Proximity}, \emph{Sparsity}, \emph{Deviation}, and \emph{Sensitivity}.  
    These metrics capture how closely adversarial examples align with original data distributions, extending the notion of ``indistinguishability'' beyond image norms to heterogeneous tabular features.

    \item \textbf{Empirical Analysis of Trade-offs.}  
    Through extensive experiments across multiple datasets and models, we reveal quantitative trade-offs between attack success and imperceptibility.  
    Our findings highlight which attack families tend to favour realism versus disruption, offering practical guidance for future algorithm design.
\end{itemize}

The remainder of this paper unfolds as follows. 
Section~\ref{sec:background} reviews existing research and taxonomies that define the scope of this benchmark. 
Section~\ref{sec:method} then introduces the proposed benchmarking methodology, detailing how both effectiveness and imperceptibility are systematically assessed across representative attacks and datasets. 
Section~\ref{sec:evaluation} reports the experimental results and analyses the quantitative trade-offs between attack success and imperceptibility. 
Finally, Section~\ref{sec:discussion} reflects on the key insights, practical implications, and future research directions arising from this work.


\section{Background and Related Work}
\label{sec:background}



The study of adversarial machine learning originates from the recognition that predictive models, despite demonstrating strong performance under standard evaluation, can be highly fragile when faced with carefully crafted perturbations. The seminal work by \citet{szegedy2014intriguing} first revealed that small, often imperceptible perturbations applied to image data could cause state-of-the-art classifiers to produce incorrect predictions. Subsequent work by \citet{goodfellow2015explaining} provided both a theoretical explanation and efficient algorithms for generating such adversarial examples, sparking a broad and active line of research. At its core, adversarial machine learning investigates the interaction between an adversary, who deliberately perturbs inputs within allowable modifications, and a predictive model, whose robustness is evaluated against these perturbations. 

Robustness analysis requires us to extend the standard learning framework by considering how models behave when their inputs are deliberately modified. Not all modifications are meaningful: some perturbations would lead to invalid or semantically inconsistent inputs. To address this, ML introduces the notion of \emph{perturbation sets}, which formalise the allowable neighbourhoods around a given instance. 

\begin{definition}[Perturbation Set]
For an instance $\bm{x} \in \mathcal{X}$, a \emph{perturbation set} $\Delta(\bm{x}) \subseteq \mathcal{X}$ 
is the collection of all instances $\bm{x}'$ that are considered valid modifications of $x$.
Formally, $\Delta(\bm{x})$ defines the allowable neighbourhood of $\bm{x}$ under which adversarial examples may be sought.
\end{definition}

In the context of \emph{tabular data}, the construction of $\Delta(\bm{x})$ must respect the heterogeneous nature of features. 
For continuous features, perturbations are typically restricted to bounded changes (e.g., $\|\bm{x}'-\bm{x}\|_\infty \leq \epsilon$) or minimal perturbation (e.g., $\|\bm{x}'-\bm{x}\|_2$), while for categorical features, perturbations are constrained to valid substitutions within the feature's domain. This ensures that all perturbed instances $\bm{x}' \in \Delta(\bm{x})$ remain semantically meaningful and valid.



An \emph{adversarial example} is an instance obtained by deliberately perturbing a valid input within its perturbation set, with the goal of challenging the model's prediction. 

\begin{definition}[Adversarial Example]
Let $f_\theta : \mathcal{X} \to \mathcal{Y}$ be an ML classifier, and $(\bm{x},y) \in \mathcal{X} \times \mathcal{Y}$ 
be a correctly classified instance, i.e., $f_\theta(\bm{x}) = y$. An \emph{adversarial example} for $\bm{x}$ is a perturbed instance $\tilde{\bm{x}} \in \Delta(\bm{x})$, often expressed as $\tilde{\bm{x}} = \bm{x} + \delta$ with $\delta \in \Delta(\bm{x})$, such that $f_\theta(\tilde{\bm{x}}) \neq y$.
If the prediction is instead forced into a specified target label $y_t \neq y$, 
then $\tilde{\bm{x}}$ is referred to as a \emph{targeted adversarial example}, 
satisfying $f_\theta(\tilde{\bm{x}}) = y_t$.  
\end{definition}

Formally, adversarial examples are generated by an \emph{adversarial attack}, which searches for $\tilde{\bm{x}} \in \Delta(\bm{x})$ that maximises some misclassification objective. Whether the resulting adversarial example actually causes a prediction change depends on the model under attack. 

\begin{definition}[Adversarial Attack]
Let $f_\theta : \mathcal{X} \to \mathcal{Y}$ be a classifier, 
$\ell : \mathcal{Y} \times \mathcal{Y} \to \mathbb{R}_{\geq 0}$ a loss function, 
and $\Delta(x) \subseteq \mathcal{X}$ a perturbation set. 
An \emph{adversarial attack} is an optimisation procedure that, given a correctly classified instance $(x,y)$, 
searches over candidates $\bm{x}' \in \Delta(\bm{x})$ and outputs an adversarial example $\tilde{\bm{x}} \in \Delta(\bm{x})$ that satisfies a misclassification objective. 
\end{definition}

\subsection{The Taxonomy of Adversarial Attacks}
\label{sec:taxonomy}

\begin{figure*}[htp!]
    \centering
    \includegraphics[width=\textwidth]{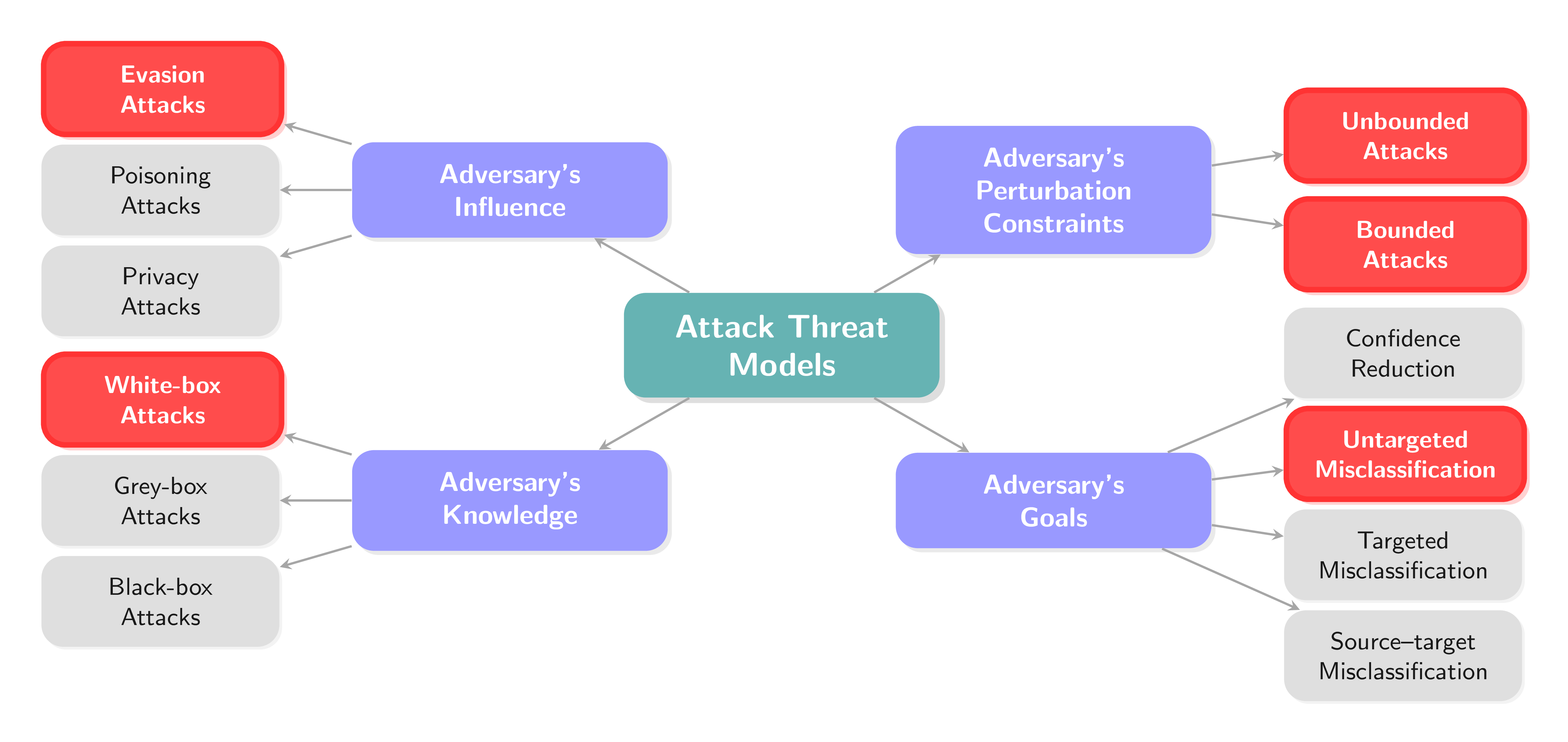}
    \caption{A taxonomy of adversarial attack threat models. The taxonomy spans four primary dimensions: \emph{Adversary's Influence}, \emph{Adversary's Knowledge}, \emph{Adversary's Perturbation Constraints} and \emph{Adversary's Goals}. Highlighted categories indicate the specific the research scope in this benchmark.}
    \label{fig:taxonomy}
\end{figure*}

Given the wide variety of adversarial attack strategies proposed in the literature, researchers have established several taxonomies to characterise and compare them~\citep{papernot2016limitation,biggio2018wild,assion2019attackgenerator,sadeghi2020system,pelekis2025adversarial}. These taxonomies are not mutually exclusive, but together they provide a systematic framework for understanding the assumptions, objectives, and mechanisms of different attacks. Taken collectively, these dimensions define the \emph{threat model} of an adversary, which specifies the conditions under which an attack is mounted. Four of the most widely adopted dimensions for classification in Figure \ref{fig:taxonomy}, including: (1) the attacking stage in the model lifecycle (\emph{Adversary's Influence}),  (2) the adversary's knowledge of the model (\emph{Adversary's Knowledge}), (3) the constraints imposed on perturbations (\emph{Adversary's Capabilities}), and (4) the goal of the attack in terms of prediction outcomes (\emph{Adversary's Goals}).

\subsubsection{Adversary's Influence}
A complementary dimension for classifying adversarial attacks concerns the \emph{influence} or attack surface available to the adversary. 
This axis describes when and where the adversary acts relative to the model life-cycle and data pipeline. 
Three primary influence categories are commonly recognised:
\begin{itemize}
  \item \textbf{Evasion (test-time) attacks~\citep{biggio2013evasion,biggio2018wild}:} the adversary perturbs inputs at inference time so that the deployed model produces incorrect outputs without altering the training data or model parameters.
  \item \textbf{Poisoning (training-time) attacks~\citep{biggio2018wild}:} the adversary injects or modifies examples in the training set (or manipulates the training process) to compromise model behaviour at deployment, e.g., by raising the error on a target subgroup or enabling backdoor triggers.
  \item \textbf{Privacy (information) attacks~\citep{shokri2017membership}:} the adversary seeks to extract sensitive information about training data, model internals, or users (e.g., via membership inference, model inversion or parameter extraction), thereby compromising confidentiality rather than directly causing misclassification.
\end{itemize}

In this benchmark, we concentrate on \emph{evasion attacks} at test time, as they directly instantiate the adversarial risk framework introduced earlier and provide the most immediate measure of the model robustness when deployed in real-world decision-making settings.

\subsubsection{Adversary's Knowledge} 
Attacks are commonly distinguished by the amount of information available to the adversary. They are categorised into three primary types of attacks based on the adversary's knowledge of the target system~\citep{papernot2016limitation}:
\begin{itemize}
    \item \textbf{White-box settings:} 
    the adversary has full access to the model architecture, parameters, and gradients, enabling highly effective gradient-based optimisation methods.

    \item \textbf{Grey-box settings:} 
    partial information may be available (e.g., the architecture is known but weights are hidden), requiring approximate or transfer-based approaches.

    \item \textbf{Black-box settings:} 
    the adversary has no access to the internal details of the model and can only query the model to observe outputs, motivating query-efficient or transferability-based attacks.

\end{itemize}

Among these settings, this paper focuses on the \emph{white-box} setting, as it represents the worst-case scenario for model vulnerability, offering a rigorous basis for analysing robustness before extending consideration to more restrictive adversarial assumptions.

\subsubsection{Adversary's Constraints}

A further dimension of adversarial taxonomies concerns the \emph{constraints} imposed on perturbations. These constraints determine the extent to which inputs may be altered while still being considered valid, thereby balancing the adversary's power against the requirement that perturbed instances remain realistic and meaningful. 
Two broad categories are typically recognised:  

\paragraph{Unbounded attacks~\citep{carlini2017towards,Moosavi2016deepfool}} the adversary faces no explicit restriction on the magnitude of perturbations. The objective is to find the smallest possible modification $\bm{\delta}$ that changes the classifier's prediction.

\begin{definition}[Unbounded Adversarial Attack]
Let $f_\theta : \mathcal{X} \to \mathcal{Y}$ be an ML classifier, 
and $(\bm{x},y) \in \mathcal{X} \times \mathcal{Y}$ a correctly classified instance. 
An \emph{unbounded adversarial attack} seeks the adversarial example $\tilde{\bm{x}} = \bm{x} + \delta$ 
with the smallest possible perturbation $\delta$ that changes the classifier's decision, i.e.,
\[
\min_{\delta} \ \lVert \delta \rVert 
\quad \text{subject to } f_\theta(\bm{x}+\delta) \neq y .
\]
\end{definition}

\paragraph{Bounded attacks~\citep{goodfellow2015explaining,madry2017towards}} the adversary is restricted by a perturbation budget, typically expressed as $\lVert \bm{\delta} \rVert \leq \epsilon$. Within this budget, the adversary seeks to maximise prediction error while keeping perturbed inputs close to the original.  

\begin{definition}[Bounded Adversarial Attack]
Let $f_\theta : \mathcal{X} \to \mathcal{Y}$ be an ML classifier, 
and $\Delta(\bm{x}) = \{\bm{x}' \in \mathcal{X} : \lVert \bm{x}'-\bm{x} \rVert_p \leq \epsilon\}$ a norm-bounded perturbation set. 
A \emph{bounded adversarial attack} restricts the adversary to perturbations within the budget $\epsilon$. 
Formally, it searches for an adversarial example $\tilde{\bm{x}} \in \Delta(\bm{x})$ such that:
\[
f_\theta(\tilde{\bm{x}}) \neq y ,
\]
with the perturbation bounded by $\lVert \tilde{\bm{x}} - \bm{x} \rVert_p \leq \epsilon$.

\end{definition}

\subsubsection{Adversary's Goals}

The final dimension of the taxonomy concerns the \emph{goals} of the adversary \citep{papernot2016limitation,assion2019attackgenerator}, which specify the intended effect of perturbations on model predictions. 
These goals determine the success criterion of an attack and influence both the choice of optimisation objective and the evaluation metrics. 
Four canonical categories are widely recognised:  

\begin{itemize}
    \item \textbf{Confidence reduction:} 
    the adversary reduces the model's prediction confidence without necessarily changing the predicted class.  

    \item \textbf{Untargeted misclassification:} 
    the adversary seeks to induce any incorrect prediction, without control over which incorrect class is produced.  

    \item \textbf{Targeted misclassification:} 
    the adversary forces the model to predict a specific incorrect class.  

    \item \textbf{Source--target misclassification:} 
    the adversary specifies both the source class and the desired incorrect target class.  

\end{itemize}

Among these objectives, this benchmark concentrates on \emph{untargeted misclassification}. This choice reflects the practical relevance of arbitrary misclassification in tabular decision systems, where any incorrect prediction can undermine reliability, and the computational efficiency of untargeted attacks, which avoid the additional complexity of enforcing specific outcomes.

Taken together, these four dimensions provide a systematic framework for characterising adversarial attacks in Figure~\ref{fig:taxonomy}. 
In this benchmark, we adopt the \emph{evasion} setting under a \emph{white-box} threat model, with perturbations considered in both \emph{bounded} and \emph{unbounded} forms to capture practical as well as theoretical perspectives, and adversarial goals defined in terms of \emph{untargeted misclassification}. This combination represents the most rigorous and practically relevant formulation of adversarial robustness in tabular data.

\subsection{White-box Adversarial Attacks}


Most adversarial attack algorithms were originally developed and studied in the context of Computer Vision (CV), where inputs are represented as high-dimensional images and perturbations can be visualised directly. 
This domain has served as the primary testbed for formulating adversarial objectives, designing optimisation procedures, and benchmarking robustness. 
Although the focus of this benchmark is on tabular data, understanding these canonical attack methods is essential, as many tabular attack strategies are adapted or inspired from techniques first introduced 
in the CV literature.

\subsubsection{Fast Gradient Sign Method}

The fast gradient sign method (FGSM) attack \citep{goodfellow2015explaining} works by calculating the gradient of the neural network with respect to the input data and using the sign of this gradient to determine the direction of the perturbation. 

\begin{definition}[FGSM]
Given a classifier $f_\theta : \mathcal{X} \to \mathcal{Y}$ with loss function $\ell$, the FGSM adversarial example for input $\bm{x}$ with label $y$ is
\[
\tilde{\bm{x}} = \bm{x} + \epsilon \,\mathrm{sign}\!\big(\nabla_{\bm{x}} \ell(f_\theta(\bm{x}), y)\big),
\]
where $\epsilon > 0$ controls the perturbation magnitude and $\mathrm{sign}(\cdot)$ is applied elementwise.
\end{definition}

It is equivalent to taking a single $\ell_\infty$-bounded step of size $\epsilon$ in the steepest ascent direction. Its single-step nature makes it weaker than iterative attacks and prone to failure against robust models. Moreover, the uniform $\ell_\infty$ perturbation assumption does not align well with the heterogeneous constraints of tabular data.

\subsubsection{Iterative Method} 


The \textit{Basic Iterative Method (BIM)}~\citep{Kurakin2017adversarial} extends FGSM by applying it multiple times with a smaller step size, iteratively updating the adversarial example. Projected Gradient Descent~\citep{madry2017towards} generalises this idea by introducing random starts and projecting perturbed inputs back into the allowed perturbation set at each iteration, making it one of the strongest first-order adversaries.

\begin{definition}[BIM and PGD]
Given a classifier $f_\theta : \mathcal{X} \to \mathcal{Y}$, loss function $\ell$, and perturbation budget $\epsilon$, iterative adversarial examples are generated as
\[
\bm{x}^{(t+1)} = \mathrm{Proj}_{\bm{x},\epsilon}\!\left(\bm{x}^{(t)} + \alpha \,\mathrm{sign}\!\big(\nabla_{\bm{x}} \ell(f_\theta(\bm{x}^{(t)}), y)\big)\right),
\]
where $\alpha$ is the step size and $\mathrm{Proj}_{\bm{x},\epsilon}(\cdot)$ ensures that $\bm{x}^{(t)}$ remains within the $\ell_\infty$ ball of radius $\epsilon$ around $\bm{x}$. The final adversarial example is denoted $\tilde{\bm{x}} = \bm{x}^{(T)}$.

\begin{itemize}
  \item \textbf{BIM:} uses deterministic initialisation $\bm{x}^{(0)} = \bm{x}$, with projection implemented as clipping.  
  \item \textbf{PGD:} generalises BIM by starting from a random perturbation $\bm{x}^{(0)} \in B_\infty(\bm{x}, \epsilon)$ and projecting back at every step.  
\end{itemize}
\end{definition}

\subsubsection{Carlini and Wagner Attack} 

The Carlini and Wagner attack  \citep{carlini2017towards}, also known as the C\&W attack, is a family of optimisation-based methods that aims to find the minimum perturbation that can cause a misclassification in a targeted deep neural network. Unlike gradient-sign methods, the C\&W attack formulates the problem as a constrained optimisation under different $\ell_p$ norms, yielding adversarial examples with minimal perturbation.

\begin{definition}[C\&W Attack]
Given an input $\bm{x}$ with label $y$ and classifier $f_\theta$, the C\&W attack seeks
\[
\min_{\bm{\delta}} \ \lVert \bm{\delta} \rVert_p + c \cdot g(\bm{x}+\bm{\delta}),
\]
where $\bm{\delta}$ is the perturbation, $c > 0$ balances perturbation and misclassification, and $g(\cdot)$ is a surrogate loss term that enforces adversarial misclassification. A common choice is the logit-based hinge:
\[
g(\bm{x}') = \max\!\big(f_\theta(\bm{x}')_y - \max_{i \neq y} f_\theta(\bm{x}')_i, \, -\kappa\big),
\]
where $f_\theta(\bm{x}')_i$ is the logit score for class $i$ and $\kappa \geq 0$ is a confidence parameter controlling attack strength. The adversarial example is $\tilde{\bm{x}} = \bm{x} + \bm{\delta}$.
\end{definition}

\subsubsection{DeepFool}

DeepFool~\citep{Moosavi2016deepfool} formulates adversarial perturbations as the minimal change required to cross the classifier's decision boundary. Unlike FGSM or PGD, which use fixed-magnitude gradient steps, DeepFool approximates the boundary locally as linear and iteratively projects the input toward the nearest boundary until misclassification occurs. 

\begin{definition}[DeepFool]
Formulated under the $\ell_2$ norm, DeepFool assumes a locally linear decision boundary $\mathcal{F} = \{\bm{x} : \bm{w}^\top \bm{x} + b = 0\}$ for some weight vector $\bm{w}$ and bias $b$. 
It computes the minimal perturbation
\[
\bm{r}_*(\bm{x}) = \argmin_{\bm{\delta}} \ \lVert \bm{\delta} \rVert_2 
\quad \text{s.t.} \quad f(\bm{x}+\bm{\delta}) \neq f(\bm{x}),
\]
which corresponds to the orthogonal projection of $\bm{x}$ onto the separating hyperplane $\mathcal{F}$. 
The adversarial example is then updated iteratively as
\[
\bm{x}^{(t+1)} = \bm{x}^{(t)} + \bm{\delta}^{(t)},
\]
where $\bm{\delta}^{(t)}$ is the projection step, until misclassification is achieved. The final adversarial example is denoted $\tilde{\bm{x}} = \bm{x}^{(T)}$. 

In the linear binary case, this reduces to the closed-form solution
\[
r_*(\bm{x}) = - \frac{f(\bm{x})}{\lVert \bm{w} \rVert_2^2} \bm{w}, 
\quad \bm{x}^{adv} = \bm{x} + r_*(\bm{x}).
\]
\end{definition}





\subsection{Adversarial Attacks on Tabular Data}

Adversarial attacks on tabular data aim to manipulate inputs such that ML models produce incorrect predictions while keeping modifications imperceptible \citep{he2025investigating}. Within this context, adversarial machine learning on tabular data faces two central challenges, including \emph{effectiveness}, i.e., the ability to reliably degrade model performance through computationally efficient attacks; and \emph{imperceptibility}, i.e., ensuring that adversarial examples respect domain constraints, preserve feature interdependencies, and avoid unrealistic or easily detectable modifications. 

\paragraph{White-box Gradient-based Effective Attacks}
A canonical baseline is the PGD attack~\citep{madry2017towards}, which iteratively updates adversarial examples by ascending the gradient of the loss while constraining perturbations within a bounded $\ell_p$ ball. In the context of tabular data, PGD has been widely adopted as a benchmark due to its simplicity and strong empirical performance. 
Building on this foundation, several variants introduce domain-specific constraints to account for the heterogeneous and semantically structured nature of tabular features. Early work by \citet{ballet2019imperceptible} demonstrate the adaptation of canonical methods such as FGSM and PGD to tabular data, showing that even simple gradient-based attacks can mislead models when applied to structured inputs. More recently, Constrained Adaptive PGD (CAPGD) \citep{simonetto2021unified,simonetto2024constrained,simonetto2024tabularbench} refines this approach by adapting step sizes and directions to respect feasibility conditions such as valid categorical values and bounded numerical ranges. These methods illustrate how canonical gradient-based attacks can be adapted for the particular challenges of tabular domains.

\paragraph{Imperceptibility and Constraint-Aware Attacks}

A key challenge in tabular attack is imperceptibility: adversarial examples should not only cause misclassification but also remain realistic and semantically valid. LowProFool~\citep{ballet2019imperceptible} introduces feature-importance weighting, biasing perturbations toward low-salience features to improve imperceptibility. Similarly, The Feature Importance Guided Attack (FIGA) \citep{gressel2021feature} perturbs the most influential features identified by importance scores, maximising impact with minimal changes.
Subsequent research refined constraint handling: \citet{Mathov2022not} enforce consistency across feature interdependencies, while \citet{Chernikova2022FENCE} ensures feasibility in security-critical datasets. Cost-aware frameworks~\citep{kireev2022adversarial} further restrict perturbations according to domain-specific budgetary limits, such as financial cost in fraud detection.
More recently, CAPGD \citep{simonetto2024constrained} unified gradient-based and search-based optimisation to efficiently generate valid adversarial examples, accompanied by a benchmarking framework across diverse tabular datasets and models~\citep{simonetto2024tabularbench}.

\section{Methodology}
\label{sec:method}

\begin{figure*}[!htp]
    \centering
    \includegraphics[width=\linewidth]{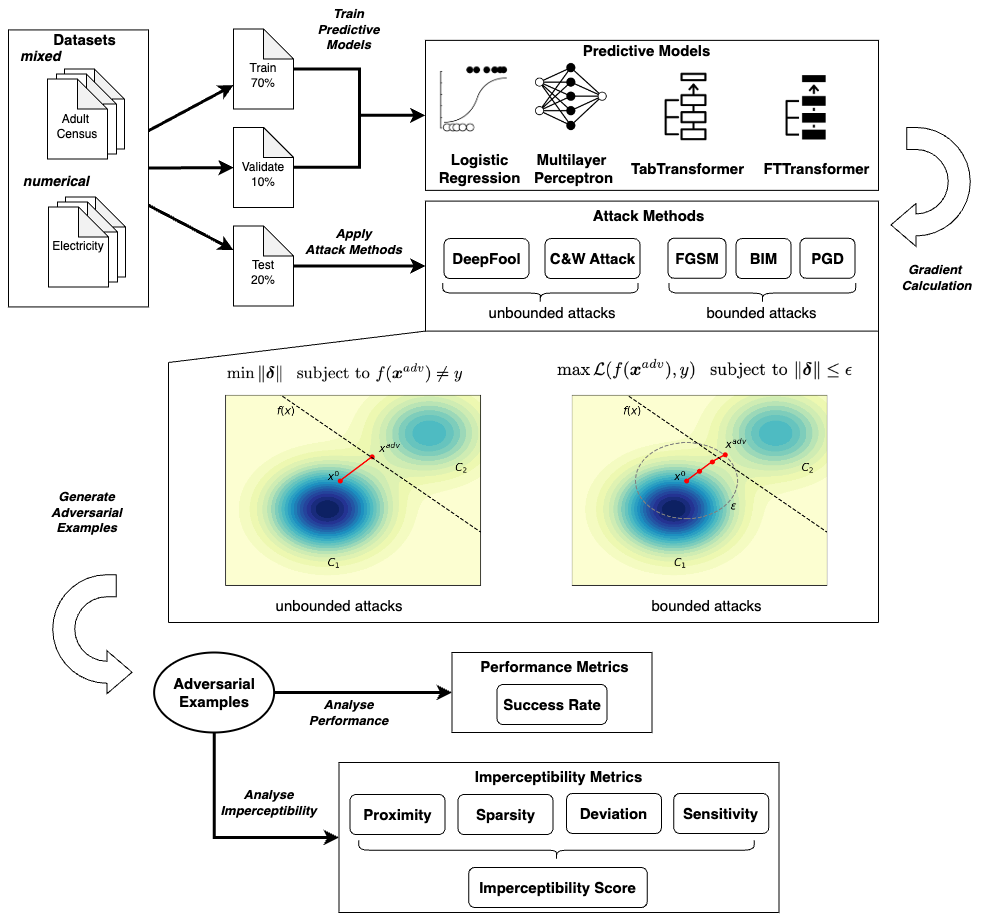}
    \caption{Roadmap of the proposed evaluation framework for benchmarking adversarial attacks on tabular data. The pipeline extends standard machine learning workflows by adding an adversarial attack stage, linking dataset preparation, model training, attack generation, and evaluation to provide a comprehensive measure of effectiveness and imperceptibility.}
    \label{fig:roadmap}
\end{figure*}

Integrating adversarial attacks as an additional phase in the standard machine learning pipeline enables a systematic assessment of model robustness against adversarial perturbations. Inspired by the established machine learning benchmark guidelines~\citep{thiyagalingam2022scientific}, we design our benchmark to evaluate adversarial attacks on tabular data. It comprises three sequential stages: (i) data preparation and model training, where clean tabular datasets are pre-processed and predictive models are optimised to establish baseline performance; (ii) adversarial example generation, in which attack algorithms craft perturbed inputs under a white-box assumption with full access to model parameters and gradients; and (iii) robustness evaluation, where the perturbed samples are assessed using both effectiveness (attack success rate) and imperceptibility metrics. This design enables consistent, reproducible comparison across models and attacks while isolating the influence of perturbations from data preprocessing or training variability. Figure \ref{fig:roadmap} summarises the overall benchmark pipeline and its component interactions.

\subsection{Datasets}\label{subsec:datasets}


The selection of datasets for the benchmark follows several criteria to ensure both suitability and effectiveness. 
\begin{enumerate}
    \item Each dataset must support classification tasks and contain at least two classes, enabling the evaluation of binary classification performance. 
    \item To maintain computational feasibility and avoid excessive dimensionality, the total number of features, including those generated by one-hot encoding of categorical attributes, is restricted to no more than~$200$.
    \item Datasets should be representative of real-world scenarios, with sufficient diversity of instances to meaningfully assess model robustness and generalisation.  
\end{enumerate}

In addition to our selection criteria, we also refer to datasets that are commonly adopted in the deep learning benchmark for tabular data \citep{gorishniy2021revisiting}, ensuring consistency with prior work and comparability across studies. Table~\ref{tab:data-profile} summarises the profiles of the $11$ datasets\footnote{WineQuality (White) and WineQuality (Red) originate from the same dataset but are treated as two separate datasets in this benchmark.} used, including their size, feature composition, and data splits for training, validation, and adversarial evaluation.



\begin{table*}[htb!]
\centering
\caption{Data profiles of the 11 datasets used in the benchmark, including the total number of instances ($n_{total}$), instances for training ($n_{train}$), validation ($n_{validate}$), and adversarial perturbation ($n_{test}$), as well as the number of numerical features ($d_{num}$), categorical features ($d_{cat}$), one-hot encoded (categorical) features ($d_{encoded}$), and the total number of features ($d_{total}$).
}
\label{tab:data-profile}
\resizebox{\textwidth}{!}{%
\begin{tabular}{@{}llrrrrrrrr@{}}
\toprule
\textbf{Dataset} & \textbf{Domain} & \(n_{total}\) & \(n_{train}\) & \(n_{val}\) & \(n_{test}\) & \(d_{num}\) & \(d_{cat}\) & \(d_{encoded}\) & \(d\) \\ \midrule
Adult & Finance & 32561 & 22792 & 3256 & 6513 & 6 & 8 & 99 & 105 \\
Electricity & Energy & 45312 & 31717 & 4532 & 9063 & 7 & 1 & 7 & 14 \\
COMPAS & Justice & 16644 & 11650 & 1665 & 3329 & 8 & 8 & 50 & 58 \\ \midrule
Higgs & Physics & 1000000 & 700000 & 100000 & 200000 & 28 & 0 & 0 & 28 \\ 
house\_16H & Finance & 22784 & 15948 & 2279 & 4557 & 16 & 0 & 0 & 16 \\
jm1 & Software & 10885 & 7619 & 1089 & 2177 & 21 & 0 & 0 & 21 \\
BreastCancer & Healthcare & 569 & 398 & 57 & 114 & 30 & 0 & 0 & 30 \\
WineQuality-White & Chemistry & 4898 & 3428 & 490 & 980 & 11 & 0 & 0 & 11 \\
WineQuality-Red & Chemistry & 1599 & 1119 & 160 & 320 & 11 & 0 & 0 & 11 \\
phoneme & Speech & 5404 & 3782 & 541 & 1081 & 5 & 0 & 0 & 5 \\
MiniBooNE & Physics & 130064 & 91044 & 13007 & 26013 & 50 & 0 & 0 & 50 \\ \bottomrule
\end{tabular}%
}
\end{table*}

\subsection{Adversarial Attacks}

Following the scope established in Section \ref{sec:taxonomy}, all adversarial examples in this benchmark are generated using white-box, untargeted attacks with full gradient access to the predictive models. Although adversarial methods for tabular data have begun to emerge, no systematic benchmark currently exists. 
Consequently, our selection process draws on established benchmarks in computer vision, with careful screening to ensure that the chosen methods can be extended to tabular settings.
We prioritise attack algorithms that:

\begin{enumerate}
    \item The selected attack methods should be applicable to tabular data.
    \item The selected attack methods should be designed for white-box attack.
\end{enumerate}

Based on these considerations, five widely used attack methods were selected (Table \ref{tab:attack-profile}). These span different levels of complexity and optimisation strategy, and all have demonstrated strong effectiveness in prior domains:
\begin{itemize}
    \item \textbf{Bounded gradient-based attacks}: FGSM~\citep{goodfellow2015explaining}, BIM~\citep{Kurakin2017adversarial} and PGD~\citep{madry2017towards}.
    \item \textbf{Unbounded optimisation-based attacks}: DeepFool~\citep{Moosavi2016deepfool} and C\&W~\citep{carlini2017towards}.
\end{itemize}

In addition, Gaussian noise is included as a baseline. This serves as a reference to compare deliberate adversarial perturbations against random noise injection. 


\begin{table*}[htp!]
\centering
\caption[Profiles of the five adversarial attack methods and the Gaussian noise baseline.]{Profiles of the five adversarial attack methods and the Gaussian noise baseline. Type abbreviations: G = Gradient-based, O = Optimisation-based, R = Random baseline; B = Bounded, U = Unbounded.}
\label{tab:attack-profile}
\begin{tabular}{@{}llll@{}}
\toprule
\textbf{Method} & \textbf{Type} & \textbf{Optimisation Strategy} & \textbf{Constraint} \\ \midrule
FGSM & G, B & One-step gradient sign & $\ell_\infty$ ball, $\epsilon$ budget \\
BIM & G, B & Iterative gradient sign & $\ell_\infty$ ball, $\epsilon$ budget  \\
PGD & G, B & Iterative with projection & $\ell_\infty$ ball, $\epsilon$ budget \\ \midrule
DeepFool & O, U & Iterative boundary projection & Closest decision boundary ($\ell_2$) \\
C\&W & O, U & Gradient descent with custom loss & Minimal $\ell_2$ perturbation \\ \midrule
Gaussian Noise & R, B & Random sampling from $\mathcal{N}(0,\sigma^2)$ & $\ell_\infty$ norm, $\epsilon$ budget \\ \bottomrule
\end{tabular}
\end{table*}

\subsection{Predictive Models} 

The choice of predictive models is guided by three criteria: \emph{diversity} (covering classical and modern architectures), \emph{interpretability} (balancing transparency with complexity), and \emph{performance} (ensuring competitive accuracy on tabular tasks). Based on these principles, we evaluate four representative models:

\begin{enumerate}
    \item \textbf{Logistic Regression (LR):} A simple, interpretable linear baseline that establishes performance lower bounds and vulnerability benchmarks.

    \item \textbf{Multilayer Perceptron (MLP):} A foundational neural network adept at capturing nonlinear patterns, offering a mid-complexity comparison point.

    \item \textbf{TabTransformer~\citep{huang2020tabtransformer}:} An attention-based model that processes tabular features via transformer layers, leveraging contextual relationships among features.

    \item \textbf{FTTransformer~\citep{gorishniy2021revisiting}:} A transformer-based architecture that tokenizes numerical and categorical features, enabling unified processing through self-attention mechanisms.

\end{enumerate}

\subsection{Evaluation Metrics}\label{subsec:metrics}

These metrics are grouped into two complementary categories: \emph{effectiveness}, which measures whether adversarial examples succeed in misleading the predictive model, and \emph{imperceptibility}, which evaluates how closely the perturbed instances adhere to the properties of imperceptible tabular data.

\subsubsection{Effectiveness} 
The primary metric of attack effectiveness is the attack success rate (ASR), 
which is defined as the proportion of adversarial examples that cause misclassification relative to the total number of perturbed instances. This metric provides a direct indication of how vulnerable the predictive model is to each attack.

\[
\mathrm{ASR}(f_\theta) 
= \frac{1}{n} \sum_{i=1}^n \mathbbm{1}\Big\{ \exists \, \bm{x}' \in \Delta(\bm{x}_i) : f_\theta(\bm{x}') \neq y_i \Big\}
\]

\subsubsection{Imperceptibility} 

Following our prior work \citep{he2025investigating}, we employ four quantitative imperceptibility metrics including proximity, sparsity, sensitivity and deviation. 

\begin{definition}[Proximity]
For an input $\bm{x} \in \mathcal{X}$ and its adversarial example $\tilde{\bm{x}} \in \Delta(\bm{x})$, 
the perturbation size can be quantified using $\ell_p$ norms:
\[
D_p(\tilde{\bm{x}},\bm{x}) = \lVert \tilde{\bm{x}} - \bm{x} \rVert_p =
\begin{cases}
      \Bigl(\sum_{j=1}^d \lvert \tilde{x}_j - x_j \rvert^p \Bigr)^{1/p}, & p \in \{1,2\}, \\[1ex]
      \max_{j \in \{1, \dots, d\}}  \lvert \tilde{x}_j - x_j \rvert, & p \to \infty .
\end{cases}
\]
Here, $d$ denotes the feature dimension. Over a dataset $\mathcal{D} = \{(\bm{x}_i,y_i)\}_{i=1}^n$, the \emph{average $\ell_p$ proximity} is
\[
\overline{D}_p(\mathcal{D}) = \frac{1}{n} \sum_{i=1}^n D_p(\tilde{x}_i, x_i),
\]
which measures the mean perturbation size across all adversarial examples. In this benchmark, we utilise the $\ell_2$ norm (Euclidean distance) for evaluating proximity.
\end{definition}

\begin{definition}[Sparsity]
For an input $\bm{x} \in \mathcal{X}$ and its adversarial example $\tilde{\bm{x}} \in \Delta(\bm{x})$, 
the \emph{sparsity} of the perturbation is the number of features whose values differ between $\bm{x}$ and $\tilde{\bm{x}}$. 
It is measured by the $\ell_0$ pseudo-norm:
\[
\mathrm{Spa}(\tilde{\bm{x}}, \bm{x}) = \lVert \tilde{\bm{x}} - \bm{x} \rVert_0 
= \sum_{j=1}^{d} \mathbb{1}\{ \tilde{x}_j \neq x_j \},
\]
where $d$ denotes the feature dimension.
The corresponding \emph{fractional sparsity} measures the fraction of features changed per instance:
\[
\mathrm{SpaFrac}(\tilde{\bm{x}},\bm{x}) = \frac{1}{d}\,\mathrm{Spa}(\tilde{\bm{x}},\bm{x}).
\]

Over a dataset $\mathcal{D} = \{(\bm{x}_i,y_i)\}_{i=1}^n$, 
the \emph{sparsity rate} is defined as the average fractional sparsity of adversarial examples:
\[
\mathrm{SpaR}(\mathcal{D}) = \frac{1}{n}\sum_{i=1}^n \mathrm{SpaFrac}(\tilde{\bm{x}}_i, \bm{x}_i)
= \frac{1}{nd}\sum_{i=1}^n \sum_{j=1}^{d} \mathbbm{1}\{\tilde{x}_{i,j} \neq x_{i,j}\}.
\]

\end{definition}

\begin{definition}[Deviation]
For an input $\bm{x} \in \mathcal{X}$, its adversarial example 
$\tilde{\bm{x}} \in \Delta(\bm{x})$, and the covariance matrix 
$\Sigma \in \mathbb{R}^{d \times d}$ of the data distribution $\mathcal{P}$ with mean $\bm{\mu}$, 
the \emph{deviation} of $\tilde{\bm{x}}$ from $\mathcal{P}$ 
is measured by the \emph{Mahalanobis Distance (MD)}:
\[
\mathrm{MD}(\tilde{\bm{x}}, \bm{\mu}) 
= \sqrt{(\tilde{\bm{x}} - \bm{\mu})^\top \, \Sigma^{-1} \, (\tilde{\bm{x}} - \bm{\mu})} .
\]

An adversarial example is considered an \emph{outlier} if
\[
\mathrm{MD}^2(\tilde{\bm{x}},\bm{\mu}) > \chi^2_{\alpha,d},
\]
where $\chi^2_{\alpha,d}$ denotes the critical value of the chi-squared distribution with $d$ degrees of freedom at significance level $\alpha$. Over a dataset $\mathcal{D} = \{(\bm{x}_i,y_i)\}_{i=1}^n \sim \mathcal{P}$, 
the \emph{outlier rate} is defined as
\[
\mathrm{OR}(\mathcal{D}) = \frac{1}{n}\sum_{i=1}^n 
\mathbbm{1}\!\left( \mathrm{MD}^2(\tilde{\bm{x}}_i, \bm{x}_i) > \chi^2_{\alpha,d} \right),
\]
which measures the proportion of adversarial examples flagged as outliers.
\end{definition}

\begin{definition}[Sensitivity]
Over a dataset $\mathcal{D} = \{(\bm{x}_i,y_i)\}_{i=1}^n$, where each feature vector $\bm{x}_i = (x_{i,1},\dots,x_{i,d})$ has $d$ features.
Among them, let $d_{\mathrm{num}} \leq d$ denote the number of numerical features.
For the $j$-th numerical feature, let $\sigma_j$ denote its standard deviation across the dataset $\mathcal{D}$. 
Given an input $\bm{x} \in \mathcal{X}$ and its adversarial example $\tilde{\bm{x}} \in \Delta(\bm{x})$, the \emph{sensitivity} of the perturbation is defined as the variance-normalised $\ell_1$ distance:
\[
\mathrm{SEN}(\tilde{\bm{x}},\bm{x}) 
= \sum_{j=1}^{d_{\mathrm{num}}} \frac{|\,\tilde{x}_j - x_j\,|}{\sigma_j}.
\]

The \emph{average sensitivity} over the dataset $\mathcal{D}$ is defined as
\[
\overline{\mathrm{SEN}}(\mathcal{D}) 
= \frac{1}{n}\sum_{i=1}^n \mathrm{SEN}(\tilde{\bm{x}}_i, \bm{x}_i)
= \frac{1}{n}\sum_{i=1}^n \sum_{j=1}^{d_{\mathrm{num}}} \frac{|\,\tilde{x}_j - x_j\,|}{\sigma_j},
\]
which measures the mean variance-normalised perturbation size 
across all adversarial examples.
\end{definition}

\begin{definition}[Imperceptibility Score]
The overall \emph{imperceptibility} of an adversarial attack is quantified by a composite metric that integrates four complementary measures: average $\ell_2$ proximity ($\overline{D}_2$), sparsity rate ($\mathrm{SpaR}$), outlier rate ($\mathrm{OR}$), and average sensitivity ($\overline{\mathrm{SEN}}$). 
Each captures a distinct aspect of how perturbations alter data realism and is assigned equal importance ($w_i=1$). 
To ensure comparability, all metrics are normalised to the range $[0,1]$: sparsity and outlier rates are inherently bounded, while $\overline{D}_2$ and $\overline{\mathrm{SEN}}$ are rescaled via min–max normalisation.
The resulting \emph{Imperceptibility Score (IS)} is defined as the weighted harmonic mean:
\[
\mathrm{IS} = 
\frac{4}{
    \frac{1}{\overline{D}_2} + 
    \frac{1}{\mathrm{SpaR}} + 
    \frac{1}{\mathrm{OR}} + 
    \frac{1}{\overline{\mathrm{SEN}}}
}.
\]
\end{definition}

\subsection{Implementation Details and Reproducibility}

To ensure transparency and reproducibility, all experiments in this benchmark were implemented in \texttt{Python~3.12} using \texttt{PyTorch~2.4} and executed on a workstation equipped with an NVIDIA~A100 GPU (80\,GB VRAM), AMD EPYC 7543 CPU, and 256\,GB RAM. The random seed was fixed to~42 across all stages, including dataset partitioning, model training, and adversarial generation, to guarantee deterministic behaviour. All code, pretrained models, and configuration files are publicly available in the companion repository\footnote{\url{https://github.com/ZhipengHe/TabAttackBench}}, ensuring full reproducibility of the reported results.

\paragraph{Data Partitioning and Preprocessing.}
Each dataset was partitioned using stratified sampling to preserve class distributions, allocating 70\% of the data for training, 10\% for validation, and 20\% for testing and adversarial evaluation. Constant or duplicate features were removed, missing values were imputed using the median for numerical features and the mode for categorical features, and categorical variables were one-hot encoded.
Numerical features were normalised to the $[0,1]$ range using min–max scaling. These consistent procedures maintain comparability across datasets while preserving their intrinsic statistical
characteristics.

\paragraph{Training Protocol.}
All predictive models were trained for~20~epochs with a batch size of~512, using the Adam optimiser with a learning rate of~$1\times10^{-3}$ and cross-entropy loss. Dropout~($p=0.2$) was applied to the MLP and transformer-based models for regularisation. Each transformer used~6~layers and~8~attention heads, with hidden-layer dimensionality fixed at~64~units to maintain comparable model capacity across architectures. The same hyperparameters were used for every dataset to facilitate consistent benchmarking rather than per-dataset optimisation. Table~\ref{tab:model-hp} summarises the main configurations.

\begin{table}[h]
\centering
\caption{Key hyperparameter configurations for the four predictive models. The `--' denotes that the corresponding hyperparameter does not apply in this model.}
\label{tab:model-hp}
\begin{tabular}{@{}lcccc@{}}
\toprule
\textbf{Parameter} & \textbf{LR} & \textbf{MLP} & \textbf{TabTransformer} & \textbf{FTTransformer} \\ \midrule
Training epochs & \multicolumn{4}{c}{20} \\
Batch size & \multicolumn{4}{c}{512} \\
Learning rate & \multicolumn{4}{c}{$1\times10^{-3}$} \\
Optimizer & \multicolumn{4}{c}{Adam} \\
Loss function & \multicolumn{4}{c}{Cross-entropy} \\ \midrule
Hidden layers & -- & 64$\!\rightarrow$32 & -- & -- \\
Activation & -- & ReLU & ReLU & ReLU \\
Dropout rate & -- & 0.2 & 0.2 & 0.2 \\ \midrule
Transformer depth & -- & -- & 6 & 6 \\
Attention heads & -- & -- & 8 & 8 \\
Embedding dimension & -- & -- & 64 & 64 \\ \bottomrule
\end{tabular}
\end{table}

\paragraph{Attack Configuration.}
All adversarial examples were generated under the white-box assumption with full gradient access. Five attack methods were implemented: FGSM, BIM, PGD, DeepFool, and C\&W, along with Gaussian noise as a random baseline. For FGSM, BIM, and PGD, perturbation budgets $\epsilon\in\{0.01,0.03,0.05,0.1,0.3,0.5,1.0\}$ were applied, with BIM and PGD using $T{=}10$~iterations and a relative step size~$\alpha=\epsilon/T$. DeepFool employed a maximum of~50~iterations and an overshoot factor of~0.02, while C\&W used~10~binary search steps, an initial constant~$c_0{=}0.001$, and confidence~$\kappa{=}0$. Each attack was run on all trained models, and evaluation metrics were averaged over~five independent runs.


\section{Evaluation}
\label{sec:evaluation}

Building on the benchmark design and implementation described in Section~\ref{sec:method}, this section presents the empirical evaluation of the proposed framework.

\subsection{Task 1: Effectiveness Evaluation}\label{subsec:RQ1}


We begin by evaluating model accuracy to establish a baseline prior to adversarial testing. If accuracy falls below $60$\%, a model may be trivially deceived, rendering robustness evaluation less meaningful.
Table~\ref{tab:accuracy} reports the accuracy of the four predictive models across the eleven datasets. The three deep learning models generally outperform LR, although LR achieves comparable results on \emph{Adult}, \emph{jm1}, and \emph{COMPAS}. Overall, All models exceed $63$\% accuracy, providing a reliable basis for subsequent adversarial evaluation.

\begin{table}[!htp]
\centering
\caption{Accuracy of four predictive models across 11 datasets. Deep learning models generally outperform LR and all models exceed 63\% accuracy.}
\label{tab:accuracy}
\begin{tabular}{@{}lrrrr@{}}
\toprule
\textbf{Dataset}   & \textbf{LR} & \textbf{MLP} & \textbf{TabTransformer} & \textbf{FTTransformer} \\ \midrule
\textit{Adult}             & 0.834                                  & 0.8337                  & 0.8328                             & 0.799                             \\
\textit{BreastCancer}      & 0.9386                                 & 0.9737                  & 0.9035                             & 0.9737                            \\
\textit{Compas}            & 0.6654                                 & 0.6738                  & 0.7053                             & 0.6858                            \\
\textit{Electricity}       & 0.6607                                 & 0.7635                  & 0.762                              & 0.7712                            \\
\textit{Higgs}             & 0.6366                                 & 0.7234                  & 0.6951                             & 0.7296                            \\
\textit{MiniBooNE}         & 0.7724                                 & 0.8372                  & 0.8497                             & 0.8402                            \\
\textit{WineQuality-Red}   & 0.7219                                 & 0.7344                  & 0.7281                             & 0.7344                            \\
\textit{WineQuality-White} & 0.6745                                 & 0.7469                  & 0.7316                             & 0.752                             \\
\textit{house\_16H}        & 0.7029                                 & 0.8578                  & 0.8251                             & 0.8466                            \\
\textit{jm1}               & 0.8075                                 & 0.8098                  & 0.8066                             & 0.8107                            \\
\textit{phoneme}           & 0.7095                                 & 0.7872                  & 0.7882                             & 0.8002                            \\ \bottomrule
\end{tabular}
\end{table}


\subsubsection{Effectiveness: Mixed Datasets}  

\begin{figure}[!ht]
    \centering
    \includegraphics[width=\linewidth]{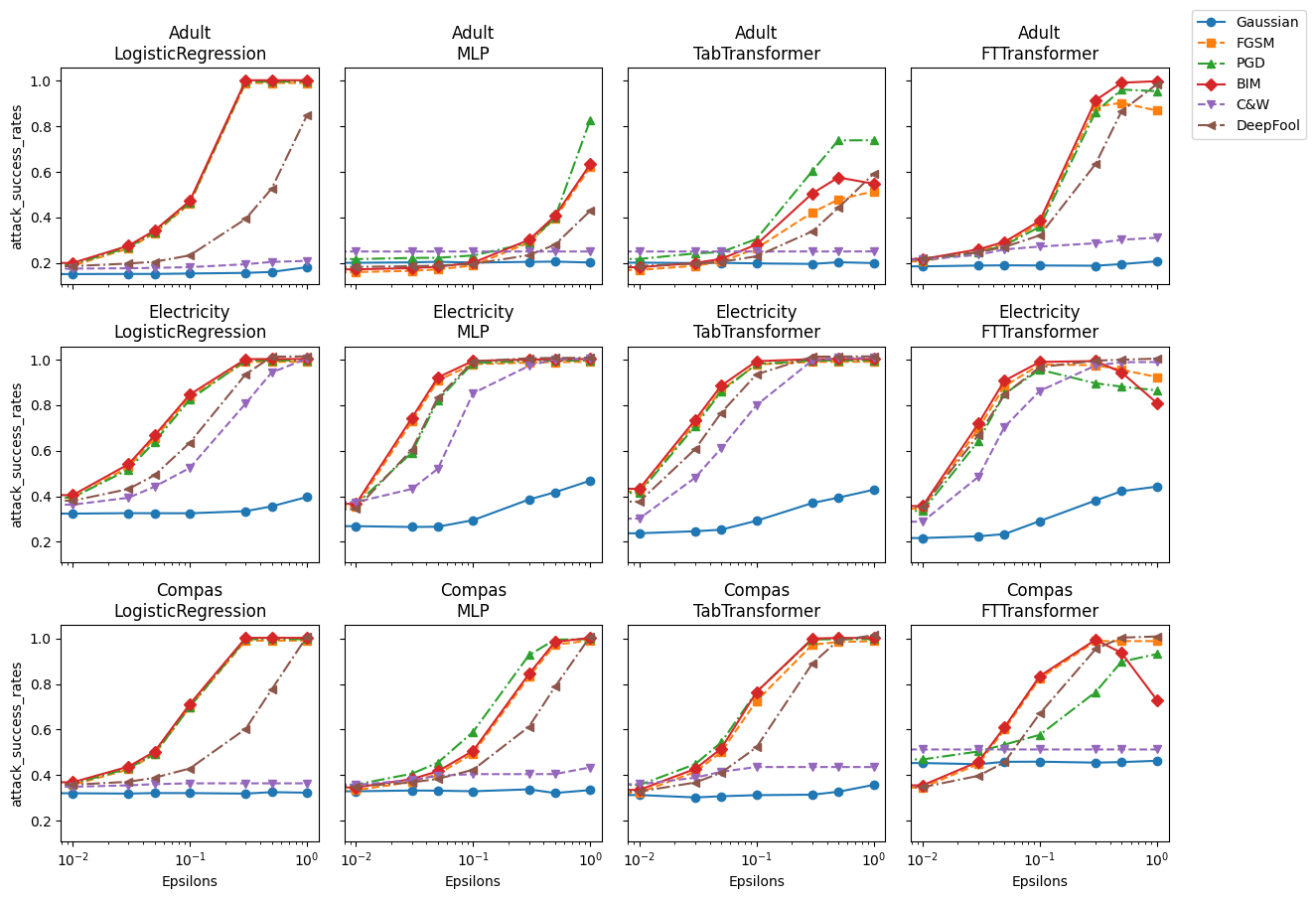}
    \caption{Attack success rate (ASR) of evaluated attack methods on all three \emph{mixed} datasets (\emph{Adult}, \emph{COMPAS}, \emph{Electricity}) and four models. \emph{Electricity} shows uniformly high vulnerability across all attacks and models, while Adult and COMPAS reveal clear differences between $\ell_\infty$-based (FGSM, BIM, PGD) and $\ell_2$-based (DeepFool, C\&W attack) methods. Transformer models generally require larger perturbations to reach comparable success rates, indicating greater robustness than LR and MLP.}
    \label{fig:asr_by_epsilon_mixed}
\end{figure}

Figure \ref{fig:asr_by_epsilon_mixed} shows that the \emph{Electricity} dataset demonstrates unique characteristics compared to other mixed datasets. Across all model architectures, every attack achieves consistently high success rates, all approaching 100\% success on LR, MLP, and TabTransformer. This indicates that the feature distribution or decision boundaries in \emph{Electricity} are especially susceptible to adversarial perturbations. Unlike \emph{Adult} and \emph{COMPAS}, where attack types diverge in performance, Electricity exhibits broadly uniform vulnerability.

For \emph{Adult} and \emph{COMPAS}, clearer differences emerge. The \(\ell_\infty\)-based bounded attacks (FGSM, BIM, PGD) consistently outperform \(\ell_2\)-based unbounded attacks (C\&W attack and DeepFool) on these datasets. 
This suggests that bounded perturbations under the \(\ell_\infty\) ball are particularly effective for mixed-type tabular data, as they can alter critical features directly without being constrained by overall perturbation magnitude.

Architecturally, Transformer-based models exhibit greater robustness than LR and MLP. Both TabTransformer and FTTransformer require larger $\epsilon$ values before achieving comparable attack success rate (ASR), especially on \emph{Adult} dataset. 
This points to attention mechanisms and deeper architectures providing some inherent robustness to adversarial manipulation in mixed tabular data.

\subsubsection{Effectiveness: Numerical Datasets} 

\begin{figure}[htp!]
    \centering
    \includegraphics[width=\linewidth]{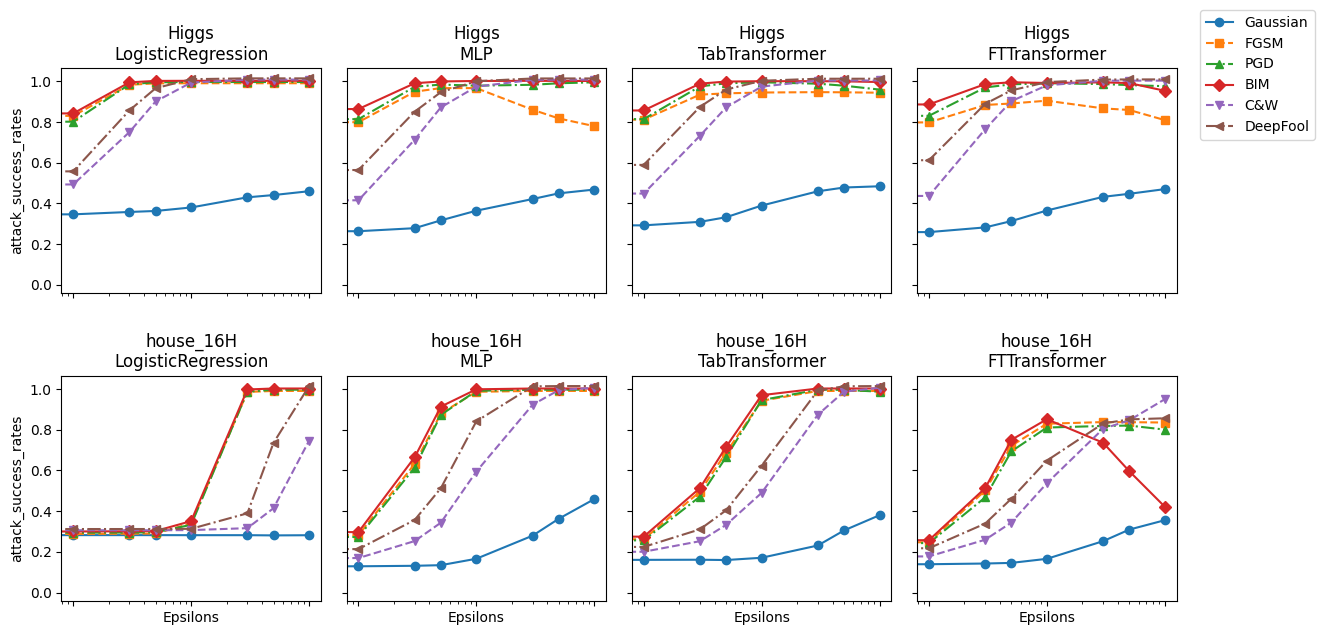}
    \caption{Attack success rate (ASR) of evaluated attack methods on two (out of eight) \textit{numerical} datasets and four ML models.}
    \label{fig:asr_by_epsilon_num_1}
\bigskip
    \centering
    \includegraphics[width=\linewidth]{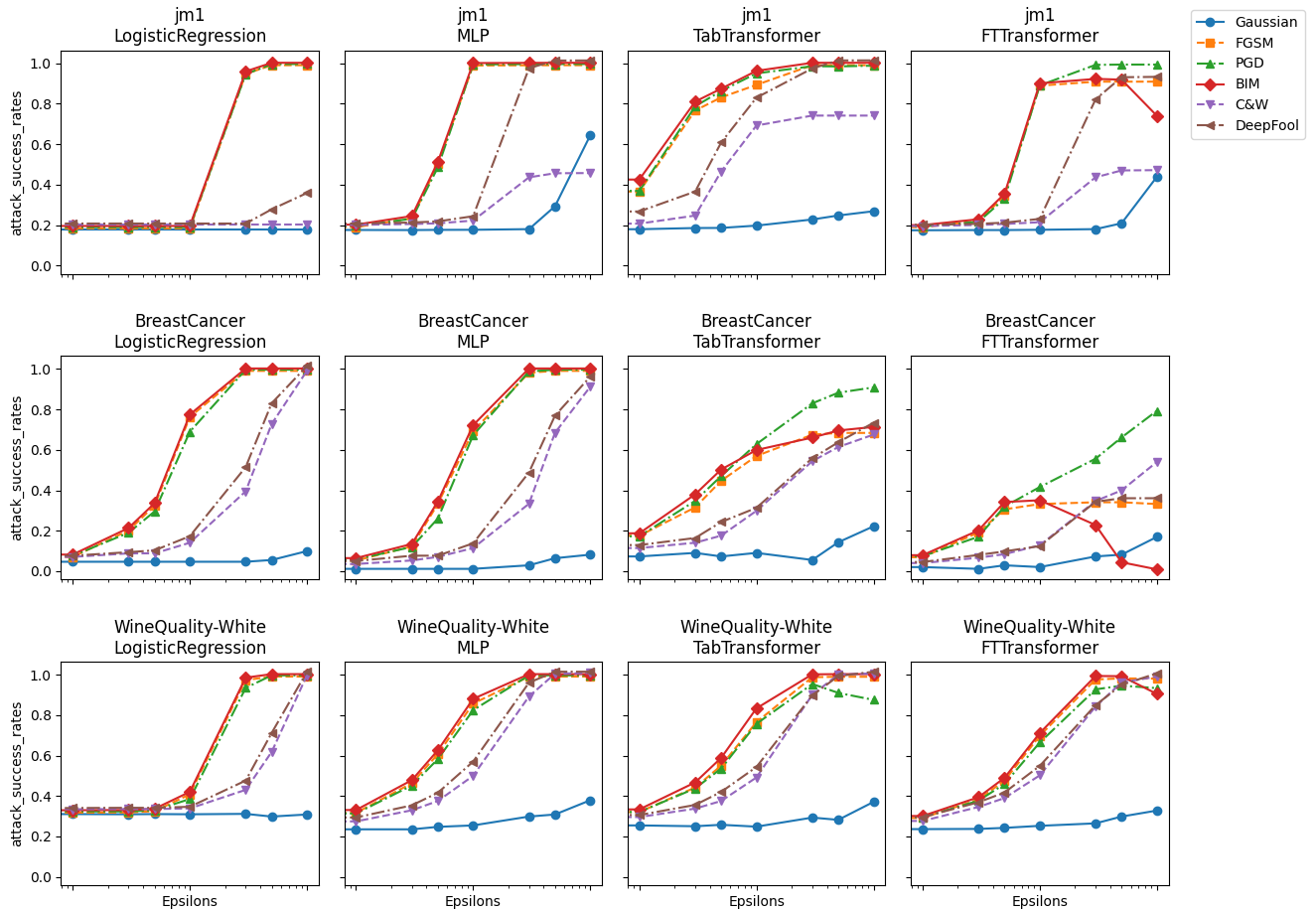}
    \caption{(Cont.) Attack success rate (ASR) of evaluated attack methods on another three (out of eight) \textit{numerical} datasets and four ML models.}
    \label{fig:asr_by_epsilon_num_2}
\end{figure}

\begin{figure}[!ht]
    \centering
    \includegraphics[width=\linewidth]{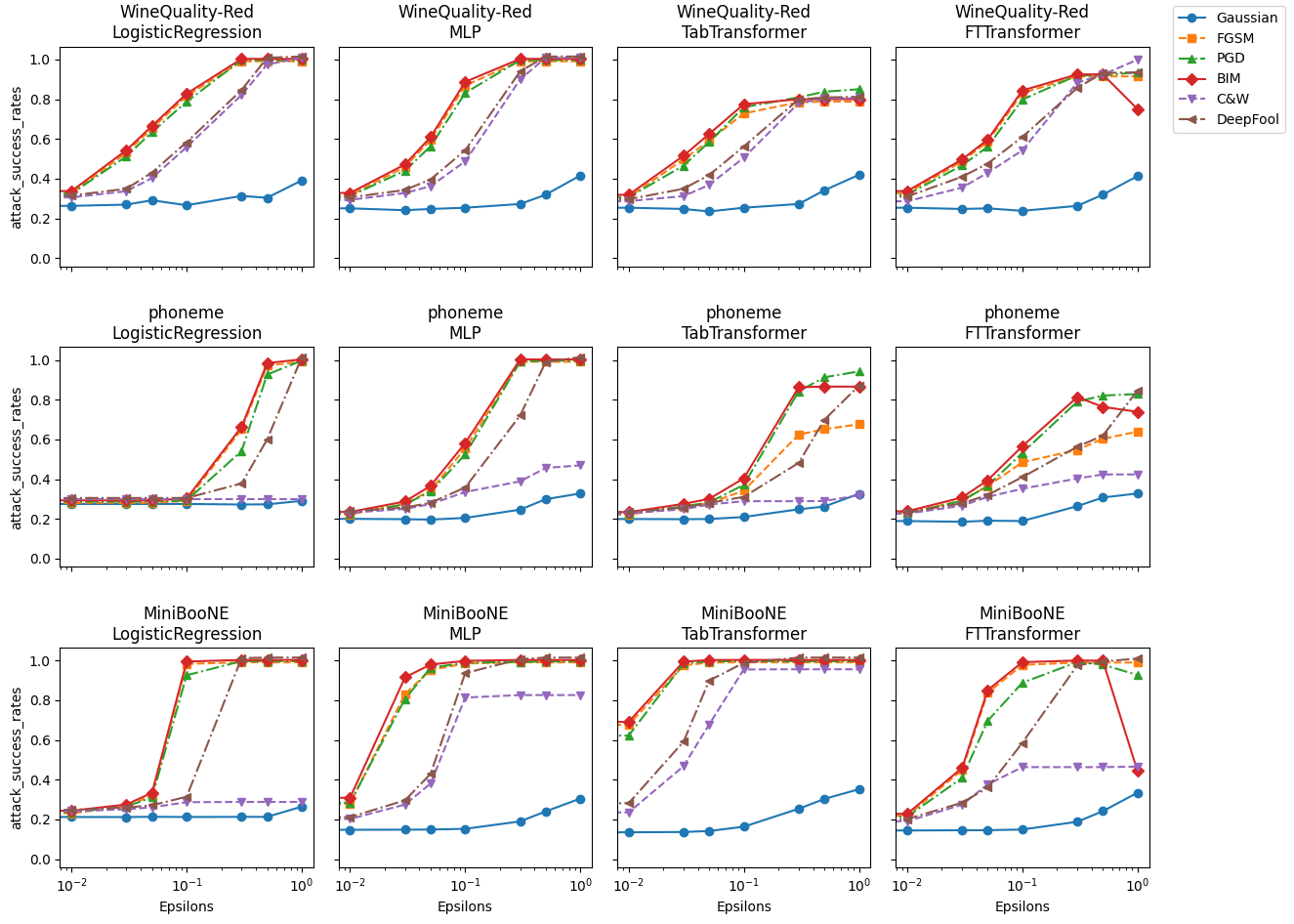}
    \caption{(Cont.) Attack success rate (ASR) of evaluated attack methods on the remaining three (out of eight) \textit{numerical} datasets and four ML models.}
    \label{fig:asr_by_epsilon_num_3}
\end{figure}

Figure \ref{fig:asr_by_epsilon_num_1}, \ref{fig:asr_by_epsilon_num_2} and \ref{fig:asr_by_epsilon_num_3} summarise results on the eight numerical datasets, which exhibit more consistent patterns than mixed datasets, while still showing dataset-specific behaviours.

On \emph{Higgs} and \emph{house\_16H} (Figure \ref{fig:asr_by_epsilon_num_1}), all three $\ell_\infty$-based attacks (FGSM, BIM, PGD) produce nearly identical success curves across models. 
The \emph{BreastCancer} dataset (Figure \ref{fig:asr_by_epsilon_num_2}) provides the most diverse response to different attack methods. Here, PGD consistently outperforms other attacks against transformer-based models. FTTransformer under BIM shows a non-monotonic trend, with attack success occasionally decreasing as $\epsilon$ increases. 
The \emph{WineQuality} datasets (\emph{Red} and \emph{White}; Figure \ref{fig:asr_by_epsilon_num_2} and \ref{fig:asr_by_epsilon_num_3}) show that \(\ell_\infty\)-based attacks require substantially lower epsilon values to achieve high success rates compared to \(\ell_2\)-based approaches. DeepFool eventually achieves comparable success but requires much larger perturbation budgets, limiting its efficiency. The \emph{phoneme} and \emph{MiniBooNE} datasets (Figure \ref{fig:asr_by_epsilon_num_3}) further confirm the superiority of \(\ell_\infty\)-based attacks, with all three methods (FGSM, PGD, BIM) demonstrating nearly identical performance trajectories and consistently outperform $\ell_2$-based approaches across models.

Across the numerical datasets, LR models are consistently the most vulnerable, often showing sharp threshold effects where ASR increases abruptly at particular $\epsilon$ values. This reflects the ease with which linear decision boundaries can be exploited by adversarial perturbations. In contrast, MLP and transformer models generally resist small perturbations and only degrade at larger $\epsilon$ values, though the exact patterns vary by dataset. The C\&W attack exhibits the least consistent behaviour, sometimes barely outperforming random noise yet occasionally achieving competitive success on specific dataset–model combinations. Its strong dependence on data and architecture characteristics limits its reliability as a general-purpose attack for tabular domains.

\begin{findingbox}[Overall insights for Task 1: Effectiveness Evaluation]
$\ell_\infty$ bounded attacks (FGSM, BIM, PGD) prove to be the most consistently effective, often achieving high success rates with relatively small attack budgets. 
LR models are generally the most vulnerable, while MLP and transformer architectures require larger perturbations before showing comparable degradation. 
Although more sophisticated methods such as PGD and C\&W can occasionally outperform simpler approaches, the results indicate that even basic gradient-based attacks (e.g., FGSM) are already highly effective against tabular classifiers.
\end{findingbox}

\subsection{Task 2: Imperceptibility Assessment}

To evaluate imperceptibility in a fair and consistent manner, we first establish a systematic strategy for selecting attack budgets. For each model--dataset--attack combination, we identify the $\epsilon$ value at which the ASR curve first reaches a plateau, beyond which further increases yield little or no improvement. The specific attack budgets selected through this methodology are detailed in \ref{appendix:epsilons}. However, direct comparisons across different models and datasets would be misleading, since the optimal \(\epsilon\) values vary significantly between these contexts. To address this, we identify the most frequently occurring $\epsilon$ value for each attack method across all experiments. These representative settings, listed in Table \ref{tab:epsilon_by_attacks}, are then adopted as our standardised benchmark parameters for subsequent analyses in \textbf{Task 2} and \textbf{Task 3}. Full imperceptibility results can be found in \ref{appendix:full}.

\begin{table}[!htp]
\centering
\caption{Representative $\epsilon$ values for each attack method, identified as the most frequent across all model–dataset combinations. 
These standardised settings balance effectiveness with comparability, and are used in the imperceptibility (\textbf{Task 2}) and trade-off (\textbf{Task 3}) analyses.}
\label{tab:epsilon_by_attacks}
\begin{tabular}{@{}crrrrrr@{}}
\toprule
  \textbf{Attacks}  & Gaussian & FGSM & BIM & PGD & C\&W & DeepFool \\ \midrule
\textbf{\(\epsilon\)} & 1 & 0.3 & 0.3 & 0.3 & 1 & 1 \\ \bottomrule
\end{tabular}%
\end{table}

\subsubsection{Task 2.1: Sparsity}

Our analysis of sparsity patterns reveals distinct behavioural characteristics among adversarial attack methods while highlighting the influence of dataset dimensionality and model architecture. The \emph{sparsity rate} measure quantifies the proportion of features modified in adversarial examples, with higher values indicating more features being perturbed.

\paragraph{Sparsity: Mixed Datasets}

\begin{figure}[htb!]
\begin{subfigure}[t]{0.32\linewidth}
    \centering
    \includegraphics[trim={0.5cm 0 2cm 0},clip,width=\linewidth]{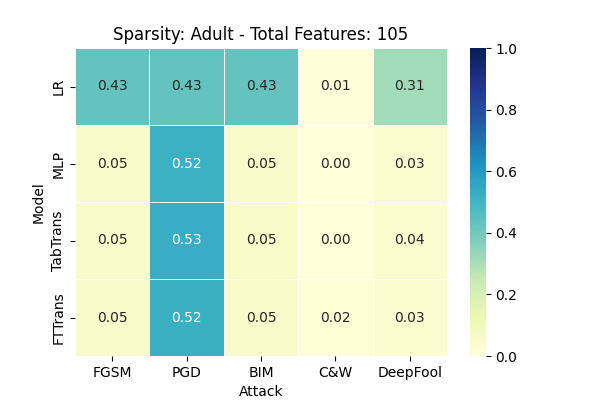}
    \caption{All features}
    \label{fig:spa:Adult:all}
\end{subfigure}
\begin{subfigure}[t]{0.32\linewidth}
    \centering
    \includegraphics[trim={0.5cm 0 2cm 0},clip,width=\linewidth]{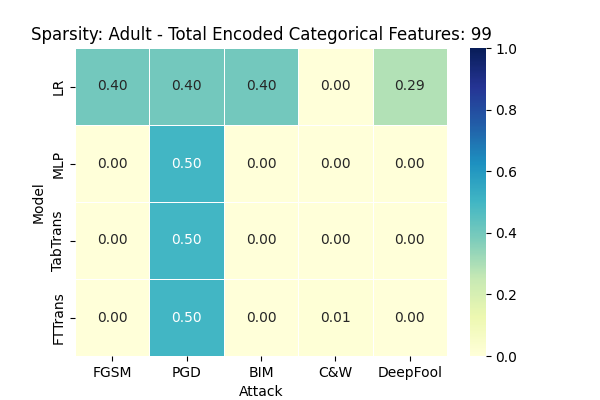}
    \caption{Categorical Features}
    \label{fig:spa:Adult:cat}
\end{subfigure}
\begin{subfigure}[t]{0.32\linewidth}
    \centering
    \includegraphics[trim={0.5cm 0 2cm 0},clip,width=\linewidth]{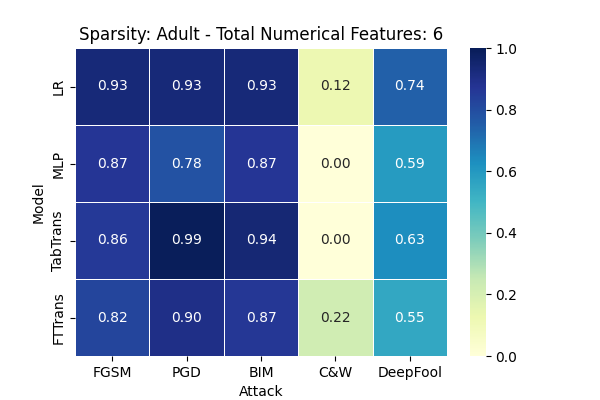}
    \caption{Numerical Features}
    \label{fig:spa:Adult:num}
\end{subfigure}
    \caption{Sparsity results (\emph{sparsity rate}) of five evaluated attack methods and four models on the \emph{Adult} dataset. Despite categorical dominance (99 of 105 features), most attacks primarily perturb numerical features, leaving categorical ones virtually untouched.}
    \label{fig:spa:Adult}
\end{figure}

\begin{figure}[htb!]
\begin{subfigure}[t]{0.32\linewidth}
    \centering
    \includegraphics[trim={0.5cm 0 2cm 0},clip,width=\linewidth]{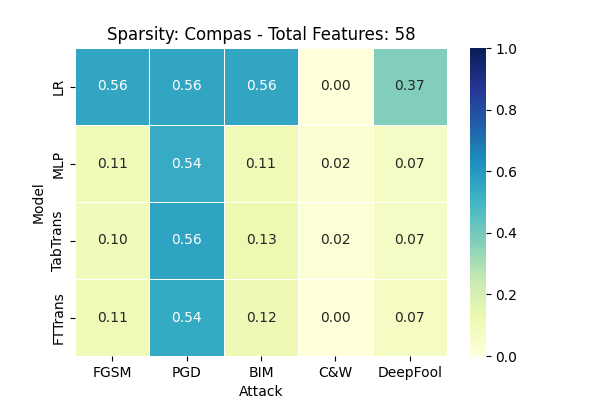}
    \caption{All features}
    \label{fig:spa:Compas:all}
\end{subfigure}
\begin{subfigure}[t]{0.32\linewidth}
    \centering
    \includegraphics[trim={0.5cm 0 2cm 0},clip,width=\linewidth]{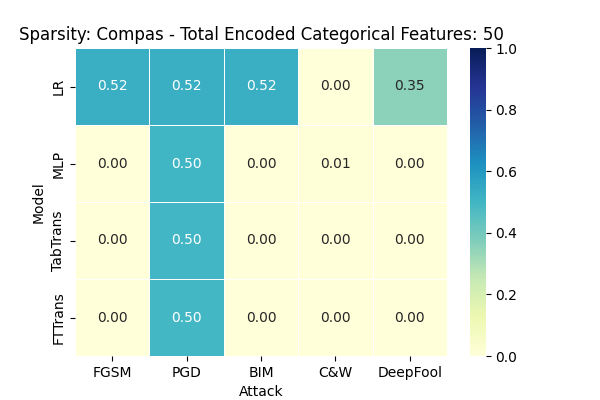}
    \caption{Categorical Features}
    \label{fig:spa:Compas:cat}
\end{subfigure}
\begin{subfigure}[t]{0.32\linewidth}
    \centering
    \includegraphics[trim={0.5cm 0 2cm 0},clip,width=\linewidth]{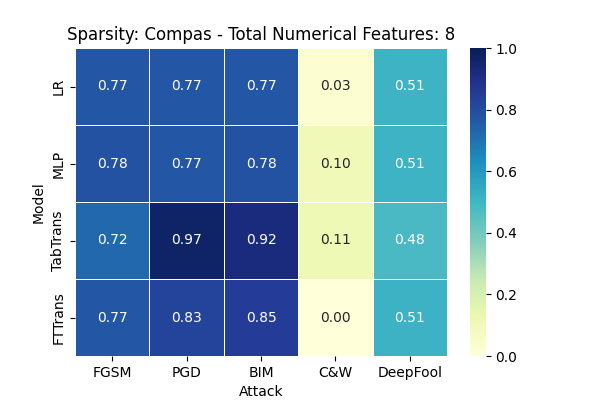}
    \caption{Numerical Features}
    \label{fig:spa:Compas:num}
\end{subfigure}
    \caption{Sparsity results (\emph{sparsity rate}) of evaluated attack methods and four models on the \emph{COMPAS} dataset. As with \emph{Adult}, attacks largely ignore categorical features, though LR shows moderate categorical sparsity, highlighting model-specific vulnerability.}
    \label{fig:spa:Compas}
\end{figure}

\begin{figure}[htb!]
\begin{subfigure}[t]{0.32\linewidth}
    \centering
    \includegraphics[trim={0.5cm 0 2cm 0},clip,width=\linewidth]{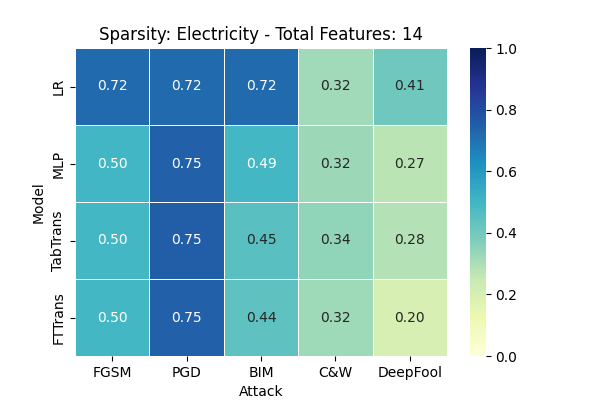}
    \caption{All features}
    \label{fig:spa:Elec:all}
\end{subfigure}
\begin{subfigure}[t]{0.32\linewidth}
    \centering
    \includegraphics[trim={0.5cm 0 2cm 0},clip,width=\linewidth]{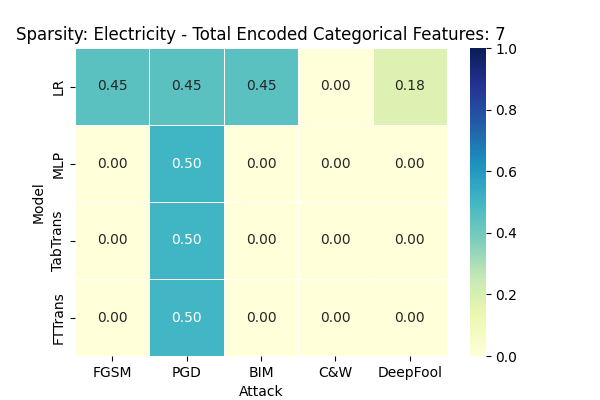}
    \caption{Categorical Features}
    \label{fig:spa:Elec:cat}
\end{subfigure}
\begin{subfigure}[t]{0.32\linewidth}
    \centering
    \includegraphics[trim={0.5cm 0 2cm 0},clip,width=\linewidth]{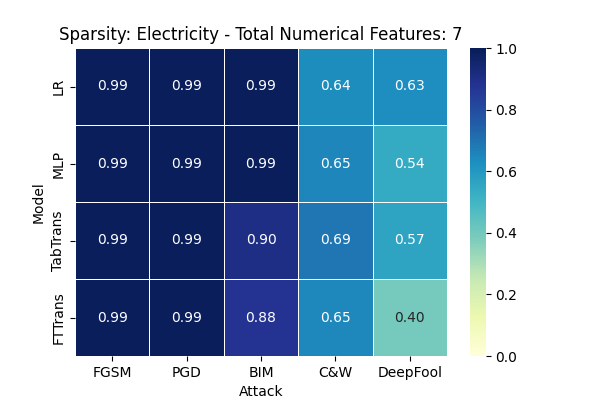}
    \caption{Numerical Features}
    \label{fig:spa:Elec:num}
\end{subfigure}
    \caption{Sparsity results (\emph{sparsity rate}) of evaluated attack methods and four models on the \emph{Electricity} dataset. Even with a balanced feature space (7 categorical, 7 numerical), attacks strongly prioritise numerical features, with categorical perturbations near zero.}
    \label{fig:spa:Elec}
\end{figure}

Results for the mixed datasets (Figures \ref{fig:spa:Adult}, \ref{fig:spa:Compas} and \ref{fig:spa:Elec}) reveal systematic differences between categorical and numerical features.  

A consistent pattern emerges across most attacks: FGSM, BIM, C\&W, and DeepFool strongly prioritise numerical features when targeting neural architectures (MLP, TabTransformer, FT-Transformer). In categorical-dominant datasets such as \emph{Adult} (105 features, 99 categorical; Figure \ref{fig:spa:Adult}) and \emph{COMPAS} (58 features, 50 categorical; Figure \ref{fig:spa:Adult}), these methods perturb numerical features heavily (72--99\% sparsity) while leaving categorical ones nearly untouched (0--1\%). This systematic selectivity indicates that these attacks inherently favour continuous variables, largely ignoring one-hot encoded categorical dimensions despite their prevalence.

Among the evaluated attacks, PGD stands out as the only method that consistently modifies categorical features, achieving around 50\% sparsity across all models and datasets. Its projection mechanism appears uniquely suited to navigating the discrete one-hot space, distinguishing it from other $\ell_\infty$-based bounded attacks.

Turning to model-specific effects, the behaviour of LR differs markedly from neural models. Under FGSM, PGD, and BIM, LR exhibits moderate categorical sparsity (40--52\%) across all mixed datasets. This suggests that LR encodes information in a way that makes categorical features more vulnerable, an effect particularly pronounced in \emph{COMPAS}, where categorical sparsity under FGSM and BIM reaches 52\%, compared to 0\% for neural networks.  

The \emph{Electricity} dataset (14 features: 7 categorical, 7 numerical; Figure \ref{fig:spa:Elec}) illustrates attack behaviour in a balanced feature space. Even here, $\ell_\infty$-based attacks retain their numerical preference: FGSM and BIM achieve 99\% sparsity on numerical features while leaving categorical features unaltered (0\%) when attacking neural models. This persistence confirms the algorithmic bias of these methods toward continuous features. $\ell_2$-based attacks show mixed outcomes. C\&W generally produces very low overall sparsity (0--2\%) across mixed datasets but achieves moderate sparsity on numerical features in \emph{Electricity} (64--69\%). DeepFool shows an even stronger numerical preference, altering 40--74\% of numerical features while leaving categorical features virtually untouched (0--0.4\%). These patterns underscore the difficulty of $\ell_2$-based methods in handling one-hot encoded categorical variables.

These findings collectively demonstrate that with the exception of PGD, current adversarial attacks on tabular data overwhelmingly target numerical features while almost entirely neglecting categorical dimensions, regardless of their prevalence.





\paragraph{Sparsity: Numerical Datasets} 

\begin{figure}[tb!]
\begin{subfigure}[t]{0.32\linewidth}
    \centering
    \includegraphics[trim={0.5cm 0 2cm 0},clip,width=\linewidth]{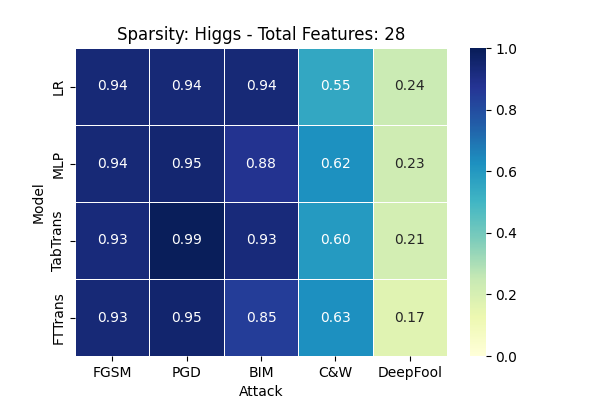}
    \caption{Higgs}
    \label{fig:spa:Higgs}
\end{subfigure}
\begin{subfigure}[t]{0.32\linewidth}
    \centering
    \includegraphics[trim={0.5cm 0 2cm 0},clip,width=\linewidth]{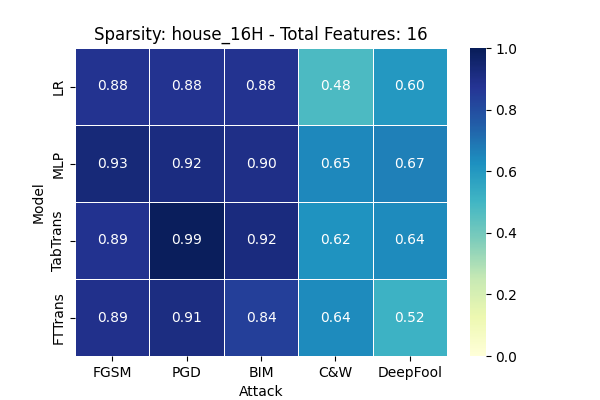}
    \caption{house\_16H}
    \label{fig:spa:house_16H}
\end{subfigure}
\begin{subfigure}[t]{0.32\linewidth}
    \centering
    \includegraphics[trim={0.5cm 0 2cm 0},clip,width=\linewidth]{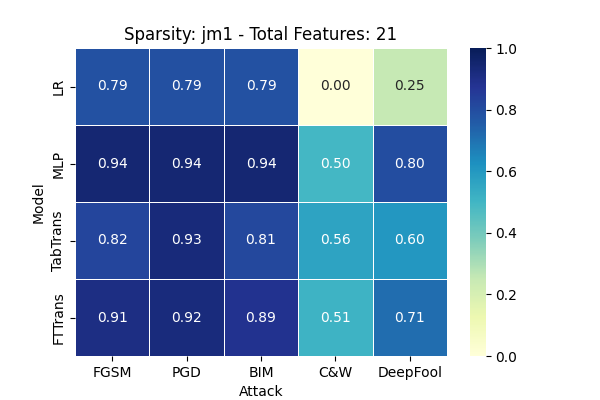}
    \caption{jm1}
    \label{fig:spa:jm1}
\end{subfigure}
\begin{subfigure}[t]{0.32\linewidth}
    \centering
    \includegraphics[trim={0.5cm 0 2cm 0},clip,width=\linewidth]{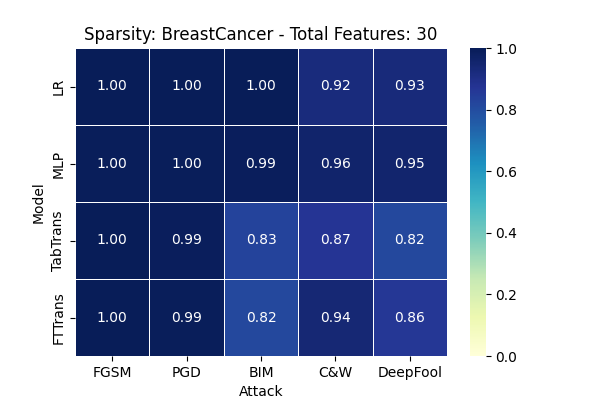}
    \caption{BreastCancer}
    \label{fig:spa:BreastCancer}
\end{subfigure}
\begin{subfigure}[t]{0.32\linewidth}
    \centering
    \includegraphics[trim={0.5cm 0 2cm 0},clip,width=\linewidth]{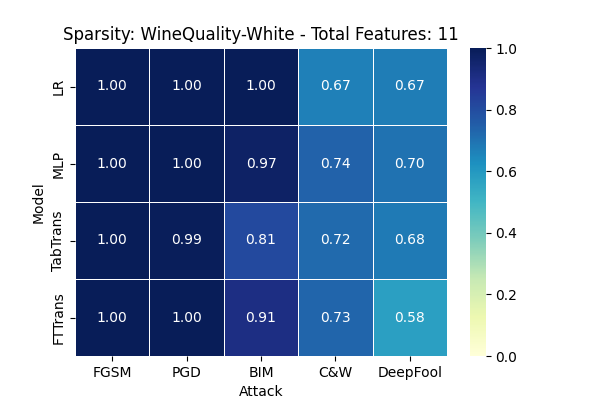}
    \caption{WineQuality-White}
    \label{fig:spa:WineQuality-White}
\end{subfigure}
\begin{subfigure}[t]{0.32\linewidth}
    \centering
    \includegraphics[trim={0.5cm 0 2cm 0},clip,width=\linewidth]{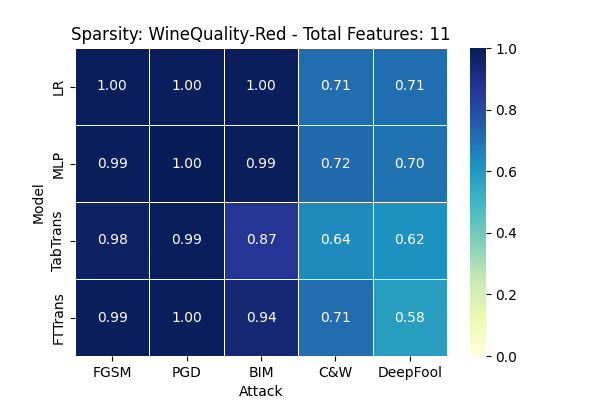}
    \caption{WineQuality-Red}
    \label{fig:spa:WineQuality-Red}
\end{subfigure}

\hspace{0.08\linewidth}
\begin{subfigure}[t]{0.32\linewidth}
    \centering
    \includegraphics[trim={0.5cm 0 2cm 0},clip,width=\linewidth]{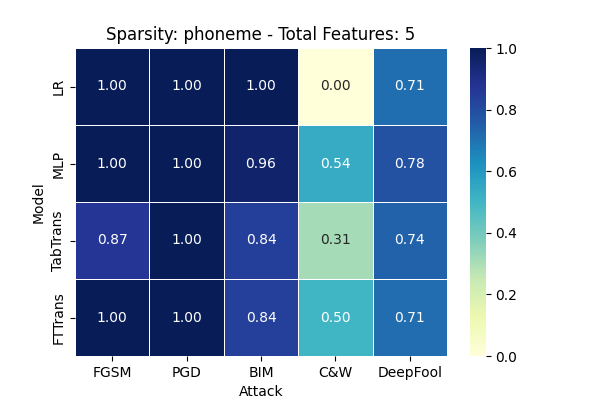}
    \caption{phoneme}
    \label{fig:spa:phoneme}
\end{subfigure}
\hspace{0.12\linewidth}
\begin{subfigure}[t]{0.32\linewidth}
    \centering
    \includegraphics[trim={0.5cm 0 2cm 0},clip,width=\linewidth]{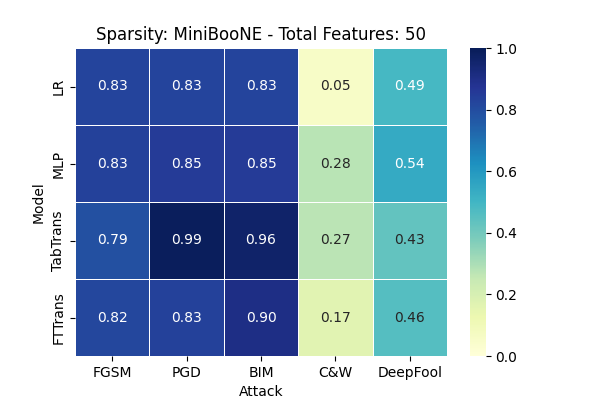}
    \caption{MiniBooNE}
    \label{fig:spa:MiniBooNE}
\end{subfigure}
    \caption{Sparsity results of evaluated attack methods and four models on eight \emph{numerical} datasets. $\ell_\infty$ attacks (FGSM, BIM, PGD) consistently modify nearly all features, while $\ell_2$ attacks (C\&W, DeepFool) are more selective, with sparsity strongly influenced by dataset dimensionality.}
    \label{fig:spa:num_1}
\end{figure}

For numerical datasets, as illustrated in Figure \ref{fig:spa:num_1}, we observe distinct patterns of feature perturbation across different attack methods. 

The three $\ell_\infty$-based attacks (FGSM, PGD, and BIM) consistently modify nearly all available features, producing sparsity rates between 80--100\% across models and datasets. This tendency holds regardless of feature dimensionality, from the low-dimensional \emph{phoneme} dataset (5 features, Figure \ref{fig:spa:phoneme}) to the high-dimensional \emph{MiniBooNE} dataset (50 features, Figure \ref{fig:spa:MiniBooNE}). Among them, PGD is particularly aggressive, frequently driving sparsity close to 99--100\%, especially on transformer-based models.

In contrast, the two $\ell_2$-based optimisation attacks show far more selective behaviour. C\&W is the most variable: in some cases (e.g., LR on \emph{jm1}, Figure \ref{fig:spa:jm1}) it modifies almost no features at all (0\%), while in others it perturbs around half (50--74\%). DeepFool generally occupies a middle ground, with sparsity rates between 17--80\%, and its most selective behaviour is observed on the \emph{Higgs} dataset (17--24\%, Figure \ref{fig:spa:Higgs}). Overall, C\&W appears highly sensitive to dataset--model combinations, aggressively pruning its perturbations to a small subset of influential features, while DeepFool is more stable across architectures but still adapts its sparsity according to dataset characteristics.

Feature dimensionality exerts a clear influence on $\ell_2$ attacks. Both C\&W and DeepFool reduce their sparsity on larger datasets, concentrating on fewer features as dimensionality grows. For example, on the 50-feature \emph{MiniBooNE} dataset, C\&W modifies only 5--28\% of features, compared with 50--74\% on small datasets such as \emph{phoneme} (5 features). An important counterexample is the \emph{BreastCancer} dataset (Figure \ref{fig:spa:BreastCancer}), where all methods exhibit very high sparsity rates ($>80$\%), suggesting that every feature contributes strongly to classification and thus becomes a viable target.

Model architecture further shapes sparsity outcomes. LR exhibits the greatest variability, particularly under C\&W and DeepFool, with sparsity rates ranging from negligible to substantial depending on the dataset. TabTransformer also shows notable fluctuations across attack types, whereas MLP yields more consistent sparsity patterns.

\begin{findingbox}[Overall insights for Task 2.1: Sparsity]
Sparsity analysis shows a clear divide: $\ell_\infty$-based attacks (FGSM, BIM, PGD) modify nearly all features, while $\ell_2$-based methods (C\&W, DeepFool) are far more selective but inconsistent across datasets and models. 
Notably, only PGD reliably perturbs categorical features, with most attacks overwhelmingly targeting numerical ones. This indicates that current methods achieve effectiveness mainly through broad numerical changes, limiting their imperceptibility in mixed-feature tabular data.
\end{findingbox}

\subsubsection{Task 2.2: Proximity}

\begin{figure}[htp!]
\begin{subfigure}[t]{0.32\linewidth}
    \centering
    \includegraphics[trim={0.5cm 0 2cm 0},clip,width=\linewidth]{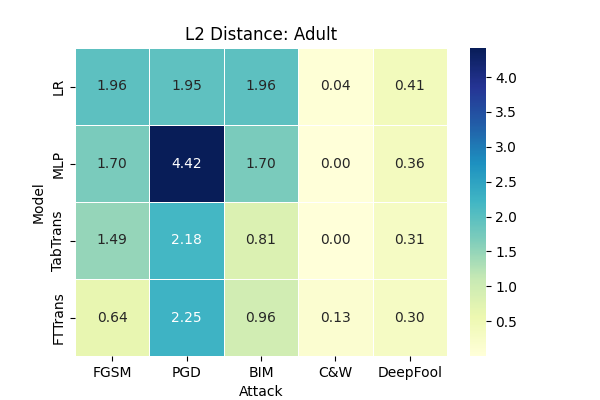}
    \caption{Adult}
    \label{fig:pro:Adult}
\end{subfigure}
\begin{subfigure}[t]{0.32\linewidth}
    \centering
    \includegraphics[trim={0.5cm 0 2cm 0},clip,width=\linewidth]{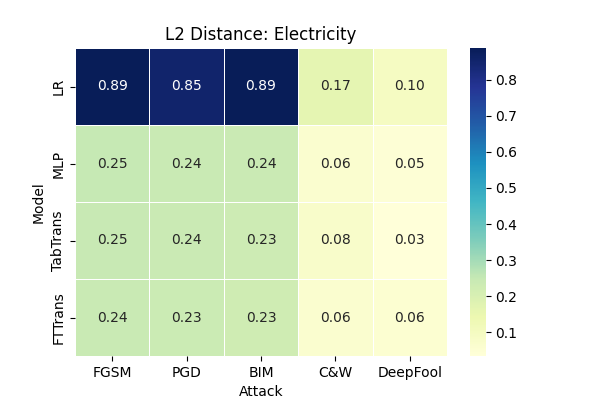}
    \caption{Electricity}
    \label{fig:pro:Elec}
\end{subfigure}
\begin{subfigure}[t]{0.32\linewidth}
    \centering
    \includegraphics[trim={0.5cm 0 2cm 0},clip,width=\linewidth]{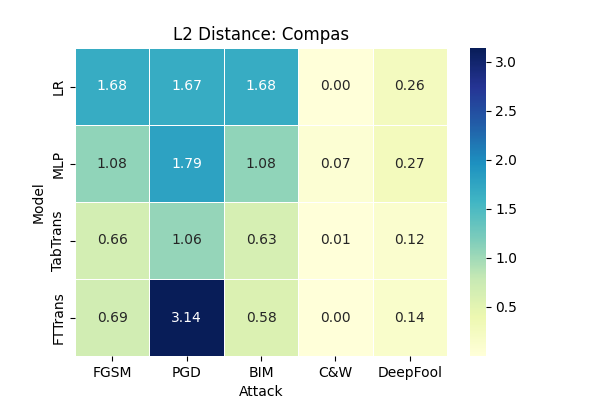}
    \caption{COMPAS}
    \label{fig:pro:Compas}
\end{subfigure}
    \caption{Proximity results (\emph{$\ell_2$ distance}) of evaluated attack methods and four models on all three \emph{mixed} datasets. $\ell_2$-based unbounded attacks (C\&W, DeepFool) consistently produce adversarial examples much closer to the originals, while $\ell_\infty$-based methods (FGSM, BIM, PGD) generate larger perturbations. }
    \label{fig:pro:mixed}
\end{figure}

\begin{figure}[tb!]
\begin{subfigure}[t]{0.32\linewidth}
    \centering
    \includegraphics[trim={0.5cm 0 2cm 0},clip,width=\linewidth]{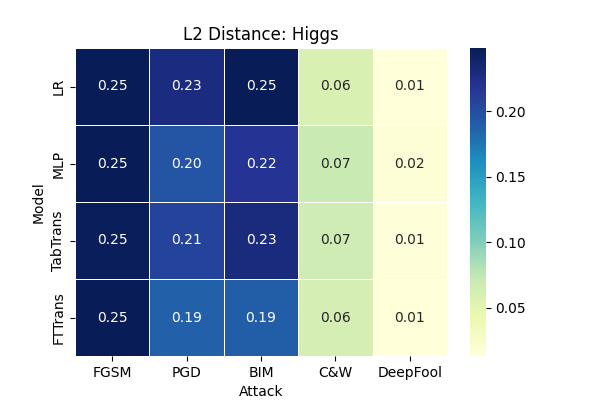}
    \caption{Higgs}
    \label{fig:pro:Higgs}
\end{subfigure}
\begin{subfigure}[t]{0.32\linewidth}
    \centering
    \includegraphics[trim={0.5cm 0 2cm 0},clip,width=\linewidth]{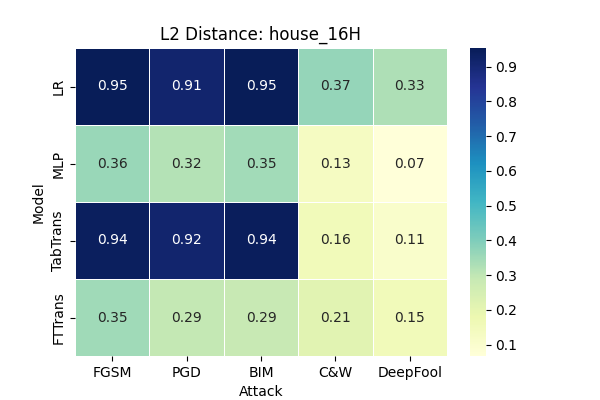}
    \caption{house\_16H}
    \label{fig:pro:house_16H}
\end{subfigure}
\begin{subfigure}[t]{0.32\linewidth}
    \centering
    \includegraphics[trim={0.5cm 0 2cm 0},clip,width=\linewidth]{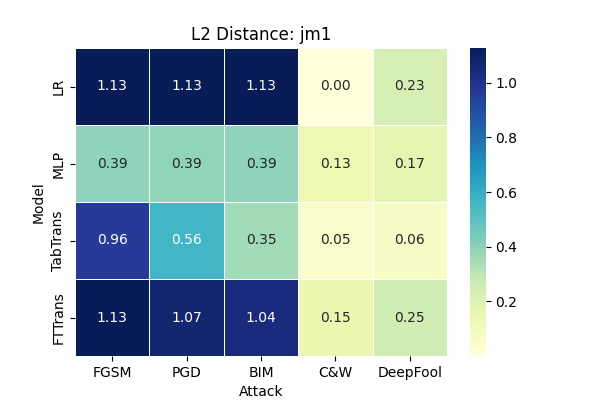}
    \caption{jm1}
    \label{fig:pro:jm1}
\end{subfigure}
\begin{subfigure}[t]{0.32\linewidth}
    \centering
    \includegraphics[trim={0.5cm 0 2cm 0},clip,width=\linewidth]{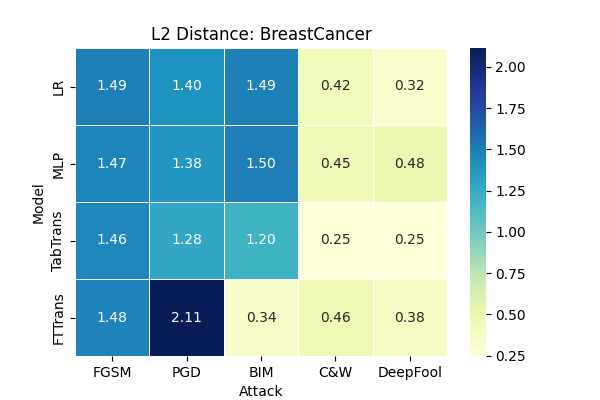}
    \caption{BreastCancer}
    \label{fig:pro:BreastCancer}
\end{subfigure}
\begin{subfigure}[t]{0.32\linewidth}
    \centering
    \includegraphics[trim={0.5cm 0 2cm 0},clip,width=\linewidth]{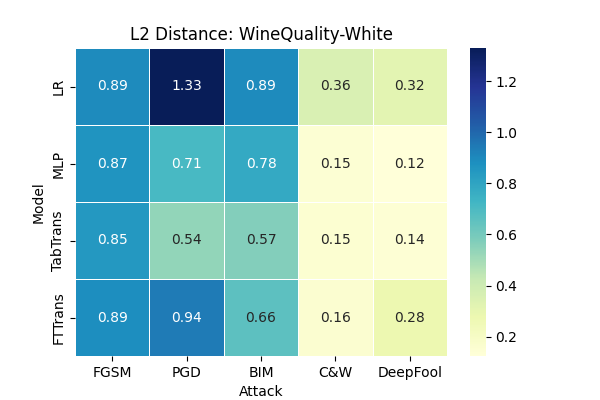}
    \caption{WineQuality-White}
    \label{fig:pro:WineQuality-White}
\end{subfigure}
\begin{subfigure}[t]{0.32\linewidth}
    \centering
    \includegraphics[trim={0.5cm 0 2cm 0},clip,width=\linewidth]{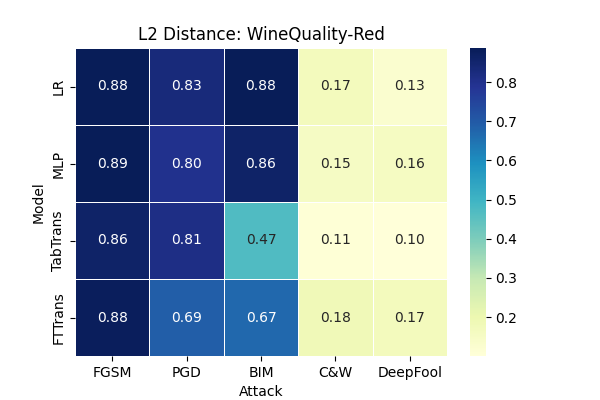}
    \caption{WineQuality-Red}
    \label{fig:pro:WineQuality-Red}
\end{subfigure}

\hspace{0.08\linewidth}
\begin{subfigure}[t]{0.32\linewidth}
    \centering
    \includegraphics[trim={0.5cm 0 2cm 0},clip,width=\linewidth]{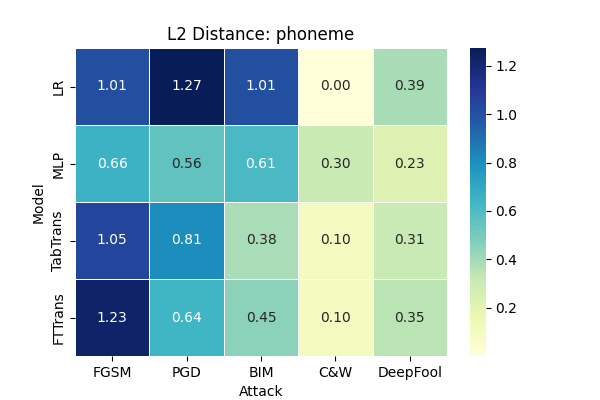}
    \caption{phoneme}
    \label{fig:pro:phoneme}
\end{subfigure}
\hspace{0.12\linewidth}
\begin{subfigure}[t]{0.32\linewidth}
    \centering
    \includegraphics[trim={0.5cm 0 2cm 0},clip,width=\linewidth]{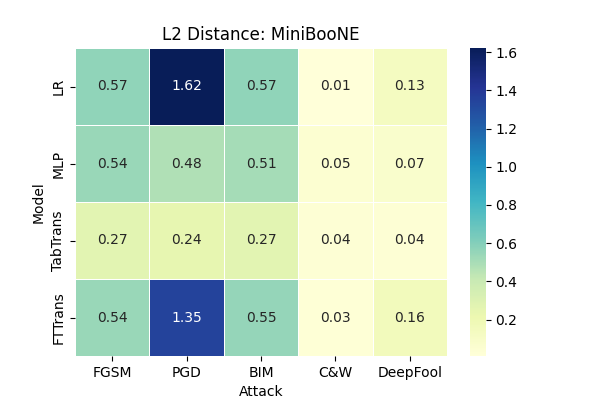}
    \caption{MiniBooNE}
    \label{fig:pro:MiniBooNE}
\end{subfigure}
    \caption{Proximity results (\emph{$\ell_2$ distance}) of evaluated attack methods and four models on eight \emph{numerical} datasets. Overall, $\ell_2$-based attacks (C\&W, DeepFool) yield much closer adversarial examples, while $\ell_\infty$-based methods (FGSM, BIM, PGD) often generate more distant perturbations. Architectural effects are evident, with LR generally requiring larger perturbations than MLP and transformer models.}
    \label{fig:pro:num_1}
\end{figure}

Our proximity analysis measures how close adversarial examples remain to their original samples in the feature space using \(\ell_2\) distance metrics. The heatmaps in Figures \ref{fig:pro:mixed} and \ref{fig:pro:num_1} reveal distinct patterns across attack types, model architectures, and datasets that provide important insights into the imperceptibility of different adversarial approaches.

\paragraph{Proximity: Mixed Dataset}

The mixed datasets (\emph{Adult}, \emph{Electricity}, and \emph{COMPAS}) reveal a consistent divide between attack types: 
\(\ell_2\)-based methods (C\&W, DeepFool) generate adversarial examples that remain much closer to the originals, while $\ell_\infty$-based methods (FGSM, BIM, PGD) produce substantially larger perturbations. Within this overall pattern, each dataset shows distinctive behaviour.

For the \emph{Adult} dataset (Figure \ref{fig:pro:Adult}), C\&W achieves the smallest distances (0.00--0.13), followed by DeepFool (0.30–0.41), whereas $\ell_\infty$-based methods range much higher (0.64–1.96). PGD stands out for producing the most distant adversarial examples, reaching 4.42 in LR and 2.25 in FTTransformer. This suggests that PGD's iterative optimisation, while effective, does so at the cost of proximity in this dataset.

The \emph{Electricity} dataset (Figure \ref{fig:pro:Elec}) exhibits smaller distances overall compared to \emph{Adult}, yet the $\ell_2$ vs $\ell_\infty$ gap persists. The difference is particularly pronounced in LR, where $\ell_\infty$ attacks average around 0.89, while  C\&W and DeepFool remain far closer (0.17 and 0.10, respectively). Neural models show minimal variation within each attack type, indicating that proximity here depends more on dataset structure than model architecture.

The \emph{COMPAS} dataset (Figure \ref{fig:pro:Compas}) continues the trend of $\ell_2$-based methods achieving the closest adversarial examples (0.00–0.07 for C\&W). However, PGD again produces extreme outcomes, particularly in FTTransformer, where distances reach the highest (3.14) across all mixed datasets. This highlights that certain dataset–model–attack combinations can yield disproportionately poor proximity, underscoring the variability of imperceptibility in mixed-feature tabular data.

\paragraph{Proximity: Numerical Dataset}

Numerical datasets generally exhibit smaller proximity distances than mixed datasets, with adversarial examples often remaining closer to the original inputs. The gap between $\ell_2$- and $\ell_\infty$-based methods is narrower here, though $\ell_2$ attacks (C\&W, DeepFool) still tend to preserve proximity better overall. 

The \emph{Higgs} dataset (Figure \ref{fig:pro:Higgs}) provides the clearest case of uniformly close adversarial examples. Across all models and attacks, $\ell_2$ distances remain very small (0.01--0.25), with $\ell_2$-based methods slightly closer (0.01--0.07) than $\ell_\infty$-based methods (0.19--0.25). This suggests that the feature space in \emph{Higgs} makes it easy to find nearby adversarial examples regardless of attack strategy.  

In contrast, the \emph{house\_16H} dataset (Figure \ref{fig:pro:house_16H}) shows pronounced variability across models. For LR and TabTransformer, $\ell_\infty$-based attacks yield much larger distances (0.91--0.95), while for MLP and FTTransformer, the same attacks produce considerably closer adversarial examples (0.29--0.36). Similarly, the \emph{jm1} dataset (Figure \ref{fig:pro:jm1}) highlights how optimisation choices within $\ell_\infty$ methods affect proximity. While distances are consistent for LR (1.13), TabTransformer shows wide variation: BIM finds relatively close examples (0.35) whereas FGSM produces much more distant ones (0.96). 

Some datasets exhibit particularly extreme proximity behaviour. 
In \emph{BreastCancer} (Figure \ref{fig:pro:BreastCancer}), $\ell_\infty$-based attacks generate some of the highest distances observed across all numerical datasets. For instance, PGD reaches 2.11 against FTTransformer, more than six times the distance of VIM (0.34) on the same model. The \emph{WineQuality} datasets (Figure \ref{fig:pro:WineQuality-White}, \ref{fig:pro:WineQuality-Red}) show moderate distances overall, but with dataset-specific patterns: in \emph{WineQuality-White}, PGD produces consistently higher values (up to 1.33 in LR) compared to FGSM and BIM, while this gap is less evident in \emph{WineQuality-Red}. 
The \emph{phoneme} dataset (Figure~\ref{fig:pro:phoneme}) reveals strong model dependence: PGD yields the most distant examples for LR (1.27), while FGSM does so for TabTransformer (1.05). In contrast, C\&W achieves exceptionally close adversarial examples across models, reaching zero for LR and just 0.10 for transformer models. Finally, the \emph{MiniBooNE} dataset (Figure \ref{fig:pro:MiniBooNE}) shows similar extremes, with PGD generating distances of 1.62 (LR) and 1.35 (FTTransformer), far above those produced by other methods.

From an architectural perspective, LR generally produces the largest $\ell_2$ distances, particularly under $\ell_\infty$ attacks, reflecting the need for larger perturbations to cross its linear decision boundaries. Transformer-based models exhibit more variable proximity patterns: in some cases, such as \emph{jm1}, they yield closer adversarial examples, while in others, such as \emph{BreastCancer}, they require substantially larger perturbations. MLP models typically fall between these two extremes, showing more stable values across datasets.

\begin{findingbox}[Overall insights for Task 2.2: Proximity]
Across both mixed and numerical datasets, $\ell_2$-based attacks (C\&W, DeepFool) consistently generate adversarial examples that remain much closer to the original inputs than $\ell_\infty$-based methods (FGSM, BIM, PGD). However, this proximity advantage often coincides with lower attack success, showing that imperceptibility cannot be interpreted in isolation from effectiveness.
\end{findingbox}

\subsubsection{Task 2.3: Deviation}

\begin{figure}[htb!]
\begin{subfigure}[t]{0.32\linewidth}
    \centering
    \includegraphics[trim={0.5cm 0 2cm 0},clip,width=\linewidth]{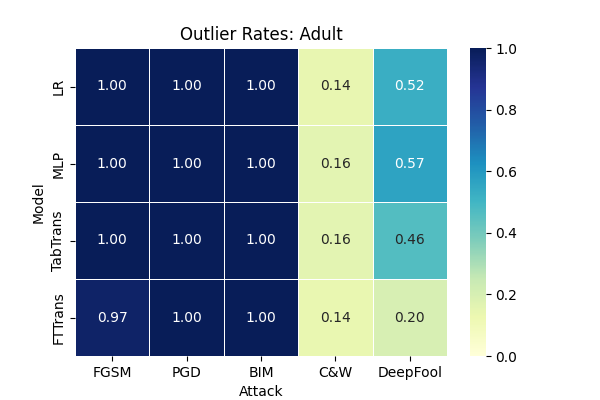}
    \caption{Adult}
    \label{fig:dev:Adult}
\end{subfigure}
\begin{subfigure}[t]{0.32\linewidth}
    \centering
    \includegraphics[trim={0.5cm 0 2cm 0},clip,width=\linewidth]{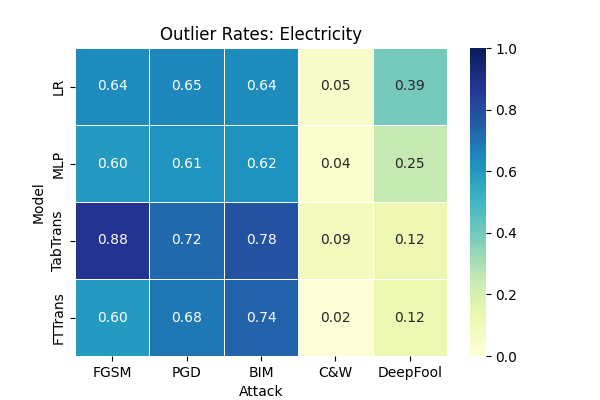}
    \caption{Electricity}
    \label{fig:dev:Elec}
\end{subfigure}
\begin{subfigure}[t]{0.32\linewidth}
    \centering
    \includegraphics[trim={0.5cm 0 2cm 0},clip,width=\linewidth]{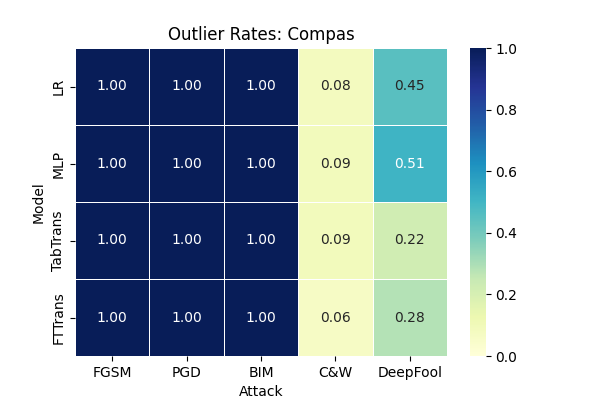}
    \caption{COMPAS}
    \label{fig:dev:Compas}
\end{subfigure}
    \caption{Deviation results (\emph{outlier rates}) of evaluated attack methods and four models on all three \emph{mixed} dataset. $\ell_\infty$-based attacks (FGSM, BIM, PGD) consistently push adversarial examples far outside the data distribution (outlier rates near 1.0), while $\ell_2$-based methods (C\&W, DeepFool) generally preserve distributional alignment.}
    \label{fig:dev:mixed}
\end{figure}

\begin{figure}[htb!]
\begin{subfigure}[t]{0.32\linewidth}
    \centering
    \includegraphics[trim={0.5cm 0 2cm 0},clip,width=\linewidth]{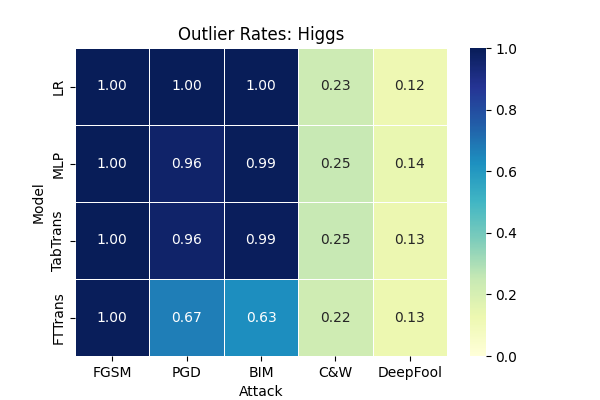}
    \caption{Higgs}
    \label{fig:dev:Higgs}
\end{subfigure}
\begin{subfigure}[t]{0.32\linewidth}
    \centering
    \includegraphics[trim={0.5cm 0 2cm 0},clip,width=\linewidth]{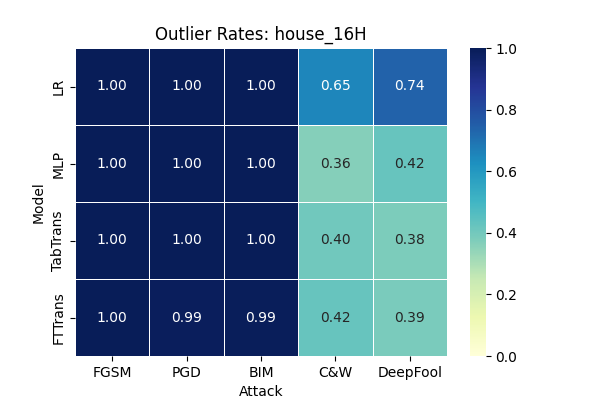}
    \caption{house\_16H}
    \label{fig:dev:house_16H}
\end{subfigure}
\begin{subfigure}[t]{0.32\linewidth}
    \centering
    \includegraphics[trim={0.5cm 0 2cm 0},clip,width=\linewidth]{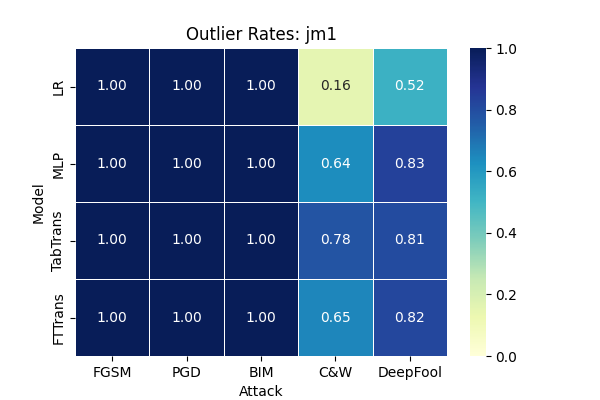}
    \caption{jm1}
    \label{fig:dev:jm1}
\end{subfigure}
\begin{subfigure}[t]{0.32\linewidth}
    \centering
    \includegraphics[trim={0.5cm 0 2cm 0},clip,width=\linewidth]{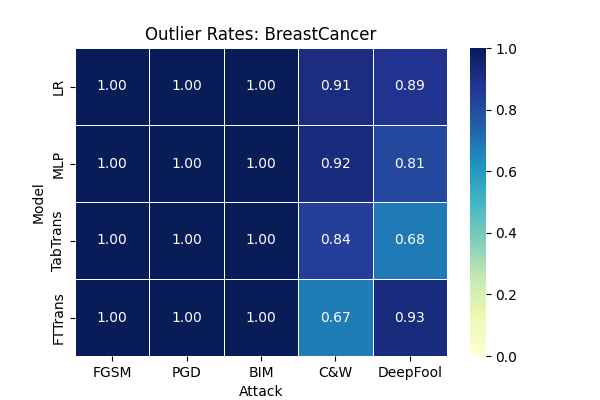}
    \caption{BreastCancer}
    \label{fig:dev:BreastCancer}
\end{subfigure}
\begin{subfigure}[t]{0.32\linewidth}
    \centering
    \includegraphics[trim={0.5cm 0 2cm 0},clip,width=\linewidth]{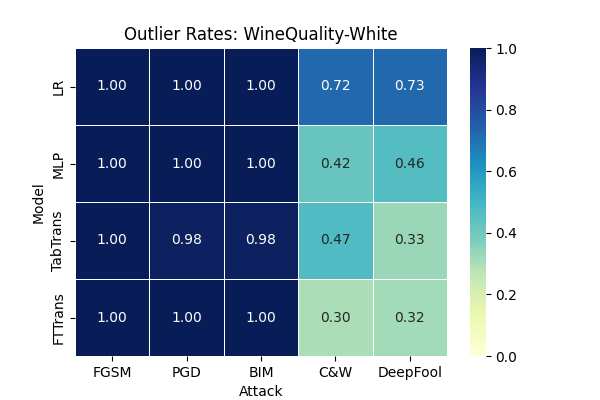}
    \caption{WineQuality-White}
    \label{fig:dev:WineQuality-White}
\end{subfigure}
\begin{subfigure}[t]{0.32\linewidth}
    \centering
    \includegraphics[trim={0.5cm 0 2cm 0},clip,width=\linewidth]{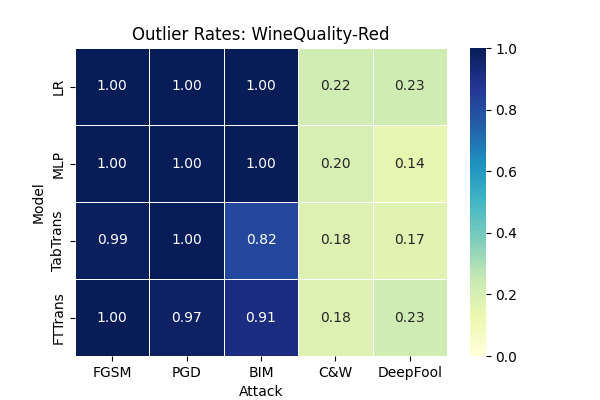}
    \caption{WineQuality-Red}
    \label{fig:dev:WineQuality-Red}
\end{subfigure}

\hspace{0.08\linewidth}
\begin{subfigure}[t]{0.32\linewidth}
    \centering
    \includegraphics[trim={0.5cm 0 2cm 0},clip,width=\linewidth]{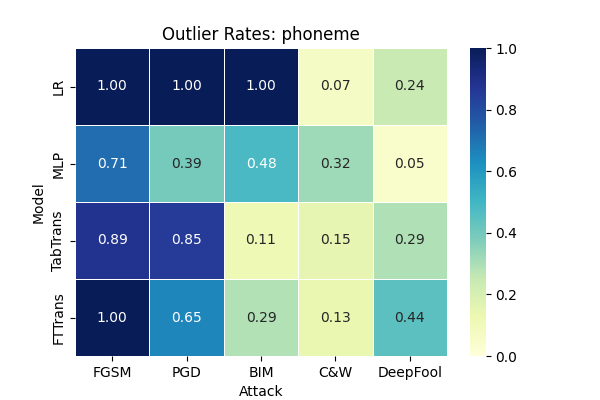}
    \caption{phoneme}
    \label{fig:dev:phoneme}
\end{subfigure}
\hspace{0.12\linewidth}
\begin{subfigure}[t]{0.32\linewidth}
    \centering
    \includegraphics[trim={0.5cm 0 2cm 0},clip,width=\linewidth]{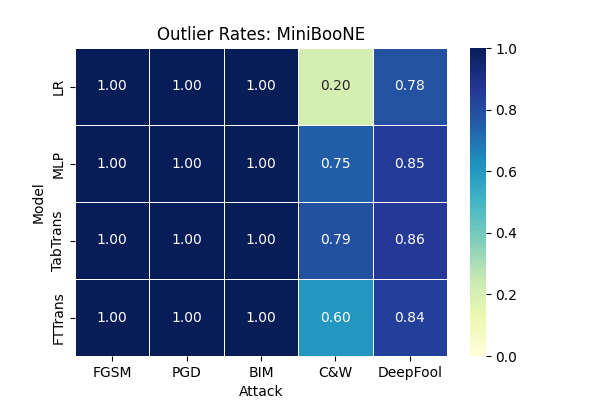}
    \caption{MiniBooNE}
    \label{fig:dev:MiniBooNE}
\end{subfigure}
    \caption{Deviation results (\emph{outlier rate}) of evaluated attack methods and four models on eight \emph{numerical} datasets. $\ell_\infty$ attacks (FGSM, BIM, PGD) generally push nearly all samples out-of-distribution, while $\ell_2$ methods (C\&W, DeepFool) preserve distributions more effectively but can still produce high deviation on compact datasets such as \emph{BreastCancer}.}
    \label{fig:dev:num_1}
\end{figure}

Our deviation analysis examines how significantly the adversarial examples differ from the original data distribution. The heatmaps presented in Figure \ref{fig:dev:mixed} - \ref{fig:dev:num_1} reveal clear patterns in the outlier rates produced by different attack algorithms across model architectures and datasets.

\paragraph{Deviation: Mixed Dataset}

Across three mixed datasets, a consistent divide emerges between attack types. $\ell_\infty$-based methods (FGSM, BIM, PGD) typically drive adversarial examples far outside the original data distribution, yielding very high outlier rates, whereas $\ell_2$-based methods (C\&W, DeepFool) preserve distributional alignment far more effectively.

The \emph{Adult} dataset (Figure \ref{fig:dev:Adult}) exemplifies this pattern: $\ell_\infty$-based attacks achieve near-perfect outlier rates (close to 1.00) across all models, while $\ell_2$-based methods remain much lower, with C\&W at 0.14--0.34 and DeepFool at 0.20--0.46. In \emph{Electricity} (Figure \ref{fig:dev:Elec}), the overall magnitude of outlier rates decreases, yet the divide persists. 
$\ell_\infty$-based attacks still dominate (0.60--0.88), with TabTransformer particularly vulnerable, while $\ell_2$-based methods again stay closer to the original distribution (0.02--0.39). The \emph{COMPAS} dataset (Figure \ref{fig:dev:Elec}) shows the starkest contrast: all $\ell_\infty$-based attacks reach a perfect outlier rate of 1.00 across models, 
while C\&W achieves consistently minimal rates (0.06--0.09) and DeepFool produces moderate ones (0.22--0.51).

\paragraph{Deviation: Numerical Dataset}

Across numerical datasets, a clear pattern emerges: $\ell_\infty$ attacks (FGSM, BIM, PGD) generally drive outlier rates close to 1.0, while $\ell_2$ methods (C\&W, DeepFool) tend to preserve distributional alignment more effectively, though with important dataset-specific variations.

The \emph{Higgs} dataset (Figure \ref{fig:dev:Higgs}) provides the clearest baseline: $\ell_\infty$ attacks yield near-perfect outlier rates, while $\ell_2$ attacks remain low (0.12--0.25). By contrast, both \emph{house\_16H} and \emph{jm1} (Figure \ref{fig:dev:house_16H}, Figure \ref{fig:dev:jm1}) show elevated outlier rates even for $\ell_2$ methods, reaching up to 0.65 for C\&W and above 0.80 for DeepFool. A similar trend is seen in \emph{MiniBooNE} (Figure \ref{fig:dev:MiniBooNE}), where adversarial examples from all methods frequently fall outside the data distribution, indicating that these feature spaces are particularly sensitive to perturbations regardless of norm constraints.

The \emph{BreastCancer} dataset (Figure \ref{fig:dev:BreastCancer}) is the strongest anomaly. 
Here, even small perturbations push samples out of distribution, with all methods, including  C\&W and DeepFool, producing high outlier rates (0.67–0.93). 
This suggests that the dataset's compact, tightly clustered structure leaves little room for imperceptible modification.

The \emph{WineQuality} datasets (Figure \ref{fig:dev:WineQuality-White}, \ref{fig:dev:WineQuality-Red}) show contrasting behaviours. White wine data align with the overall trend: almost all examples by  $\ell_\infty$ attacks are out-of-distribution (0.98--1.00), while $\ell_2$ methods yield lower but still non-trivial deviation (0.30--0.73). In contrast, the red wine dataset is less extreme and outlier rates are lower and more variable, with $\ell_\infty$ attacks occasionally dropping below 1.0 (e.g., BIM at 0.82 on TabTransformer). 
Finally, the \emph{phoneme} dataset (Figure \ref{fig:dev:phoneme}) illustrates the role of model architecture. While LR collapses under $\ell_\infty$ attacks (outlier rates of 1.0), TabTransformer shows remarkable resilience, with BIM reducing deviation to just 0.11. This highlights how complex architectures can sometimes accommodate perturbations without generating deviated adversarial examples.

\begin{findingbox}[Overall insights for Task 2.3: Deviation]
Our deviation analysis confirms that \(\ell_\infty\)-based attacks (FGSM, BIM, PGD) consistently generate out-of-distribution adversarial examples that significantly deviate from original data distributions, while \(\ell_2\)-based attacks (C\&W, DeepFool) tend to produce more in-distribution perturbations. However, the specific patterns vary notably by dataset characteristics and model architecture, highlighting the complex interplay between attack methods and the underlying data structures they attempt to exploit.
\end{findingbox}

\subsubsection{Task 2.4: Sensitivity}

\begin{figure}[!htb]
\begin{subfigure}[t]{0.32\linewidth}
    \centering
    \includegraphics[trim={0.5cm 0 2cm 0},clip,width=\linewidth]{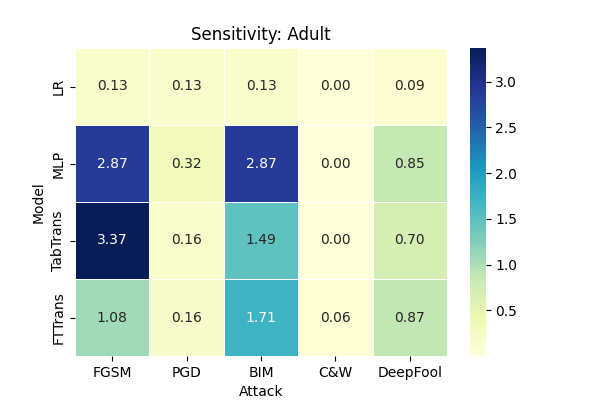}
    \caption{Adult}
    \label{fig:sen:Adult}
\end{subfigure}
\begin{subfigure}[t]{0.32\linewidth}
    \centering
    \includegraphics[trim={0.5cm 0 2cm 0},clip,width=\linewidth]{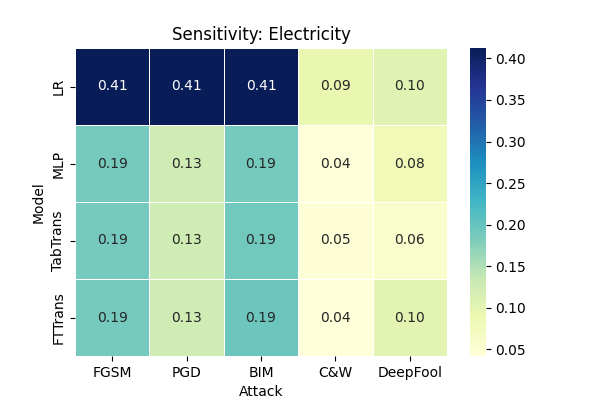}
    \caption{Electricity}
    \label{fig:sen:Elec}
\end{subfigure}
\begin{subfigure}[t]{0.32\linewidth}
    \centering
    \includegraphics[trim={0.5cm 0 2cm 0},clip,width=\linewidth]{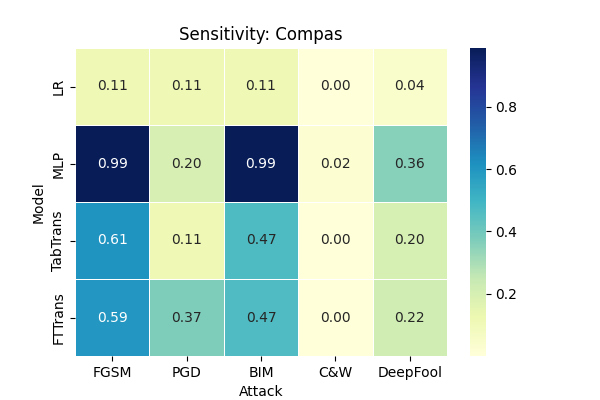}
    \caption{COMPAS}
    \label{fig:sen:Compas}
\end{subfigure}
    \caption{Sensitivity results (\emph{average sensitivity}) of evaluated attack methods and four models on all three \emph{mixed} dataset. $\ell_2$-based attacks (C\&W, DeepFool) consistently yield the lowest sensitivity, preserving narrow-distribution features, while $\ell_\infty$-based methods (FGSM, BIM, PGD) show higher variation across architectures and datasets.}
    \label{fig:sen:mixed}
\end{figure}

\begin{figure}[htb!]
\begin{subfigure}[t]{0.32\linewidth}
    \centering
    \includegraphics[trim={0.5cm 0 2cm 0},clip,width=\linewidth]{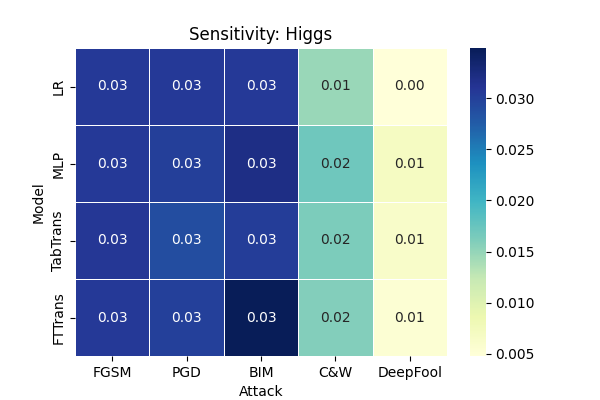}
    \caption{Higgs}
    \label{fig:sen:Higgs}
\end{subfigure}
\begin{subfigure}[t]{0.32\linewidth}
    \centering
    \includegraphics[trim={0.5cm 0 2cm 0},clip,width=\linewidth]{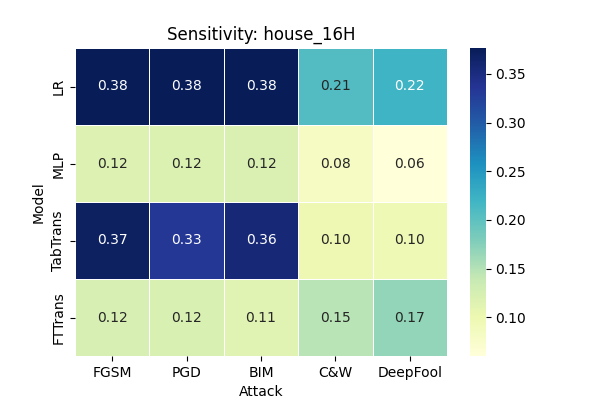}
    \caption{house\_16H}
    \label{fig:sen:house_16H}
\end{subfigure}
\begin{subfigure}[t]{0.32\linewidth}
    \centering
    \includegraphics[trim={0.5cm 0 2cm 0},clip,width=\linewidth]{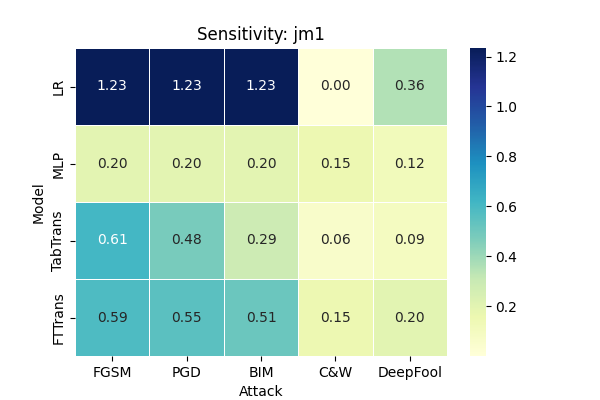}
    \caption{jm1}
    \label{fig:sen:jm1}
\end{subfigure}
\begin{subfigure}[t]{0.32\linewidth}
    \centering
    \includegraphics[trim={0.5cm 0 2cm 0},clip,width=\linewidth]{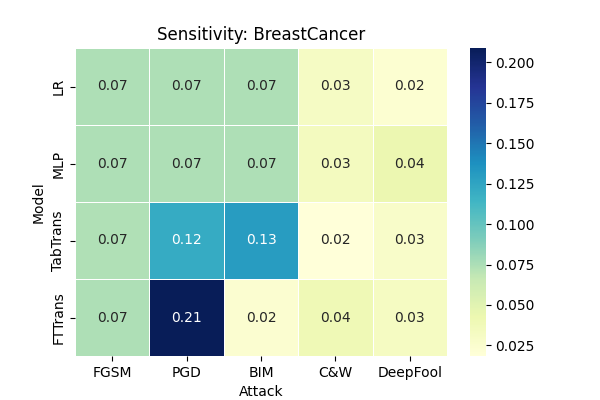}
    \caption{BreastCancer}
    \label{fig:sen:BreastCancer}
\end{subfigure}
\begin{subfigure}[t]{0.32\linewidth}
    \centering
    \includegraphics[trim={0.5cm 0 2cm 0},clip,width=\linewidth]{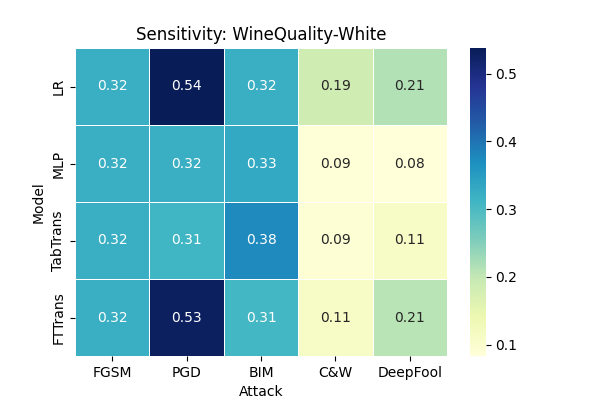}
    \caption{WineQuality-White}
    \label{fig:sen:WineQuality-White}
\end{subfigure}
\begin{subfigure}[t]{0.32\linewidth}
    \centering
    \includegraphics[trim={0.5cm 0 2cm 0},clip,width=\linewidth]{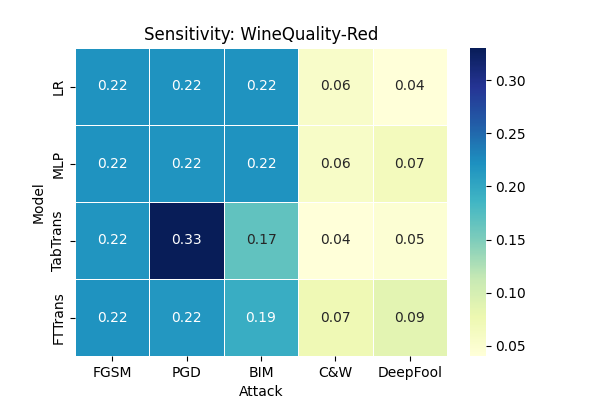}
    \caption{WineQuality-Red}
    \label{fig:sen:WineQuality-Red}
\end{subfigure}

\hspace{0.08\linewidth}
\begin{subfigure}[t]{0.32\linewidth}
    \centering
    \includegraphics[trim={0.5cm 0 2cm 0},clip,width=\linewidth]{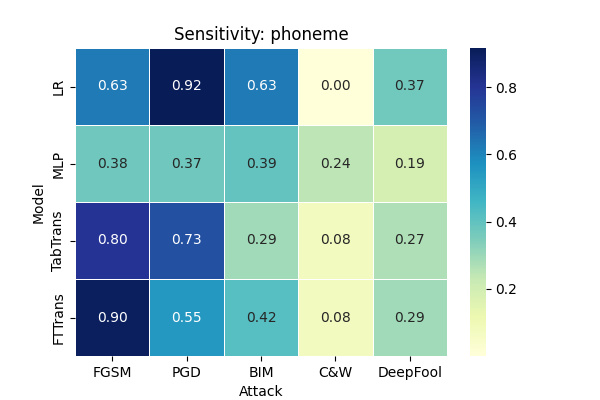}
    \caption{phoneme}
    \label{fig:sen:phoneme}
\end{subfigure}
\hspace{0.12\linewidth}
\begin{subfigure}[t]{0.32\linewidth}
    \centering
    \includegraphics[trim={0.5cm 0 2cm 0},clip,width=\linewidth]{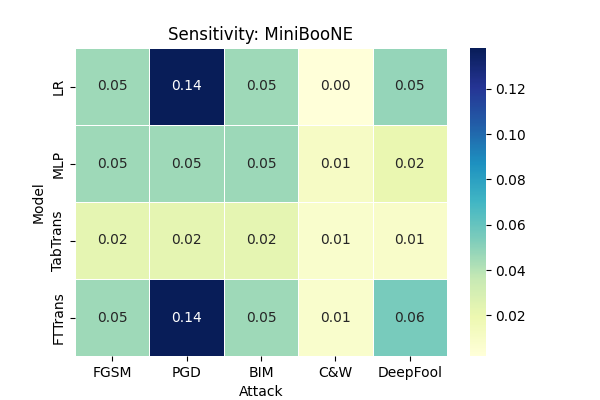}
    \caption{MiniBooNE}
    \label{fig:sen:MiniBooNE}
\end{subfigure}
    \caption{Sensitivity results (\emph{average sensitivity}) of evaluated attack methods and four models on eight \emph{numerical} datasets. Overall, $\ell_2$-based attacks (C\&W, DeepFool) again preserve narrow-distribution features most effectively, while $\ell_\infty$-based attacks (FGSM, BIM, PGD) show strong dataset- and model-dependent variability, with \emph{jm1} and \emph{phoneme} producing the highest sensitivities and \emph{Higgs} and \emph{MiniBooNE} the lowest.}
    \label{fig:sen:num_1}
\end{figure}

Our sensitivity analysis examines how adversarial attacks handle narrow-guard feature perturbation, particularly for features with narrow distributions in tabular data. The heatmaps in Figure \ref{fig:sen:mixed} and \ref{fig:sen:num_1} reveal complex patterns that vary significantly across datasets, attack algorithms, and model architectures. Rather than showing consistent behaviours, the sensitivity metrics highlight the contextual nature of how perturbations interact with narrowly distributed features.

\paragraph{Sensitivity: Mixed Dataset}

Across mixed datasets, a clear divide emerges between attack types. 
C\&W consistently yields the lowest sensitivity scores (0.00--0.06), reflecting its $\ell_2$ objective that discourages large deviations in any single feature. By contrast, $\ell_\infty$-based attacks generally induce higher sensitivity, with FGSM and BIM often producing the most extreme perturbations to narrow-distribution features, while PGD shows a somewhat more controlled pattern across models and datasets.

Model architecture further shapes these outcomes. LR models typically exhibit relatively low sensitivity, even under $\ell_\infty$ attacks (e.g., 0.13 in \emph{Adult}, 0.41 in \emph{Electricity}), suggesting that their linear boundaries limit perturbations to narrow-guard features. In contrast, MLP models can amplify sensitivity substantially, as seen in \emph{COMPAS} where FGSM and BIM reach 0.99. Transformer-based models usually fall between these extremes, but can occasionally produce very high spikes, such as the TabTransformer in \emph{Adult} where FGSM reaches 3.37 --- the highest sensitivity score observed across all mixed datasets.

Dataset-specific outcomes illustrate these broader trends. 
In \emph{Adult} (Figure \ref{fig:sen:Adult}), the gap between LR (0.13) and TabTransformer (3.37) under FGSM shows how architectural complexity can magnify perturbations on narrow features. 
The \emph{Electricity} dataset (Figure \ref{fig:sen:Elec}) displays uniformity across $\ell_\infty$ methods within each model: all three attacks yield identical scores (e.g., 0.41 in LR, 0.13--0.19 in neural models), indicating that the dataset's distribution constrains perturbation behaviour. 
In \emph{COMPAS} (Figure \ref{fig:sen:Compas}), MLP records much larger scores (0.99 for FGSM and BIM) than transformer models (0.47--0.61), while PGD consistently produces smaller values than the other $\ell_\infty$ methods, reflecting differences in optimisation strategy.

\paragraph{Sensitivity: Numerical Dataset}

Numerical datasets reveal diverse patterns of sensitivity, with some showing very limited perturbation of narrow-distribution features and others displaying pronounced vulnerability. The \emph{Higgs} (Figure \ref{fig:sen:Higgs}) and \emph{MiniBooNE} (Figure \ref{fig:sen:MiniBooNE}) datasets stand out for their uniformly low sensitivity scores across all attack–model combinations (typically 0.00--0.06). This suggests that either narrow-distribution features are absent or that adversarial examples can be generated without significantly altering them. 

In contrast, the \emph{jm1} dataset (Figure \ref{fig:sen:jm1}) produces substantially higher scores, particularly for LR under $\ell_\infty$ attacks (up to 1.23), indicating that narrowly distributed features strongly influence predictions in this dataset. More complex models, such as TabTransformer and FTTransformer, reduce these values (0.29--0.61), suggesting a lesser reliance on such features. The \emph{phoneme} dataset (Figure \ref{fig:sen:phoneme}) also shows high sensitivity, especially in transformer models targeted by FGSM and PGD, confirming that some datasets contain narrow features that adversaries can exploit.  

The WineQuality datasets (Figure \ref{fig:sen:WineQuality-White}, \ref{fig:sen:WineQuality-Red}) occupy a middle ground. In \emph{White}, PGD produces notably higher scores (0.31--0.54), particularly in LR and FTTransformer, whereas \emph{Red} exhibits more uniform behaviour, with $\ell_\infty$ attacks clustering together and $\ell_2$ methods consistently lower.  

From an architectural perspective, LR amplifies dataset effects: it shows either very high sensitivity (as in \emph{jm1}) or very low (as in \emph{Higgs}). Transformer-based models display greater variability, sometimes producing elevated scores (as in \emph{phoneme}) and sometimes low values (as in \emph{MiniBooNE}). MLP models generally lie between these extremes.  

Across attack methods, C\&W consistently yields the lowest sensitivity values, reflecting its $\ell_2$-based optimisation that avoids strong perturbation of individual features. PGD, by contrast, shows the widest variation, producing high scores in some datasets (e.g., \emph{WineQuality-White}, \emph{phoneme}) and more moderate values elsewhere, suggesting that its iterative procedure adapts strongly to dataset-specific structures.  

Overall, sensitivity to narrow-distribution feature perturbation is highly contextual, driven by the interplay of dataset characteristics, model reliance on specific features, and the perturbation strategy of the attack. 
These results caution against generalising attack behaviours and highlight the need for dataset-specific evaluation when assessing imperceptibility from a sensitivity perspective.

\begin{findingbox}[Overall insights for Task 2.4: Sensitivity]
The analysis of sensitivity shows that no attack universally preserves or disrupts narrow-distribution features; outcomes depend heavily on dataset structure and model reliance on these features. 
$\ell_2$-based methods (C\&W, DeepFool) generally avoid strong perturbations, while $\ell_\infty$-based methods (FGSM, BIM, PGD) can produce extreme values in certain contexts. 
These results highlight that imperceptibility in tabular data cannot be judged in isolation but must account for the interaction between feature distributions, attack design, and model architecture. 
\end{findingbox}

\subsection{Task 3: Trade-off Analysis}

\begin{figure}[t!]
    \centering
    \includegraphics[width=\linewidth]{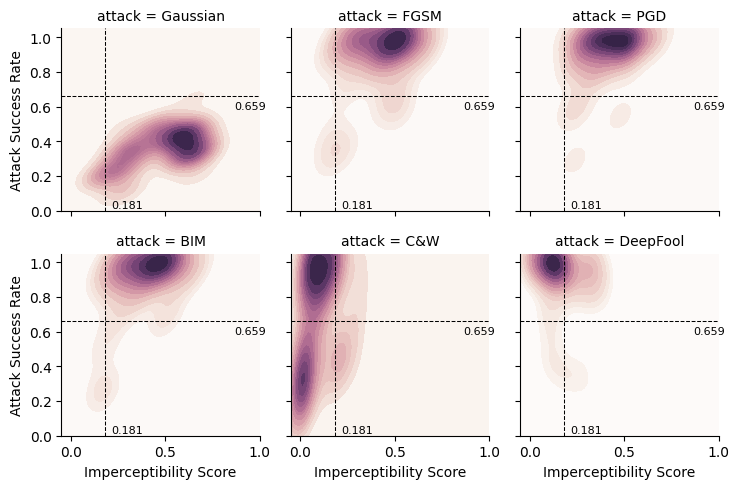}
    \caption{The 2D density plot shows the attack success rate (ASR) and imperceptibility score (IS) across Gaussian noise and five different attack methods. The plot is divided into four sectors based on the maximum ASR value (0.659) and the minimum IS value (0.181) observed among all adversarial examples generated by Gaussian noise. Gaussian noise is considered an ineffective and perceptible method for generating adversarial examples for tabular data. FGSM, PGD, and BIM are categorised as effective but perceptible methods. C\&W attack has two high-density regions: one that is Effective and Imperceptible, and another that is Ineffective but Imperceptible. Most of DeepFool attack's high-density regions fall into the Effective and Imperceptible sector.}
    \label{fig:asr-imp-2d}
\end{figure}

Analysing the relationship between attack success rate (ASR) and imperceptibility score (IS) provides critical insights into the relationship between effectiveness and imperceptibility of adversarial attacks on tabular data. By visualising this relationship through a 2D density plot in Figure \ref{fig:asr-imp-2d}, we can discern patterns that illuminate the interplay between these two crucial factors.
We take the Gaussian noise method as baseline and select the maximum ASR value (0.659) and the minimum IS value (0.181) from all adversarial examples generated by Gaussian noise. 
The graphs were divided into four distinct sections based on two thresholds, enabling us to categorise different scenarios and gain a clearer understanding of their impact. 

\paragraph{\textbf{Effective and Imperceptible} (High ASR, Low IS)}

The most desirable outcome for adversarial attacks occurs when examples successfully fool models while remaining nearly indistinguishable from original data. The density plot reveals that DeepFool consistently achieves co-optimal performance, with its highest density region falling in this quadrant. DeepFool's iterative approach of finding minimal perturbations to cross decision boundaries clearly excels at preserving tabular data characteristics while maintaining high effectiveness. 
C\&W also demonstrates strong performance in this quadrant for a portion of its examples, though it shows a bimodal distribution across both imperceptible regions. This suggests that C\&W can achieve the ideal results in many cases but may sometimes sacrifice effectiveness to maintain imperceptibility.


\paragraph{\textbf{Effective but Perceptible} (High ASR, High IS)}

This quadrant contains attacks that successfully mislead models but make noticeable modifications to the data. 
The density plots show that FGSM, PGD, and BIM consistently fall into this category, achieving high attack success rates at the cost of more significant data alterations. These \(\ell_\infty\)-based attacks effectively fool models but often modify features in ways that could compromise data integrity or be detected in quality control processes.

\paragraph{\textbf{Ineffective but Imperceptible} (Low ASR, Low IS)}

Attacks in this quadrant make subtle changes that preserve data characteristics but fail to successfully mislead models. C\&W shows a significant density in this region, indicating that it sometimes generates examples that maintain excellent imperceptibility but cannot effectively fool the model. This highlights C\&W's explicit optimisation for minimal perturbations, which can sometimes come at the expense of attack effectiveness.

\paragraph{\textbf{Ineffective and Perceptible} (Low ASR, High IS)}

The least desirable outcome occurs when attacks make noticeable changes yet fail to mislead the model. Gaussian noise predominantly falls in this category, confirming its poor performance as a baseline comparison. Its high-density region centres on moderate ASR values with high imperceptibility scores, demonstrating why random noise is considered both ineffective and easily perceptible.

\paragraph{Correlation Analysis}

To further investigate the trade-off between attack effectiveness and imperceptibility, 
we conducted a correlation analysis between Attack Success Rate (ASR) and five 
imperceptibility metrics across six attack methods. Table \ref{tab:correlation_by_attack} 
presents the Pearson correlation coefficients for each attack-metric pair. PGD, BIM, FGSM and C\&W demonstrate positive correlations with an average $r$ ranging from 0.171 to 0.273, indicating a clear trade-off where higher perturbation magnitudes are associated with higher success rates.  In contrast, DeepFool exhibits negative correlations across most metrics (avg $r = -0.235$), demonstrating efficient 
behaviour where it achieves high success rates with lower perturbation magnitudes (\(\ell_2\): $r = -0.536$, Sensitivity: $r = -0.427$). These findings suggest that the trade-off of effectiveness and imperceptibility is attack-dependent rather than universal (See more analysis in \ref{appendix:trade-off}).

\begin{table}[htbp!]
\centering
\caption{Correlation between Attack Success Rate and Imperceptibility Metrics by Attack Type. Note: Red values (positive correlation) indicate trade-off; Green values (negative correlation) indicate efficiency. Abbreviations: SpaR = Sparsity Rate, Sens = Sensitivity, OR = Outlier Rate, IS = Imperceptibility Score.}
\label{tab:correlation_by_attack}
\begin{tabular}{l|ccccc|c}
\hline
\textbf{Attack} & \textbf{$\ell_2$} & \textbf{SpaR} & \textbf{Sens} & \textbf{OR} & \textbf{IS} & \textbf{Avg} \\
\hline
Gaussian & -0.059 & 0.077 & \textcolor{red}{0.375} & \textcolor{red}{0.359} & \textcolor{red}{0.494} & \textbf{\textcolor{red}{0.249}} \\
FGSM & 0.220 & 0.229 & -0.283 & 0.278 & \textcolor{red}{0.410} & 0.171 \\
BIM & \textcolor{red}{0.344} & 0.296 & -0.269 & 0.218 & \textcolor{red}{0.472} & \textbf{\textcolor{red}{0.212}} \\
PGD & 0.081 & 0.252 & 0.212 & \textcolor{red}{0.428} & \textcolor{red}{0.390} & \textbf{\textcolor{red}{0.273}} \\
C\&W & \textcolor{red}{0.316} & \textcolor{red}{0.326} & 0.185 & 0.221 & 0.297 & \textbf{\textcolor{red}{0.269}} \\
DeepFool & \textcolor{green}{-0.536} & 0.206 & \textcolor{green}{-0.427} & -0.212 & -0.206 & \textbf{\textcolor{green}{-0.235}} \\
\hline
\end{tabular}
\end{table}

\paragraph{Overall Performance Comparison} 

The density plots and correlation analysis provide clear evidence for ranking the overall performance of different attack methods:
\begin{enumerate}
    \item  DeepFool emerges as the most favourable approach, consistently generating examples that are both highly effective and imperceptible. Its iterative linearisation of decision boundaries enables precise identification of minimal perturbations needed to cross classification boundaries, resulting in subtle modifications that maintain data integrity while achieving high ASR \citep{Moosavi2016deepfool}.
    \item C\&W shows mixed results with two distinct behaviour patterns --- one group achieving the simultaneous strength and another maintaining imperceptibility at the cost of effectiveness.
    \item The \(\ell_\infty\)-based attacks (FGSM, PGD, and BIM) prioritise effectiveness over imperceptibility, making them suitable for scenarios where attack success is more important than maintaining data characteristics.
    \item Gaussian noise serves as an appropriate baseline, demonstrating poor performance in both dimensions as expected.
\end{enumerate}

This analysis provides valuable guidance for selecting appropriate attack methods based on specific requirements for tabular data scenarios, highlighting the fundamental trade-off between effectiveness and imperceptibility in adversarial machine learning.

\section{Discussion}
\label{sec:discussion}


\subsection{Investigating the Inverse Relationship Between BIM Attack Budget and Success Rate}

As presented in Section~\ref{subsec:RQ1}, our \textbf{Task 1} evaluation results across both mixed and numerical datasets reveal an intriguing and counterintuitive phenomenon. While increasing epsilon \(\epsilon\) values generally leads to improved success rates for most attack methods, the BIM attack on the FTTransformer model shows a notable decline in success rates at higher perturbation budgets. This inverse relationship between attack budget and effectiveness contradicts conventional adversarial attack theory, where larger perturbation budgets typically enable more successful attacks.

The plots in Figure~\ref{fig:asr_by_epsilon_mixed} to \ref{fig:asr_by_epsilon_num_3} clearly demonstrate this unexpected pattern across multiple datasets, including \emph{Electricity}, \emph{COMPAS}, \emph{house\_16H}, \emph{BreastCancer}, and \emph{MiniBooNE}. In these cases, the attack success rate of BIM initially increase with epsilon values but then significantly decline at higher epsilon values, sometimes dropping dramatically. For example, on the \emph{BreastCancer} dataset, the success rate drops from approximately 35\% to nearly 0\% at the highest epsilon value, while on MiniBooNE, it plummets from 100\% to about 40\%.

\begin{figure}[t!]
    \centering
    \includegraphics[width=\linewidth]{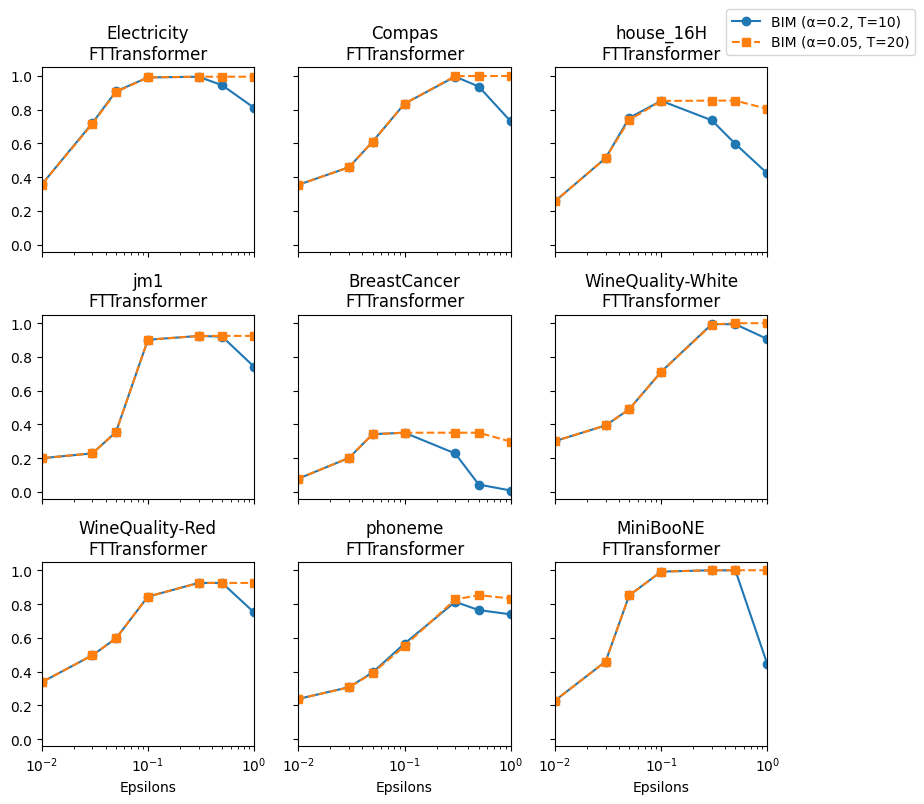}
    \caption{BIM attacks on FTTransformer stop dropping in attack success rate after adjusting step size (\(\alpha\)) hyperparameters. The orange lines, representing the adjusted BIM implementation, consistently maintain high attack success rates across all epsilon values, eliminating the dramatic drops observed with the default parameters. }
    \label{fig:bim_compare}
\end{figure}

The primary factor may explain this counterintuitive behaviour, which is gradient saturation effects. In BIM's iterative approach, the step size (\(\alpha\)) and number of iterations (\(T\)) play critical roles. When using default parameters (\(\alpha\)=0.2, \(T\)=10), the relatively large step size may cause overshooting at higher epsilon values. This occurs because BIM computes gradients with respect to the input and takes steps in that direction. As epsilon increases, these steps can become too large, causing the attack to miss optimal adversarial regions and produce less effective perturbations.

Our follow-up experiment demonstrates that adjusting BIM's hyperparameters can mitigate this issue. By reducing the step size (\(\alpha\)) from 0.2 to 0.05 and increasing iterations (\(T\)) from 10 to 20, in Figure \ref{fig:bim_compare}, we observe that the modified BIM attack maintains high success rates even at larger epsilon values across all datasets. This confirms that the original decline was primarily due to optimisation challenges rather than fundamental limitations of the attack method.

This finding has important implications for adversarial attack research on tabular data:
\begin{itemize}
    \item FTTransformer models possess unique adversarial robustness characteristics that differ from other model architectures.
    \item Attack hyperparameters require careful tuning based on both the model architecture and dataset characteristics.
\end{itemize}


\subsection{Exploring Design Strategies for Effective and Imperceptible Adversarial Attacks on Tabular Data}

In light of the results from analysing the relationship between attack success rate (ASR) and imperceptibility score (IS), achieving an optimal balance between effectiveness and imperceptibility is crucial in designing adversarial attack algorithms for tabular data. One notable observation is that $\ell_\infty$ attacks tend to generate highly effective adversarial examples, whereas $\ell_2$ attacks are more adept at producing imperceptible examples. The key challenge lies in finding the equilibrium between these two aspects.

To design effective and imperceptible adversarial attack algorithms for tabular data, several strategies can be explored:
\begin{itemize}


    \item  \textbf{Optimisation Techniques}: Employing advanced optimisation techniques can enhance the efficiency of adversarial attack algorithms. Techniques such as evolutionary algorithms, genetic algorithms, or gradient-based optimisation methods can be tailored to optimise both ASR and IS simultaneously, thereby facilitating the creation of more effective and imperceptible adversarial examples.
    \item \textbf{Feature Engineering}: Leveraging domain-specific knowledge and feature engineering techniques can enhance the robustness and imperceptibility of adversarial attacks. By identifying and manipulating key features within the tabular data that are most susceptible to manipulation, attackers can craft adversarial examples that achieve their objectives while minimising perceptible changes to the data.

\end{itemize}

\subsection{Evaluating the Suitability of One-Hot Encoding for Adversarial Attacks on Tabular Data}



Our sparsity evaluation in Section \ref{sec:evaluation}  revealed that categorical features are rarely perturbed under one-hot encoding. Since each category is represented by multiple binary indicators, altering a categorical attribute requires flipping at least two dimensions, which increases the perturbation cost. Consequently, optimisation algorithms tend to modify numerical features instead, resulting in dense perturbations within continuous attributes and limited diversity across feature types. 


\begin{table}[b!]
\centering
\caption{The encoding methods employed in recent papers on adversarial attacks targeting tabular data.}
\label{tab:encoding}
\begin{tabular}{@{}llll@{}}
\toprule
Paper                                                                                                                                      & Year & Encoding method               & Distance metric             \\
\midrule
\citet{ballet2019imperceptible}& 2019 & Drop all categorical features & $\ell_p$-norm \& Weighted $\ell_p$-norm \\
\citet{Mathov2022not} & 2021 & Label encoding                &    $\ell_2$-norm          \\ 
\citet{Chernikova2022FENCE} & 2022 & One-hot encoding              & $\ell_2$-norm                     \\ 
\citet{Cartella2021adversarial}  & 2021 & One-hot encoding              & $\ell_2$-norm                \\ 
\citet{kireev2022adversarial}  & 2022 & Discrete continuous features  & Cost function                 \\ 
\citet{zhou2022Discretization} & 2022 & Discrete continuous features  &  $\ell_1$-norm              \\
\bottomrule
\end{tabular}
\end{table}

Several existing studies have explored different strategies for handling categorical features for adversarial attacks on tabular data in Table~\ref{tab:encoding}. \citet{ballet2019imperceptible} proposed dropping all categorical features and using the $\ell_p$-norm and Weighted $\ell_p$-norm as distance metrics. \citet{Mathov2022not} suggested using label encoding for categorical data, though they did not specify the distance metric employed. \citet{Chernikova2022FENCE} and \citet{Cartella2021adversarial} both used one-hot encoding and applied the $\ell_2$-norm as their distance metric. On the other hand, \citet{kireev2022adversarial} and \citet{zhou2022Discretization} opted for discretising continuous features and used a cost function as a distance measure, without specifying a particular norm.



Moreover, exploring alternative distance metrics presents a promising direction for future research. Traditional metrics like the $\ell_p$-norm may not be well-suited for the mixed data types often found in tabular datasets. Metrics such as Gower's distance \citep{gower1971general}, which can handle mixed types of data (continuous, ordinal, and categorical), could provide a more accurate measure of similarity for tabular data. Additionally, other categorical feature similarity measures, such as those proposed by \citet{cost1993weighted} and \citet{le2005association}, offer potential improvements by considering the unique characteristics of categorical data. By integrating these distance metrics into the design of adversarial attack algorithms, researchers can develop more effective and nuanced methods that are better tailored to the complexities of tabular data.

\section{Conclusion}
\label{sec:conclusion}


In this paper, we conducted a comprehensive benchmark analysis of adversarial attacks on tabular data, focusing on both their \textit{effectiveness} and \textit{imperceptibility}. Using a diverse set of 11 datasets, encompassing both mixed and numerical data types, we evaluated the performance of five different adversarial attacks across four predictive models. The results reveal substantial variation in attack performance depending on model type and dataset characteristics, underscoring the unique challenges of adversarial robustness in structured data. 

Our benchmark demonstrates that highly effective attacks often compromise imperceptibility, while perturbations that remain realistic tend to achieve lower success rates. These findings quantify the trade-off between attack potency and subtlety, establishing a reproducible reference point for future work on tabular adversarial robustness. By providing unified evaluation metrics, standardised preprocessing, and open-source implementations, this study offers a consistent foundation for analysing vulnerabilities and improving defences in tabular machine learning systems.



\paragraph{Future Work}
While this benchmark establishes a strong foundation for evaluating adversarial robustness in tabular data, several directions remain open for future exploration:
\begin{itemize}
    \item \textbf{Incorporating defence mechanisms:} Extending the benchmark to include state-of-the-art adversarial defences (e.g., adversarial training, feature denoising, or certified robustness) will allow a more holistic assessment of both attack and defence strategies in practical deployments.
    \item \textbf{Expanding domain coverage:} Future iterations can include additional real-world domains such as cybersecurity, fraud detection, and medical diagnostics to further validate generalisability and cross-domain consistency.
    \item \textbf{Benchmarking black-box attacks:} Beyond white-box settings, future work will extend the benchmark to include black-box and adaptive attack scenarios~\citep{chen2017zoo,brendel2017decision,Cartella2021adversarial}. Such evaluations will enhance the practical relevance of the benchmark by reflecting real-world conditions where model internals are not fully observable.
    \item \textbf{Non-uniform feature importance:} This study assumes uniform feature contribution, yet real-world tabular data often contain features with heterogeneous influence. Investigating non-uniform or adaptive adversarial perturbations will provide deeper insight into model vulnerability under realistic conditions~\citep{erdemir2021adversarial,nandy2023non}.
\end{itemize}

\section*{Code Availability}

The implementation code, including data processing scripts and experimental pipelines, is openly available at \url{https://github.com/ZhipengHe/TabAttackBench/}

\section*{CRediT authorship contribution statement}


\textbf{Zhipeng He:} Conceptualisation, Methodology, Software, Investigation, Writing -- Original Draft \& Revision, Visualisation. 
\textbf{Chun Ouyang:} Conceptualisation, Methodology, Investigation, Writing -- Original Draft \& Revision. 
\textbf{Lijie Wen:} Methodology, Investigation. 
\textbf{Cong Liu:} Supervision. 
\textbf{Catarina Moreira:} Methodology, Investigation, Supervision.

\appendix

\section{Selected Attack Budgets (\(\epsilon\)) by ASR}\label{appendix:epsilons}

{\scriptsize
\begin{longtable}[c]{@{}llrrrrrr@{}}
\caption{Best attack budget (\(\epsilon\)) settings for four models on different datasets from the evaluation.}
\label{tab:my-table}\\

\toprule
\textbf{Datasets} & \textbf{Model} & \textbf{Guassian} & \textbf{FGSM} & \textbf{PGD} & \textbf{BIM} & \textbf{C\&W} & \textbf{DeepFool} \\* \midrule
\endfirsthead
\caption{Best attack budget (\(\epsilon\)) settings for four models on different datasets from the evaluation.}\\
\toprule
\textbf{Datasets} & \textbf{Model} & \textbf{Guassian} & \textbf{FGSM} & \textbf{PGD} & \textbf{BIM} & \textbf{C\&W} & \textbf{DeepFool} \\* \midrule
\endhead
\midrule
\multicolumn{8}{r}{{Continued on next page}} \\
\endfoot
\bottomrule
\endlastfoot
Adult & LR & 1 & 0.3 & 0.3 & 0.3 & 0.5 & 1 \\
Adult & MLP & 0.5 & 1 & 1 & 1 & 0.01 & 1 \\
Adult & TabTrans & 0.5 & 1 & 0.5 & 0.5 & 0.01 & 1 \\
Adult & FTTrans & 1 & 0.3 & 0.5 & 0.5 & 0.5 & 1 \\
Electricity & LR & 1 & 0.3 & 0.3 & 0.3 & 1 & 0.5 \\
Electricity & MLP & 1 & 0.1 & 0.1 & 0.1 & 0.3 & 0.3 \\
Electricity & TabTrans & 1 & 0.1 & 0.1 & 0.1 & 0.3 & 0.3 \\
Electricity & FTTrans & 1 & 0.1 & 0.1 & 0.1 & 0.3 & 0.3 \\
COMPAS & LR & 0.5 & 0.3 & 0.3 & 0.3 & 0.1 & 1 \\
COMPAS & MLP & 0.3 & 0.5 & 0.5 & 0.5 & 1 & 1 \\
COMPAS & TabTrans & 1 & 0.3 & 0.3 & 0.3 & 0.1 & 0.5 \\
COMPAS & FTTrans & 0.07 & 0.3 & 1 & 0.3 & 0.01 & 0.5 \\
Higgs & LR & 1 & 0.07 & 0.07 & 0.07 & 0.3 & 0.1 \\
Higgs & MLP & 1 & 0.07 & 0.07 & 0.07 & 0.3 & 0.1 \\
Higgs & TabTrans & 1 & 0.07 & 0.07 & 0.07 & 0.3 & 0.1 \\
Higgs & FTTrans & 1 & 0.07 & 0.07 & 0.07 & 0.3 & 0.1 \\
house\_16H & LR & 0.01 & 0.3 & 0.3 & 0.3 & 1 & 1 \\
house\_16H & MLP & 1 & 0.1 & 0.1 & 0.1 & 0.5 & 0.3 \\
house\_16H & TabTrans & 1 & 0.3 & 0.3 & 0.3 & 0.5 & 0.5 \\
house\_16H & FTTrans & 1 & 0.1 & 0.1 & 0.1 & 1 & 0.5 \\
jm1 & LR & 0.01 & 0.5 & 0.5 & 0.5 & 0.01 & 1 \\
jm1 & MLP & 1 & 0.1 & 0.1 & 0.1 & 0.5 & 0.5 \\
jm1 & TabTrans & 1 & 0.3 & 0.3 & 0.3 & 0.3 & 0.5 \\
jm1 & FTTrans & 1 & 0.3 & 0.3 & 0.3 & 0.5 & 0.5 \\
BreastCancer & LR & 1 & 0.3 & 0.3 & 0.3 & 1 & 1 \\
BreastCancer & MLP & 1 & 0.3 & 0.3 & 0.3 & 1 & 1 \\
BreastCancer & TabTrans & 1 & 0.3 & 0.5 & 0.5 & 1 & 1 \\
BreastCancer & FTTrans & 1 & 0.3 & 1 & 0.1 & 1 & 0.5 \\
Wine-White & LR & 1 & 0.3 & 0.5 & 0.3 & 1 & 1 \\
Wine-White & MLP & 1 & 0.3 & 0.3 & 0.3 & 0.5 & 0.5 \\
Wine-White & TabTrans & 1 & 0.3 & 0.3 & 0.3 & 0.5 & 0.5 \\
Wine-White & FTTrans & 1 & 0.3 & 0.5 & 0.3 & 1 & 1 \\
Wine-Red & LR & 1 & 0.3 & 0.3 & 0.3 & 1 & 0.5 \\
Wine-Red & MLP & 1 & 0.3 & 0.3 & 0.3 & 0.5 & 0.5 \\
Wine-Red & TabTrans & 1 & 0.3 & 0.5 & 0.3 & 0.5 & 0.5 \\
Wine-Red & FTTrans & 1 & 0.3 & 0.3 & 0.3 & 1 & 0.5 \\
phoneme & LR & 1 & 0.5 & 1 & 0.5 & 0.01 & 1 \\
phoneme & MLP & 1 & 0.3 & 0.3 & 0.3 & 1 & 1 \\
phoneme & TabTrans & 1 & 1 & 1 & 0.3 & 1 & 1 \\
phoneme & FTTrans & 1 & 1 & 0.5 & 0.3 & 0.5 & 1 \\
MiniBooNE & LR & 1 & 0.1 & 0.3 & 0.1 & 0.1 & 0.3 \\
MiniBooNE & MLP & 1 & 0.1 & 0.1 & 0.1 & 0.1 & 0.3 \\
MiniBooNE & TabTrans & 1 & 0.07 & 0.07 & 0.07 & 0.1 & 0.3 \\
MiniBooNE & FTTrans & 1 & 0.1 & 0.3 & 0.1 & 0.1 & 0.5 \\* \bottomrule

\end{longtable}
}

\section{Full Experimental Results}\label{appendix:full}

{\scriptsize
\begin{longtable}{lllrrrrrrrrr}
\caption{Full experimental results from all adversarial attack configurations.  ${\ell_2}$* and \textbf{SEN*} refer to normalised value.}
\label{tab:full_results}\\
\toprule
\textbf{Dataset} & \textbf{Model} & \textbf{Attack} & $\epsilon$ & \textbf{ASR} & \textbf{SpaR} & $\ell_2$ & ${\ell_2}$* & \textbf{SEN} & \textbf{SEN*} & \textbf{OR} & \textbf{IS} \\
\midrule
\endfirsthead

\caption[]{Full experimental results from all adversarial attack configurations.} \\
\toprule
\textbf{Dataset} & \textbf{Model} & \textbf{Attack} & $\epsilon$ & \textbf{ASR} & \textbf{SpaR} & $\ell_2$ & ${\ell_2}$* & \textbf{SEN} & \textbf{SEN*} & \textbf{OR} & \textbf{IS} \\
\midrule
\endhead

\midrule
\multicolumn{12}{r}{{Continued on next page}} \\

\endfoot

\bottomrule
\endlastfoot

Adult             & LR & Gaussian      & 1  & 0.1962              & 0.5216 & 0.7160      & 0.3660        & 0.0796      & 0.0740          & 0.2011        & 0.0436 \\
Adult             & LR & FGSM        & 0.3  & 0.9997              & 0.4289 & 1.9565      & 1.0000        & 0.1324      & 0.1230          & 0.9986        & 0.0807 \\
Adult             & LR & PGD         & 0.3  & 0.9997              & 0.4289 & 1.9460      & 0.9946        & 0.1324      & 0.1230          & 0.9980        & 0.0807 \\
Adult             & LR & BIM         & 0.3  & 0.9997              & 0.4289 & 1.9565      & 1.0000        & 0.1324      & 0.1230          & 0.9986        & 0.0807 \\
Adult             & LR & C\&W & 1  & 0.1999              & 0.9976 & 0.0510      & 0.0261        & 0.0058      & 0.0054          & 0.1598        & 0.0050 \\
Adult             & LR & DeepFool      & 1  & 0.8334              & 0.3272 & 0.4113      & 0.2102        & 0.0894      & 0.0830          & 0.5236        & 0.0463 \\
Adult             & MLP                & Gaussian      & 1  & 0.2168              & 0.5229 & 0.7205      & 0.3683        & 0.0800      & 0.0743          & 0.2302        & 0.0450 \\
Adult             & MLP                & FGSM        & 0.3  & 0.2985              & 0.0496 & 0.6323      & 0.3232        & 1.0415      & 0.9674          & 0.7269        & 0.0396 \\
Adult             & MLP                & PGD         & 0.3  & 0.2906              & 0.5177 & 1.3321      & 0.6809        & 0.0973      & 0.0904          & 0.3941        & 0.0593 \\
Adult             & MLP                & BIM         & 0.3  & 0.2985              & 0.0496 & 0.6323      & 0.3232        & 1.0415      & 0.9674          & 0.7269        & 0.0396 \\
Adult             & MLP                & C\&W & 1  & 0.2412              & 0.9975 & 0.0000      & 0.0000        & 0.0000      & 0.0000          & 0.1598        & 0.0005 \\
Adult             & MLP                & DeepFool      & 1  & 0.4141              & 0.0339 & 0.3624      & 0.1852        & 0.8500      & 0.7895          & 0.5651        & 0.0270 \\
Adult             & TabTrans     & Gaussian      & 1  & 0.2143              & 0.5228 & 0.7169      & 0.3664        & 0.0796      & 0.0739          & 0.2335        & 0.0450 \\
Adult             & TabTrans     & FGSM        & 0.3  & 0.4307              & 0.0491 & 0.6303      & 0.3222        & 1.0413      & 0.9672          & 0.9964        & 0.0398 \\
Adult             & TabTrans     & PGD         & 0.3  & 0.6069              & 0.5298 & 1.3461      & 0.6880        & 0.0951      & 0.0883          & 0.9992        & 0.0644 \\
Adult             & TabTrans     & BIM         & 0.3  & 0.5025              & 0.0544 & 0.5985      & 0.3059        & 0.9274      & 0.8614          & 0.9994        & 0.0426 \\
Adult             & TabTrans     & C\&W & 1  & 0.2417              & 0.9975 & 0.0009      & 0.0005        & 0.0001      & 0.0001          & 0.1614        & 0.0006 \\
Adult             & TabTrans     & DeepFool      & 1  & 0.5765              & 0.0364 & 0.3054      & 0.1561        & 0.6964      & 0.6469          & 0.4621        & 0.0272 \\
Adult             & FTTrans      & Gaussian      & 1  & 0.2220              & 0.5210 & 0.7190      & 0.3675        & 0.0804      & 0.0746          & 0.2040        & 0.0440 \\
Adult             & FTTrans      & FGSM        & 0.3  & 0.8967              & 0.0469 & 0.6353      & 0.3247        & 1.0766      & 1.0000          & 0.9663        & 0.0385 \\
Adult             & FTTrans      & PGD         & 0.3  & 0.8687              & 0.5234 & 1.3617      & 0.6960        & 0.0963      & 0.0895          & 0.9408        & 0.0647 \\
Adult             & FTTrans      & BIM         & 0.3  & 0.9154              & 0.0495 & 0.6310      & 0.3225        & 1.0201      & 0.9475          & 0.9764        & 0.0400 \\
Adult             & FTTrans      & C\&W & 1  & 0.3057              & 0.8866 & 0.1345      & 0.0687        & 0.0588      & 0.0546          & 0.1386        & 0.0246 \\
Adult             & FTTrans      & DeepFool      & 1  & 0.9754              & 0.0370 & 0.3029      & 0.1548        & 0.8710      & 0.8091          & 0.1986        & 0.0256 \\
Electricity       & LR & Gaussian      & 1  & 0.4116              & 0.7505 & 0.7795      & 0.3984        & 0.6045      & 0.5615          & 0.9332        & 0.1496 \\
Electricity       & LR & FGSM        & 0.3  & 0.9997              & 0.7224 & 0.8874      & 0.4536        & 0.4125      & 0.3831          & 0.6398        & 0.1291 \\
Electricity       & LR & PGD         & 0.3  & 0.9996              & 0.7224 & 0.8524      & 0.4357        & 0.4125      & 0.3832          & 0.6454        & 0.1278 \\
Electricity       & LR & BIM         & 0.3  & 0.9997              & 0.7224 & 0.8874      & 0.4536        & 0.4125      & 0.3831          & 0.6398        & 0.1291 \\
Electricity       & LR & C\&W & 1  & 0.9961              & 0.9911 & 0.1673      & 0.0855        & 0.0925      & 0.0859          & 0.0506        & 0.0230 \\
Electricity       & LR & DeepFool      & 1  & 1.0000              & 0.4342 & 0.0986      & 0.0504        & 0.1054      & 0.0979          & 0.3918        & 0.0291 \\
Electricity       & MLP                & Gaussian      & 1  & 0.4835              & 0.7465 & 0.7695      & 0.3933        & 0.6042      & 0.5612          & 0.9350        & 0.1488 \\
Electricity       & MLP                & FGSM        & 0.3  & 0.9960              & 0.4985 & 0.7107      & 0.3632        & 0.5645      & 0.5244          & 0.7767        & 0.1260 \\
Electricity       & MLP                & PGD         & 0.3  & 0.9972              & 0.7488 & 0.7146      & 0.3652        & 0.3812      & 0.3540          & 0.8317        & 0.1237 \\
Electricity       & MLP                & BIM         & 0.3  & 0.9977              & 0.4968 & 0.6977      & 0.3566        & 0.5669      & 0.5265          & 0.8937        & 0.1279 \\
Electricity       & MLP                & C\&W & 1  & 0.9947              & 0.9916 & 0.0732      & 0.0374        & 0.0506      & 0.0470          & 0.0407        & 0.0139 \\
Electricity       & MLP                & DeepFool      & 1  & 0.9938              & 0.3805 & 0.0487      & 0.0249        & 0.0838      & 0.0779          & 0.2522        & 0.0173 \\
Electricity       & TabTrans     & Gaussian      & 1  & 0.4433              & 0.7485 & 0.7713      & 0.3942        & 0.6023      & 0.5594          & 0.9266        & 0.1487 \\
Electricity       & TabTrans     & FGSM        & 0.3  & 0.9992              & 0.4984 & 0.7017      & 0.3587        & 0.5647      & 0.5245          & 1.0000        & 0.1301 \\
Electricity       & TabTrans     & PGD         & 0.3  & 0.9993              & 0.7520 & 0.6161      & 0.3149        & 0.3795      & 0.3525          & 1.0000        & 0.1202 \\
Electricity       & TabTrans     & BIM         & 0.3  & 0.9996              & 0.4632 & 0.5672      & 0.2899        & 0.6127      & 0.5691          & 1.0000        & 0.1198 \\
Electricity       & TabTrans     & C\&W & 1  & 0.9989              & 0.9939 & 0.0796      & 0.0407        & 0.0488      & 0.0454          & 0.0990        & 0.0177 \\
Electricity       & TabTrans     & DeepFool      & 1  & 0.9999              & 0.3804 & 0.0356      & 0.0182        & 0.0582      & 0.0541          & 0.1222        & 0.0123 \\
Electricity       & FTTrans      & Gaussian      & 1  & 0.4571              & 0.7491 & 0.7745      & 0.3958        & 0.5995      & 0.5569          & 0.9399        & 0.1491 \\
Electricity       & FTTrans      & FGSM        & 0.3  & 0.9861              & 0.4983 & 0.6976      & 0.3565        & 0.5649      & 0.5247          & 0.9549        & 0.1291 \\
Electricity       & FTTrans      & PGD         & 0.3  & 0.9015              & 0.7490 & 0.6441      & 0.3292        & 0.3800      & 0.3530          & 0.9547        & 0.1215 \\
Electricity       & FTTrans      & BIM         & 0.3  & 0.9943              & 0.4490 & 0.5273      & 0.2695        & 0.6107      & 0.5672          & 0.9649        & 0.1148 \\
Electricity       & FTTrans      & C\&W & 1  & 0.9849              & 0.9941 & 0.0674      & 0.0345        & 0.0455      & 0.0422          & 0.0166        & 0.0092 \\
Electricity       & FTTrans      & DeepFool      & 1  & 0.9949              & 0.3839 & 0.0885      & 0.0452        & 0.1510      & 0.1403          & 0.1350        & 0.0259 \\
COMPAS            & LR & Gaussian      & 1  & 0.3373              & 0.5432 & 0.7204      & 0.3682        & 0.1202      & 0.1117          & 0.7890        & 0.0681 \\
COMPAS            & LR & FGSM        & 0.3  & 1.0000              & 0.5645 & 1.6790      & 0.8582        & 0.1146      & 0.1064          & 1.0000        & 0.0755 \\
COMPAS            & LR & PGD         & 0.3  & 1.0000              & 0.5645 & 1.6686      & 0.8529        & 0.1146      & 0.1064          & 1.0000        & 0.0755 \\
COMPAS            & LR & BIM         & 0.3  & 1.0000              & 0.5645 & 1.6790      & 0.8582        & 0.1146      & 0.1064          & 1.0000        & 0.0755 \\
COMPAS            & LR & C\&W & 1  & 0.3545              & 0.9986 & 0.0017      & 0.0009        & 0.0002      & 0.0002          & 0.0822        & 0.0007 \\
COMPAS            & LR & DeepFool      & 1  & 0.9940              & 0.3953 & 0.2639      & 0.1349        & 0.0408      & 0.0379          & 0.4530        & 0.0264 \\
COMPAS            & MLP                & Gaussian      & 1  & 0.3491              & 0.5431 & 0.7180      & 0.3670        & 0.1194      & 0.1109          & 0.7866        & 0.0677 \\
COMPAS            & MLP                & FGSM        & 0.3  & 0.8414              & 0.1146 & 0.6900      & 0.3527        & 0.5851      & 0.5434          & 1.0000        & 0.0699 \\
COMPAS            & MLP                & PGD         & 0.3  & 0.9303              & 0.5463 & 1.0876      & 0.5559        & 0.1173      & 0.1090          & 1.0000        & 0.0729 \\
COMPAS            & MLP                & BIM         & 0.3  & 0.8414              & 0.1146 & 0.6900      & 0.3527        & 0.5851      & 0.5434          & 1.0000        & 0.0699 \\
COMPAS            & MLP                & C\&W & 1  & 0.4245              & 0.9444 & 0.0733      & 0.0375        & 0.0236      & 0.0219          & 0.0920        & 0.0123 \\
COMPAS            & MLP                & DeepFool      & 1  & 0.9940              & 0.0745 & 0.2722      & 0.1391        & 0.3579      & 0.3324          & 0.5056        & 0.0394 \\
COMPAS            & TabTrans     & Gaussian      & 1  & 0.3716              & 0.5415 & 0.7232      & 0.3696        & 0.1209      & 0.1123          & 0.7801        & 0.0683 \\
COMPAS            & TabTrans     & FGSM        & 0.3  & 0.9820              & 0.1060 & 0.6623      & 0.3385        & 0.6096      & 0.5662          & 0.9997        & 0.0664 \\
COMPAS            & TabTrans     & PGD         & 0.3  & 0.9988              & 0.5672 & 1.0631      & 0.5434        & 0.1121      & 0.1041          & 1.0000        & 0.0709 \\
COMPAS            & TabTrans     & BIM         & 0.3  & 0.9970              & 0.1316 & 0.6307      & 0.3223        & 0.4689      & 0.4356          & 1.0000        & 0.0718 \\
COMPAS            & TabTrans     & C\&W & 1  & 0.4263              & 0.9987 & 0.0074      & 0.0038        & 0.0010      & 0.0009          & 0.0909        & 0.0014 \\
COMPAS            & TabTrans     & DeepFool      & 1  & 0.9994              & 0.0726 & 0.1340      & 0.0685        & 0.2171      & 0.2017          & 0.2381        & 0.0270 \\
COMPAS            & FTTrans      & Gaussian      & 1  & 0.4776              & 0.5394 & 0.7173      & 0.3667        & 0.1198      & 0.1113          & 0.7736        & 0.0677 \\
COMPAS            & FTTrans      & FGSM        & 0.3  & 0.9985              & 0.1126 & 0.6855      & 0.3504        & 0.5950      & 0.5526          & 1.0000        & 0.0692 \\
COMPAS            & FTTrans      & PGD         & 0.3  & 0.7696              & 0.5499 & 1.0582      & 0.5409        & 0.1165      & 0.1082          & 0.9949        & 0.0723 \\
COMPAS            & FTTrans      & BIM         & 0.3  & 0.9949              & 0.1232 & 0.5849      & 0.2989        & 0.4662      & 0.4330          & 1.0000        & 0.0681 \\
COMPAS            & FTTrans      & C\&W & 1  & 0.5074              & 0.9982 & 0.0000      & 0.0000        & 0.0000      & 0.0000          & 0.0628        & 0.0005 \\
COMPAS            & FTTrans      & DeepFool      & 1  & 0.9991              & 0.0777 & 0.1651      & 0.0844        & 0.2515      & 0.2336          & 0.2913        & 0.0311 \\
Higgs             & LR & Gaussian      & 1  & 0.4752              & 0.9336 & 0.8177      & 0.4180        & 0.2425      & 0.2252          & 0.9998        & 0.1127 \\
Higgs             & LR & FGSM        & 0.3  & 1.0000              & 0.9362 & 1.2939      & 0.6613        & 0.1836      & 0.1706          & 1.0000        & 0.1063 \\
Higgs             & LR & PGD         & 0.3  & 1.0000              & 0.9362 & 1.2392      & 0.6334        & 0.1836      & 0.1706          & 1.0000        & 0.1056 \\
Higgs             & LR & BIM         & 0.3  & 1.0000              & 0.9362 & 1.2939      & 0.6613        & 0.1836      & 0.1706          & 1.0000        & 0.1063 \\
Higgs             & LR & C\&W & 1  & 1.0000              & 0.9857 & 0.0585      & 0.0299        & 0.0147      & 0.0137          & 0.2339        & 0.0095 \\
Higgs             & LR & DeepFool      & 1  & 1.0000              & 0.5986 & 0.0144      & 0.0073        & 0.0050      & 0.0046          & 0.1228        & 0.0032 \\
Higgs             & MLP                & Gaussian      & 1  & 0.4834              & 0.9349 & 0.8176      & 0.4179        & 0.2414      & 0.2242          & 1.0000        & 0.1124 \\
Higgs             & MLP                & FGSM        & 0.3  & 0.8697              & 0.9373 & 1.2987      & 0.6638        & 0.1833      & 0.1703          & 1.0000        & 0.1063 \\
Higgs             & MLP                & PGD         & 0.3  & 0.9864              & 0.9522 & 1.0700      & 0.5469        & 0.1803      & 0.1675          & 1.0000        & 0.1020 \\
Higgs             & MLP                & BIM         & 0.3  & 0.9999              & 0.8958 & 1.1412      & 0.5833        & 0.1928      & 0.1790          & 1.0000        & 0.1066 \\
Higgs             & MLP                & C\&W & 1  & 0.9998              & 0.9890 & 0.0703      & 0.0359        & 0.0171      & 0.0159          & 0.2470        & 0.0110 \\
Higgs             & MLP                & DeepFool      & 1  & 0.9996              & 0.6796 & 0.0188      & 0.0096        & 0.0077      & 0.0071          & 0.1499        & 0.0044 \\
Higgs             & TabTrans     & Gaussian      & 1  & 0.4995              & 0.9343 & 0.8193      & 0.4188        & 0.2426      & 0.2253          & 0.9998        & 0.1128 \\
Higgs             & TabTrans     & FGSM        & 0.3  & 0.9563              & 0.9278 & 1.2783      & 0.6534        & 0.1854      & 0.1723          & 1.0000        & 0.1067 \\
Higgs             & TabTrans     & PGD         & 0.3  & 0.9918              & 0.9976 & 1.1295      & 0.5773        & 0.1719      & 0.1597          & 1.0000        & 0.1005 \\
Higgs             & TabTrans     & BIM         & 0.3  & 0.9982              & 0.9363 & 1.2146      & 0.6208        & 0.1833      & 0.1702          & 1.0000        & 0.1051 \\
Higgs             & TabTrans     & C\&W & 1  & 0.9956              & 0.9880 & 0.0669      & 0.0342        & 0.0163      & 0.0151          & 0.2489        & 0.0105 \\
Higgs             & TabTrans     & DeepFool      & 1  & 0.9982              & 0.6394 & 0.0164      & 0.0084        & 0.0071      & 0.0066          & 0.1411        & 0.0040 \\
Higgs             & FTTrans      & Gaussian      & 1  & 0.4855              & 0.9340 & 0.8188      & 0.4185        & 0.2422      & 0.2250          & 0.9998        & 0.1126 \\
Higgs             & FTTrans      & FGSM        & 0.3  & 0.8775              & 0.9344 & 1.2902      & 0.6595        & 0.1840      & 0.1709          & 1.0000        & 0.1064 \\
Higgs             & FTTrans      & PGD         & 0.3  & 0.9919              & 0.9600 & 0.9245      & 0.4726        & 0.1788      & 0.1661          & 0.9995        & 0.0987 \\
Higgs             & FTTrans      & BIM         & 0.3  & 0.9955              & 0.8740 & 0.9650      & 0.4933        & 0.1973      & 0.1833          & 1.0000        & 0.1043 \\
Higgs             & FTTrans      & C\&W & 1  & 0.9988              & 0.9892 & 0.0625      & 0.0319        & 0.0160      & 0.0148          & 0.2232        & 0.0101 \\
Higgs             & FTTrans      & DeepFool      & 1  & 1.0000              & 0.6784 & 0.0157      & 0.0080        & 0.0065      & 0.0060          & 0.1384        & 0.0038 \\
house\_16H        & LR & Gaussian      & 1  & 0.2967              & 0.9626 & 0.7776      & 0.3975        & 0.5257      & 0.4883          & 1.0000        & 0.1517 \\
house\_16H        & LR & FGSM        & 0.3  & 0.9952              & 0.8966 & 0.9539      & 0.4876        & 0.3766      & 0.3498          & 1.0000        & 0.1427 \\
house\_16H        & LR & PGD         & 0.3  & 0.9903              & 0.8968 & 0.9142      & 0.4673        & 0.3766      & 0.3498          & 1.0000        & 0.1409 \\
house\_16H        & LR & BIM         & 0.3  & 0.9952              & 0.8966 & 0.9539      & 0.4876        & 0.3766      & 0.3498          & 1.0000        & 0.1427 \\
house\_16H        & LR & C\&W & 1  & 0.7369              & 0.9929 & 0.3681      & 0.1881        & 0.2080      & 0.1932          & 0.6498        & 0.0770 \\
house\_16H        & LR & DeepFool      & 1  & 0.9993              & 0.6018 & 0.3277      & 0.1675        & 0.2207      & 0.2050          & 0.7418        & 0.0725 \\
house\_16H        & MLP                & Gaussian      & 1  & 0.4742              & 0.9220 & 0.7923      & 0.4050        & 0.5526      & 0.5133          & 1.0000        & 0.1541 \\
house\_16H        & MLP                & FGSM        & 0.3  & 1.0000              & 0.9472 & 1.0094      & 0.5159        & 0.3517      & 0.3266          & 1.0000        & 0.1420 \\
house\_16H        & MLP                & PGD         & 0.3  & 1.0000              & 0.9299 & 0.8950      & 0.4574        & 0.3584      & 0.3329          & 1.0000        & 0.1379 \\
house\_16H        & MLP                & BIM         & 0.3  & 1.0000              & 0.9163 & 0.9432      & 0.4821        & 0.3641      & 0.3382          & 1.0000        & 0.1407 \\
house\_16H        & MLP                & C\&W & 1  & 0.9978              & 0.9964 & 0.1374      & 0.0702        & 0.0869      & 0.0807          & 0.3717        & 0.0334 \\
house\_16H        & MLP                & DeepFool      & 1  & 1.0000              & 0.8143 & 0.0675      & 0.0345        & 0.0607      & 0.0564          & 0.4237        & 0.0203 \\
house\_16H        & TabTrans     & Gaussian      & 1  & 0.3950              & 0.9279 & 0.7909      & 0.4043        & 0.5520      & 0.5127          & 1.0000        & 0.1541 \\
house\_16H        & TabTrans     & FGSM        & 0.3  & 0.9998              & 0.9068 & 0.9405      & 0.4807        & 0.3699      & 0.3435          & 1.0000        & 0.1413 \\
house\_16H        & TabTrans     & PGD         & 0.3  & 1.0000              & 0.9928 & 0.9157      & 0.4680        & 0.3337      & 0.3100          & 1.0000        & 0.1360 \\
house\_16H        & TabTrans     & BIM         & 0.3  & 1.0000              & 0.9345 & 0.9447      & 0.4828        & 0.3560      & 0.3307          & 1.0000        & 0.1399 \\
house\_16H        & TabTrans     & C\&W & 1  & 0.9976              & 0.9963 & 0.1642      & 0.0839        & 0.1022      & 0.0950          & 0.4158        & 0.0391 \\
house\_16H        & TabTrans     & DeepFool      & 1  & 1.0000              & 0.7428 & 0.1126      & 0.0575        & 0.0999      & 0.0928          & 0.3799        & 0.0315 \\
house\_16H        & FTTrans      & Gaussian      & 1  & 0.3704              & 0.9351 & 0.7878      & 0.4027        & 0.5445      & 0.5058          & 1.0000        & 0.1534 \\
house\_16H        & FTTrans      & FGSM        & 0.3  & 0.8479              & 0.9102 & 0.9585      & 0.4899        & 0.3699      & 0.3435          & 1.0000        & 0.1421 \\
house\_16H        & FTTrans      & PGD         & 0.3  & 0.8238              & 0.9304 & 0.7085      & 0.3621        & 0.3526      & 0.3275          & 1.0000        & 0.1271 \\
house\_16H        & FTTrans      & BIM         & 0.3  & 0.7364              & 0.8775 & 0.6852      & 0.3502        & 0.3466      & 0.3219          & 0.9997        & 0.1237 \\
house\_16H        & FTTrans      & C\&W & 1  & 0.9451              & 0.9812 & 0.2130      & 0.1089        & 0.1473      & 0.1368          & 0.4233        & 0.0507 \\
house\_16H        & FTTrans      & DeepFool      & 1  & 0.8466              & 0.7388 & 0.2312      & 0.1182        & 0.2510      & 0.2331          & 0.3901        & 0.0603 \\
jm1               & LR & Gaussian      & 1  & 0.1925              & 0.8952 & 0.7096      & 0.3627        & 0.8589      & 0.7978          & 1.0000        & 0.1635 \\
jm1               & LR & FGSM        & 0.3  & 0.9573              & 0.8155 & 0.6913      & 0.3533        & 0.7366      & 0.6842          & 1.0000        & 0.1537 \\
jm1               & LR & PGD         & 0.3  & 0.9499              & 0.8165 & 0.6906      & 0.3530        & 0.7362      & 0.6838          & 1.0000        & 0.1537 \\
jm1               & LR & BIM         & 0.3  & 0.9573              & 0.8155 & 0.6913      & 0.3533        & 0.7366      & 0.6842          & 1.0000        & 0.1537 \\
jm1               & LR & C\&W & 1  & 0.1925              & 0.9993 & 0.0000      & 0.0000        & 0.0000      & 0.0000          & 0.1551        & 0.0005 \\
jm1               & LR & DeepFool      & 1  & 0.3436              & 0.3140 & 0.2277      & 0.1164        & 0.3551      & 0.3299          & 0.5160        & 0.0601 \\
jm1               & MLP                & Gaussian      & 1  & 0.6587              & 0.8882 & 0.7300      & 0.3731        & 0.8799      & 0.8173          & 1.0000        & 0.1661 \\
jm1               & MLP                & FGSM        & 0.3  & 0.9995              & 0.9534 & 1.1506      & 0.5881        & 0.5868      & 0.5450          & 1.0000        & 0.1793 \\
jm1               & MLP                & PGD         & 0.3  & 0.9995              & 0.9534 & 1.1487      & 0.5871        & 0.5868      & 0.5450          & 1.0000        & 0.1793 \\
jm1               & MLP                & BIM         & 0.3  & 0.9995              & 0.9534 & 1.1505      & 0.5881        & 0.5868      & 0.5450          & 1.0000        & 0.1793 \\
jm1               & MLP                & C\&W & 1  & 0.4483              & 0.9999 & 0.1335      & 0.0682        & 0.1504      & 0.1397          & 0.6373        & 0.0415 \\
jm1               & MLP                & DeepFool      & 1  & 0.9995              & 0.7997 & 0.1679      & 0.0858        & 0.1195      & 0.1110          & 0.8336        & 0.0437 \\
jm1               & TabTrans     & Gaussian      & 1  & 0.2834              & 0.8876 & 0.6921      & 0.3538        & 0.8408      & 0.7810          & 1.0000        & 0.1607 \\
jm1               & TabTrans     & FGSM        & 0.3  & 1.0000              & 0.9004 & 0.9625      & 0.4920        & 0.6091      & 0.5658          & 1.0000        & 0.1695 \\
jm1               & TabTrans     & PGD         & 0.3  & 0.9890              & 0.9723 & 0.5614      & 0.2870        & 0.4785      & 0.4445          & 1.0000        & 0.1291 \\
jm1               & TabTrans     & BIM         & 0.3  & 1.0000              & 0.9096 & 0.3550      & 0.1814        & 0.2877      & 0.2673          & 1.0000        & 0.0884 \\
jm1               & TabTrans     & C\&W & 1  & 0.7327              & 0.9999 & 0.0497      & 0.0254        & 0.0600      & 0.0558          & 0.7781        & 0.0173 \\
jm1               & TabTrans     & DeepFool      & 1  & 1.0000              & 0.7323 & 0.0648      & 0.0331        & 0.0943      & 0.0876          & 0.8071        & 0.0231 \\
jm1               & FTTrans      & Gaussian      & 1  & 0.4534              & 0.8971 & 0.7328      & 0.3745        & 0.8681      & 0.8063          & 1.0000        & 0.1663 \\
jm1               & FTTrans      & FGSM        & 0.3  & 0.9196              & 0.9467 & 1.1269      & 0.5760        & 0.5855      & 0.5439          & 1.0000        & 0.1779 \\
jm1               & FTTrans      & PGD         & 0.3  & 0.9977              & 0.9607 & 1.0661      & 0.5449        & 0.5543      & 0.5149          & 1.0000        & 0.1721 \\
jm1               & FTTrans      & BIM         & 0.3  & 0.9233              & 0.9492 & 1.0364      & 0.5297        & 0.5149      & 0.4782          & 1.0000        & 0.1660 \\
jm1               & FTTrans      & C\&W & 1  & 0.4667              & 0.9999 & 0.1490      & 0.0761        & 0.1553      & 0.1442          & 0.6516        & 0.0447 \\
jm1               & FTTrans      & DeepFool      & 1  & 0.9233              & 0.7761 & 0.3426      & 0.1751        & 0.2691      & 0.2500          & 0.8199        & 0.0822 \\
BreastCancer      & LR & Gaussian      & 1  & 0.1140              & 1.0000 & 0.8718      & 0.4456        & 0.0988      & 0.0918          & 1.0000        & 0.0666 \\
BreastCancer      & LR & FGSM        & 0.3  & 1.0000              & 0.9977 & 1.4918      & 0.7625        & 0.0743      & 0.0690          & 1.0000        & 0.0568 \\
BreastCancer      & LR & PGD         & 0.3  & 1.0000              & 0.9977 & 1.4039      & 0.7176        & 0.0743      & 0.0690          & 1.0000        & 0.0566 \\
BreastCancer      & LR & BIM         & 0.3  & 1.0000              & 0.9977 & 1.4918      & 0.7625        & 0.0743      & 0.0690          & 1.0000        & 0.0568 \\
BreastCancer      & LR & C\&W & 1  & 0.9825              & 0.9991 & 0.4157      & 0.2125        & 0.0314      & 0.0292          & 0.9107        & 0.0251 \\
BreastCancer      & LR & DeepFool      & 1  & 1.0000              & 0.9363 & 0.3159      & 0.1615        & 0.0233      & 0.0216          & 0.8860        & 0.0190 \\
BreastCancer      & MLP                & Gaussian      & 1  & 0.0965              & 1.0000 & 0.8610      & 0.4401        & 0.0965      & 0.0896          & 1.0000        & 0.0654 \\
BreastCancer      & MLP                & FGSM        & 0.3  & 0.9912              & 0.9956 & 1.4714      & 0.7521        & 0.0744      & 0.0691          & 1.0000        & 0.0569 \\
BreastCancer      & MLP                & PGD         & 0.3  & 0.9912              & 0.9976 & 1.3779      & 0.7043        & 0.0743      & 0.0690          & 1.0000        & 0.0565 \\
BreastCancer      & MLP                & BIM         & 0.3  & 1.0000              & 0.9921 & 1.4967      & 0.7650        & 0.0747      & 0.0694          & 1.0000        & 0.0571 \\
BreastCancer      & MLP                & C\&W & 1  & 0.9035              & 0.9997 & 0.4474      & 0.2287        & 0.0344      & 0.0319          & 0.9223        & 0.0272 \\
BreastCancer      & MLP                & DeepFool      & 1  & 0.9474              & 0.9676 & 0.4797      & 0.2452        & 0.0433      & 0.0402          & 0.8148        & 0.0327 \\
BreastCancer      & TabTrans     & Gaussian      & 1  & 0.2368              & 1.0000 & 0.8826      & 0.4511        & 0.0920      & 0.0855          & 1.0000        & 0.0634 \\
BreastCancer      & TabTrans     & FGSM        & 0.3  & 0.6842              & 0.9987 & 1.4568      & 0.7446        & 0.0742      & 0.0689          & 1.0000        & 0.0567 \\
BreastCancer      & TabTrans     & PGD         & 0.3  & 0.8333              & 0.9993 & 0.9238      & 0.4722        & 0.0737      & 0.0685          & 1.0000        & 0.0540 \\
BreastCancer      & TabTrans     & BIM         & 0.3  & 0.6579              & 0.8644 & 0.8821      & 0.4508        & 0.0807      & 0.0749          & 1.0000        & 0.0570 \\
BreastCancer      & TabTrans     & C\&W & 1  & 0.6667              & 0.9991 & 0.2530      & 0.1293        & 0.0185      & 0.0172          & 0.8421        & 0.0154 \\
BreastCancer      & TabTrans     & DeepFool      & 1  & 0.7193              & 0.8890 & 0.2479      & 0.1267        & 0.0282      & 0.0262          & 0.6829        & 0.0212 \\
BreastCancer      & FTTrans      & Gaussian      & 1  & 0.1842              & 1.0000 & 0.9038      & 0.4620        & 0.0959      & 0.0891          & 1.0000        & 0.0655 \\
BreastCancer      & FTTrans      & FGSM        & 0.3  & 0.3509              & 0.9975 & 1.4770      & 0.7549        & 0.0743      & 0.0690          & 1.0000        & 0.0568 \\
BreastCancer      & FTTrans      & PGD         & 0.3  & 0.5614              & 0.9979 & 0.8884      & 0.4541        & 0.0731      & 0.0679          & 1.0000        & 0.0534 \\
BreastCancer      & FTTrans      & BIM         & 0.3  & 0.2281              & 0.8359 & 0.5681      & 0.2904        & 0.0588      & 0.0546          & 0.9231        & 0.0422 \\
BreastCancer      & FTTrans      & C\&W & 1  & 0.5351              & 0.9989 & 0.4630      & 0.2366        & 0.0400      & 0.0371          & 0.6721        & 0.0304 \\
BreastCancer      & FTTrans      & DeepFool      & 1  & 0.3509              & 0.9225 & 0.7440      & 0.3803        & 0.0608      & 0.0564          & 0.9250        & 0.0451 \\
Wine-White & LR & Gaussian      & 1  & 0.3235              & 0.9994 & 0.8094      & 0.4137        & 0.5372      & 0.4989          & 1.0000        & 0.1560 \\
Wine-White & LR & FGSM        & 0.3  & 0.9827              & 0.9997 & 0.8930      & 0.4564        & 0.3227      & 0.2997          & 1.0000        & 0.1332 \\
Wine-White & LR & PGD         & 0.3  & 0.9378              & 0.9997 & 0.8370      & 0.4278        & 0.3227      & 0.2997          & 1.0000        & 0.1306 \\
Wine-White & LR & BIM         & 0.3  & 0.9827              & 0.9997 & 0.8930      & 0.4564        & 0.3227      & 0.2997          & 1.0000        & 0.1332 \\
Wine-White & LR & C\&W & 1  & 0.9878              & 0.9936 & 0.3592      & 0.1836        & 0.1859      & 0.1727          & 0.7231        & 0.0737 \\
Wine-White & LR & DeepFool      & 1  & 1.0000              & 0.6745 & 0.3166      & 0.1618        & 0.2130      & 0.1978          & 0.7265        & 0.0713 \\
Wine-White & MLP                & Gaussian      & 1  & 0.3939              & 0.9993 & 0.8239      & 0.4211        & 0.5428      & 0.5042          & 1.0000        & 0.1575 \\
Wine-White & MLP                & FGSM        & 0.3  & 1.0000              & 0.9997 & 0.8665      & 0.4429        & 0.3227      & 0.2997          & 1.0000        & 0.1320 \\
Wine-White & MLP                & PGD         & 0.3  & 1.0000              & 0.9997 & 0.7106      & 0.3632        & 0.3223      & 0.2994          & 1.0000        & 0.1239 \\
Wine-White & MLP                & BIM         & 0.3  & 1.0000              & 0.9761 & 0.7791      & 0.3982        & 0.3312      & 0.3076          & 1.0000        & 0.1287 \\
Wine-White & MLP                & C\&W & 1  & 1.0000              & 0.9952 & 0.1535      & 0.0784        & 0.0890      & 0.0827          & 0.4265        & 0.0359 \\
Wine-White & MLP                & DeepFool      & 1  & 1.0000              & 0.7468 & 0.1251      & 0.0639        & 0.0841      & 0.0781          & 0.4643        & 0.0317 \\
Wine-White & TabTrans     & Gaussian      & 1  & 0.3857              & 0.9993 & 0.8174      & 0.4178        & 0.5450      & 0.5062          & 1.0000        & 0.1573 \\
Wine-White & TabTrans     & FGSM        & 0.3  & 0.9969              & 0.9997 & 0.8463      & 0.4326        & 0.3227      & 0.2997          & 1.0000        & 0.1311 \\
Wine-White & TabTrans     & PGD         & 0.3  & 0.9571              & 1.0000 & 0.5375      & 0.2747        & 0.3122      & 0.2900          & 0.9776        & 0.1101 \\
Wine-White & TabTrans     & BIM         & 0.3  & 0.9990              & 0.8524 & 0.5702      & 0.2914        & 0.3755      & 0.3488          & 0.9775        & 0.1180 \\
Wine-White & TabTrans     & C\&W & 1  & 0.9980              & 0.9950 & 0.1559      & 0.0797        & 0.0938      & 0.0871          & 0.4755        & 0.0372 \\
Wine-White & TabTrans     & DeepFool      & 1  & 1.0000              & 0.7315 & 0.1615      & 0.0825        & 0.1215      & 0.1129          & 0.3449        & 0.0400 \\
Wine-White & FTTrans      & Gaussian      & 1  & 0.3429              & 0.9997 & 0.8063      & 0.4121        & 0.5364      & 0.4983          & 1.0000        & 0.1557 \\
Wine-White & FTTrans      & FGSM        & 0.3  & 0.9867              & 0.9995 & 0.8852      & 0.4524        & 0.3227      & 0.2998          & 1.0000        & 0.1328 \\
Wine-White & FTTrans      & PGD         & 0.3  & 0.9327              & 0.9998 & 0.6832      & 0.3492        & 0.3215      & 0.2986          & 1.0000        & 0.1221 \\
Wine-White & FTTrans      & BIM         & 0.3  & 0.9939              & 0.9417 & 0.6582      & 0.3364        & 0.3071      & 0.2853          & 0.9990        & 0.1174 \\
Wine-White & FTTrans      & C\&W & 1  & 0.9857              & 0.9954 & 0.1579      & 0.0807        & 0.1113      & 0.1034          & 0.3002        & 0.0383 \\
Wine-White & FTTrans      & DeepFool      & 1  & 0.9969              & 0.7512 & 0.2762      & 0.1412        & 0.2088      & 0.1940          & 0.3153        & 0.0600 \\
Wine-Red   & LR & Gaussian      & 1  & 0.4063              & 0.9965 & 0.8256      & 0.4220        & 0.3677      & 0.3416          & 0.9923        & 0.1371 \\
Wine-Red   & LR & FGSM        & 0.3  & 1.0000              & 0.9955 & 0.8850      & 0.4523        & 0.2197      & 0.2040          & 1.0000        & 0.1101 \\
Wine-Red   & LR & PGD         & 0.3  & 1.0000              & 0.9955 & 0.8296      & 0.4240        & 0.2197      & 0.2040          & 1.0000        & 0.1083 \\
Wine-Red   & LR & BIM         & 0.3  & 1.0000              & 0.9955 & 0.8850      & 0.4523        & 0.2197      & 0.2040          & 1.0000        & 0.1101 \\
Wine-Red   & LR & C\&W & 1  & 1.0000              & 0.9835 & 0.1703      & 0.0871        & 0.0557      & 0.0517          & 0.2156        & 0.0278 \\
Wine-Red   & LR & DeepFool      & 1  & 1.0000              & 0.7188 & 0.1288      & 0.0658        & 0.0418      & 0.0388          & 0.2313        & 0.0218 \\
Wine-Red   & MLP                & Gaussian      & 1  & 0.4313              & 0.9960 & 0.8200      & 0.4191        & 0.3668      & 0.3407          & 0.9855        & 0.1366 \\
Wine-Red   & MLP                & FGSM        & 0.3  & 1.0000              & 0.9940 & 0.8872      & 0.4535        & 0.2200      & 0.2044          & 1.0000        & 0.1102 \\
Wine-Red   & MLP                & PGD         & 0.3  & 1.0000              & 0.9991 & 0.7989      & 0.4083        & 0.2188      & 0.2032          & 1.0000        & 0.1071 \\
Wine-Red   & MLP                & BIM         & 0.3  & 1.0000              & 0.9946 & 0.8605      & 0.4398        & 0.2199      & 0.2042          & 1.0000        & 0.1094 \\
Wine-Red   & MLP                & C\&W & 1  & 1.0000              & 0.9844 & 0.1546      & 0.0790        & 0.0570      & 0.0529          & 0.1969        & 0.0270 \\
Wine-Red   & MLP                & DeepFool      & 1  & 1.0000              & 0.7287 & 0.1618      & 0.0827        & 0.0679      & 0.0631          & 0.1563        & 0.0283 \\
Wine-Red   & TabTrans     & Gaussian      & 1  & 0.4344              & 0.9974 & 0.8297      & 0.4241        & 0.3701      & 0.3438          & 1.0000        & 0.1379 \\
Wine-Red   & TabTrans     & FGSM        & 0.3  & 0.7938              & 0.9753 & 0.8615      & 0.4404        & 0.2173      & 0.2018          & 0.9882        & 0.1083 \\
Wine-Red   & TabTrans     & PGD         & 0.3  & 0.8125              & 0.9983 & 0.5685      & 0.2906        & 0.2108      & 0.1958          & 0.9231        & 0.0944 \\
Wine-Red   & TabTrans     & BIM         & 0.3  & 0.7969              & 0.8841 & 0.4695      & 0.2399        & 0.1669      & 0.1550          & 0.8157        & 0.0774 \\
Wine-Red   & TabTrans     & C\&W & 1  & 0.7969              & 0.9811 & 0.1109      & 0.0567        & 0.0412      & 0.0383          & 0.1843        & 0.0203 \\
Wine-Red   & TabTrans     & DeepFool      & 1  & 0.7969              & 0.6510 & 0.1385      & 0.0708        & 0.0652      & 0.0606          & 0.1765        & 0.0268 \\
Wine-Red   & FTTrans      & Gaussian      & 1  & 0.4281              & 0.9960 & 0.8259      & 0.4222        & 0.3699      & 0.3435          & 1.0000        & 0.1376 \\
Wine-Red   & FTTrans      & FGSM        & 0.3  & 0.9250              & 0.9926 & 0.8800      & 0.4498        & 0.2203      & 0.2047          & 1.0000        & 0.1101 \\
Wine-Red   & FTTrans      & PGD         & 0.3  & 0.9250              & 0.9982 & 0.6888      & 0.3521        & 0.2156      & 0.2003          & 0.9730        & 0.1018 \\
Wine-Red   & FTTrans      & BIM         & 0.3  & 0.9250              & 0.9509 & 0.6710      & 0.3430        & 0.1935      & 0.1798          & 0.9122        & 0.0945 \\
Wine-Red   & FTTrans      & C\&W & 1  & 0.9938              & 0.9854 & 0.1810      & 0.0925        & 0.0736      & 0.0684          & 0.1792        & 0.0316 \\
Wine-Red   & FTTrans      & DeepFool      & 1  & 0.9250              & 0.7067 & 0.2618      & 0.1338        & 0.1308      & 0.1215          & 0.3007        & 0.0492 \\
phoneme           & LR & Gaussian      & 1  & 0.3062              & 1.0000 & 0.7827      & 0.4001        & 0.6939      & 0.6445          & 0.9879        & 0.1652 \\
phoneme           & LR & FGSM        & 0.3  & 0.6605              & 1.0000 & 0.6507      & 0.3326        & 0.3756      & 0.3489          & 0.7213        & 0.1214 \\
phoneme           & LR & PGD         & 0.3  & 0.5449              & 1.0000 & 0.5967      & 0.3050        & 0.3746      & 0.3480          & 0.6078        & 0.1140 \\
phoneme           & LR & BIM         & 0.3  & 0.6605              & 1.0000 & 0.6507      & 0.3326        & 0.3756      & 0.3489          & 0.7213        & 0.1214 \\
phoneme           & LR & C\&W & 1  & 0.2905              & 0.9178 & 0.0000      & 0.0000        & 0.0000      & 0.0000          & 0.0669        & 0.0005 \\
phoneme           & LR & DeepFool      & 1  & 0.9991              & 0.7093 & 0.3903      & 0.1995        & 0.3698      & 0.3435          & 0.2398        & 0.0743 \\
phoneme           & MLP                & Gaussian      & 1  & 0.3432              & 1.0000 & 0.8085      & 0.4132        & 0.7091      & 0.6586          & 0.9865        & 0.1683 \\
phoneme           & MLP                & FGSM        & 0.3  & 1.0000              & 1.0000 & 0.6562      & 0.3354        & 0.3756      & 0.3489          & 0.7123        & 0.1215 \\
phoneme           & MLP                & PGD         & 0.3  & 0.9954              & 1.0000 & 0.5568      & 0.2846        & 0.3746      & 0.3479          & 0.3931        & 0.1010 \\
phoneme           & MLP                & BIM         & 0.3  & 1.0000              & 0.9630 & 0.6144      & 0.3141        & 0.3931      & 0.3651          & 0.4838        & 0.1110 \\
phoneme           & MLP                & C\&W & 1  & 0.4607              & 0.9627 & 0.3044      & 0.1556        & 0.2441      & 0.2268          & 0.3233        & 0.0671 \\
phoneme           & MLP                & DeepFool      & 1  & 1.0000              & 0.7872 & 0.2305      & 0.1178        & 0.1866      & 0.1733          & 0.0463        & 0.0273 \\
phoneme           & TabTrans     & Gaussian      & 1  & 0.3414              & 1.0000 & 0.7911      & 0.4044        & 0.6967      & 0.6471          & 0.9837        & 0.1660 \\
phoneme           & TabTrans     & FGSM        & 0.3  & 0.6337              & 0.8701 & 0.5664      & 0.2895        & 0.3268      & 0.3036          & 0.8248        & 0.1101 \\
phoneme           & TabTrans     & PGD         & 0.3  & 0.8455              & 1.0000 & 0.4353      & 0.2225        & 0.3511      & 0.3261          & 0.2352        & 0.0783 \\
phoneme           & TabTrans     & BIM         & 0.3  & 0.8603              & 0.8417 & 0.3839      & 0.1962        & 0.2920      & 0.2712          & 0.1129        & 0.0534 \\
phoneme           & TabTrans     & C\&W & 1  & 0.3099              & 0.9475 & 0.0971      & 0.0496        & 0.0760      & 0.0706          & 0.1493        & 0.0241 \\
phoneme           & TabTrans     & DeepFool      & 1  & 0.8557              & 0.7524 & 0.3103      & 0.1586        & 0.2714      & 0.2521          & 0.2908        & 0.0668 \\
phoneme           & FTTrans      & Gaussian      & 1  & 0.3432              & 1.0000 & 0.8006      & 0.4092        & 0.6977      & 0.6480          & 0.9811        & 0.1668 \\
phoneme           & FTTrans      & FGSM        & 0.3  & 0.5578              & 1.0000 & 0.6559      & 0.3353        & 0.3756      & 0.3489          & 0.9287        & 0.1265 \\
phoneme           & FTTrans      & PGD         & 0.3  & 0.7956              & 1.0000 & 0.4553      & 0.2327        & 0.3584      & 0.3329          & 0.2698        & 0.0836 \\
phoneme           & FTTrans      & BIM         & 0.3  & 0.8131              & 0.8480 & 0.4499      & 0.2299        & 0.4178      & 0.3881          & 0.2856        & 0.0864 \\
phoneme           & FTTrans      & C\&W & 1  & 0.4191              & 0.9656 & 0.1861      & 0.0951        & 0.1459      & 0.1355          & 0.2075        & 0.0424 \\
phoneme           & FTTrans      & DeepFool      & 1  & 0.8363              & 0.7611 & 0.3481      & 0.1779        & 0.2914      & 0.2707          & 0.4425        & 0.0779 \\
MiniBooNE         & LR & Gaussian      & 1  & 0.2798              & 0.9978 & 0.7557      & 0.3863        & 0.1493      & 0.1387          & 1.0000        & 0.0852 \\
MiniBooNE         & LR & FGSM        & 0.3  & 1.0000              & 0.9977 & 1.6490      & 0.8429        & 0.1380      & 0.1282          & 1.0000        & 0.0915 \\
MiniBooNE         & LR & PGD         & 0.3  & 1.0000              & 0.9977 & 1.6217      & 0.8289        & 0.1380      & 0.1282          & 1.0000        & 0.0914 \\
MiniBooNE         & LR & BIM         & 0.3  & 1.0000              & 0.9977 & 1.6490      & 0.8429        & 0.1380      & 0.1282          & 1.0000        & 0.0915 \\
MiniBooNE         & LR & C\&W & 1  & 0.2794              & 0.9827 & 0.0100      & 0.0051        & 0.0022      & 0.0020          & 0.2083        & 0.0020 \\
MiniBooNE         & LR & DeepFool      & 1  & 1.0000              & 0.7701 & 0.1316      & 0.0673        & 0.0481      & 0.0447          & 0.7855        & 0.0256 \\
MiniBooNE         & MLP                & Gaussian      & 1  & 0.3204              & 0.9998 & 0.7529      & 0.3848        & 0.1477      & 0.1372          & 1.0000        & 0.0846 \\
MiniBooNE         & MLP                & FGSM        & 0.3  & 1.0000              & 0.9983 & 1.5766      & 0.8059        & 0.1376      & 0.1278          & 1.0000        & 0.0909 \\
MiniBooNE         & MLP                & PGD         & 0.3  & 0.9961              & 0.9990 & 1.4559      & 0.7441        & 0.1374      & 0.1276          & 1.0000        & 0.0899 \\
MiniBooNE         & MLP                & BIM         & 0.3  & 1.0000              & 0.9953 & 1.5173      & 0.7755        & 0.1379      & 0.1281          & 1.0000        & 0.0906 \\
MiniBooNE         & MLP                & C\&W & 1  & 0.8172              & 0.9961 & 0.0526      & 0.0269        & 0.0110      & 0.0102          & 0.7517        & 0.0078 \\
MiniBooNE         & MLP                & DeepFool      & 1  & 1.0000              & 0.8357 & 0.0889      & 0.0455        & 0.0254      & 0.0236          & 0.8508        & 0.0155 \\
MiniBooNE         & TabTrans     & Gaussian      & 1  & 0.3671              & 0.9987 & 0.7553      & 0.3860        & 0.1482      & 0.1376          & 1.0000        & 0.0848 \\
MiniBooNE         & TabTrans     & FGSM        & 0.3  & 1.0000              & 0.9983 & 1.5165      & 0.7751        & 0.1376      & 0.1278          & 1.0000        & 0.0905 \\
MiniBooNE         & TabTrans     & PGD         & 0.3  & 1.0000              & 0.9999 & 1.5001      & 0.7667        & 0.1371      & 0.1273          & 1.0000        & 0.0901 \\
MiniBooNE         & TabTrans     & BIM         & 0.3  & 1.0000              & 0.9863 & 1.4737      & 0.7532        & 0.1391      & 0.1292          & 1.0000        & 0.0908 \\
MiniBooNE         & TabTrans     & C\&W & 1  & 0.9470              & 0.9970 & 0.0362      & 0.0185        & 0.0077      & 0.0072          & 0.7906        & 0.0057 \\
MiniBooNE         & TabTrans     & DeepFool      & 1  & 1.0000              & 0.8480 & 0.0464      & 0.0237        & 0.0116      & 0.0108          & 0.8604        & 0.0078 \\
MiniBooNE         & FTTrans      & Gaussian      & 1  & 0.3499              & 0.9998 & 0.7550      & 0.3859        & 0.1478      & 0.1373          & 1.0000        & 0.0847 \\
MiniBooNE         & FTTrans      & FGSM        & 0.3  & 1.0000              & 0.9981 & 1.5529      & 0.7937        & 0.1377      & 0.1279          & 1.0000        & 0.0908 \\
MiniBooNE         & FTTrans      & PGD         & 0.3  & 0.9970              & 0.9987 & 1.3504      & 0.6902        & 0.1375      & 0.1277          & 1.0000        & 0.0892 \\
MiniBooNE         & FTTrans      & BIM         & 0.3  & 1.0000              & 0.9710 & 0.9056      & 0.4629        & 0.1140      & 0.1058          & 1.0000        & 0.0738 \\
MiniBooNE         & FTTrans      & C\&W & 1  & 0.4610              & 0.9933 & 0.0305      & 0.0156        & 0.0082      & 0.0076          & 0.6035        & 0.0056 \\
MiniBooNE         & FTTrans      & DeepFool      & 1  & 0.9997              & 0.8382 & 0.1968      & 0.1006        & 0.0668      & 0.0620          & 0.8454        & 0.0356 \\
\end{longtable}
}

\section{Trade-off Analysis}\label{appendix:trade-off}

\begin{figure}[htbp]
    \centering
    \includegraphics[width=0.75\linewidth]{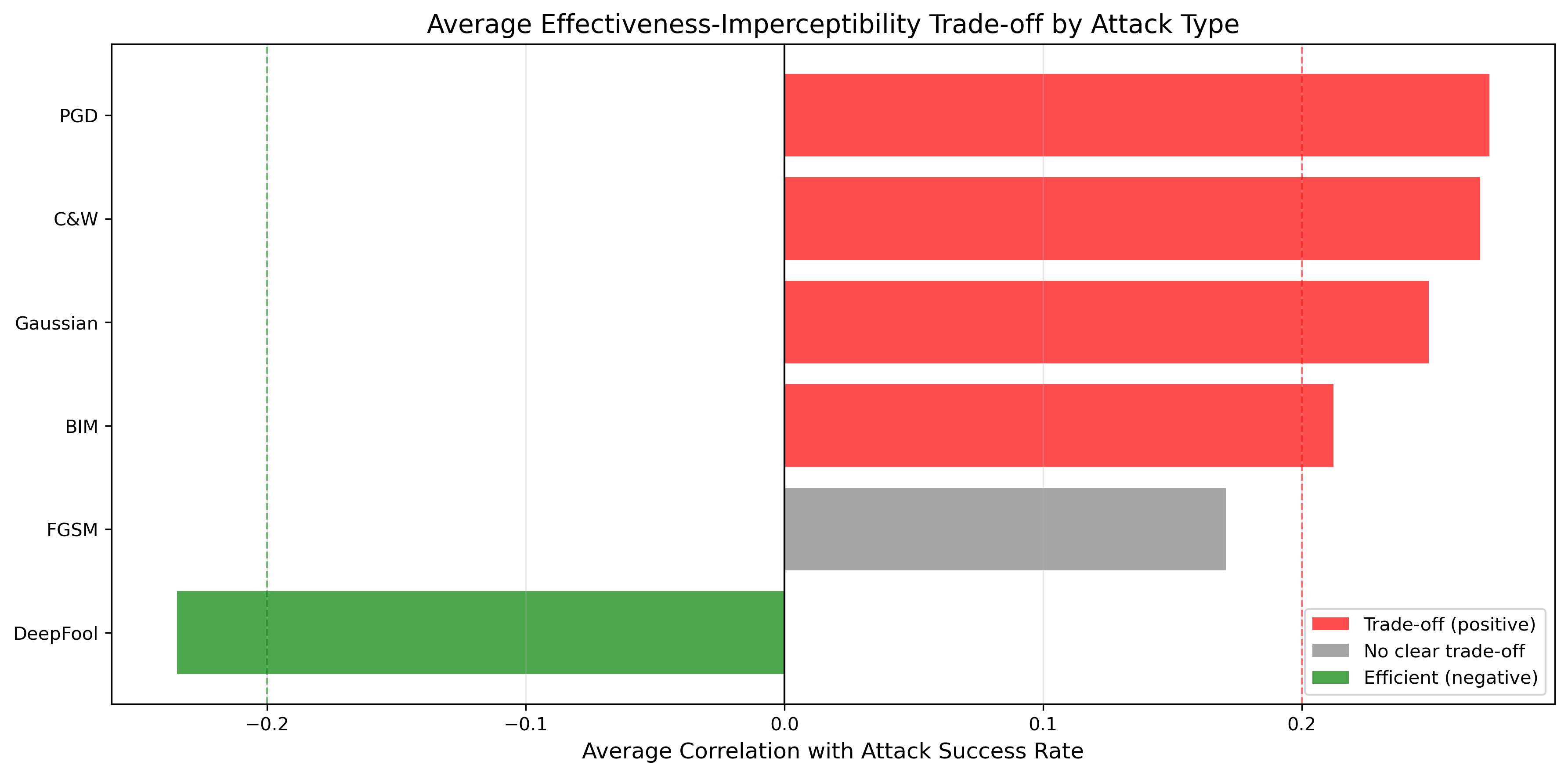}
    \caption{Average correlation between Attack Success Rate (ASR) and imperceptibility metrics by attack type. Positive correlations (red bars) indicate a trade-off where higher perturbation magnitudes lead to higher success rates. Negative correlations (green bars) indicate efficient attacks that achieve high success with lower perturbations. Gray bars indicate no clear relationship. Dashed lines mark the weak correlation threshold ($|r| = 0.2$).}
    \label{fig:trade-off-1}
\end{figure}

\begin{figure}[htbp]
    \centering
    \includegraphics[width=\linewidth]{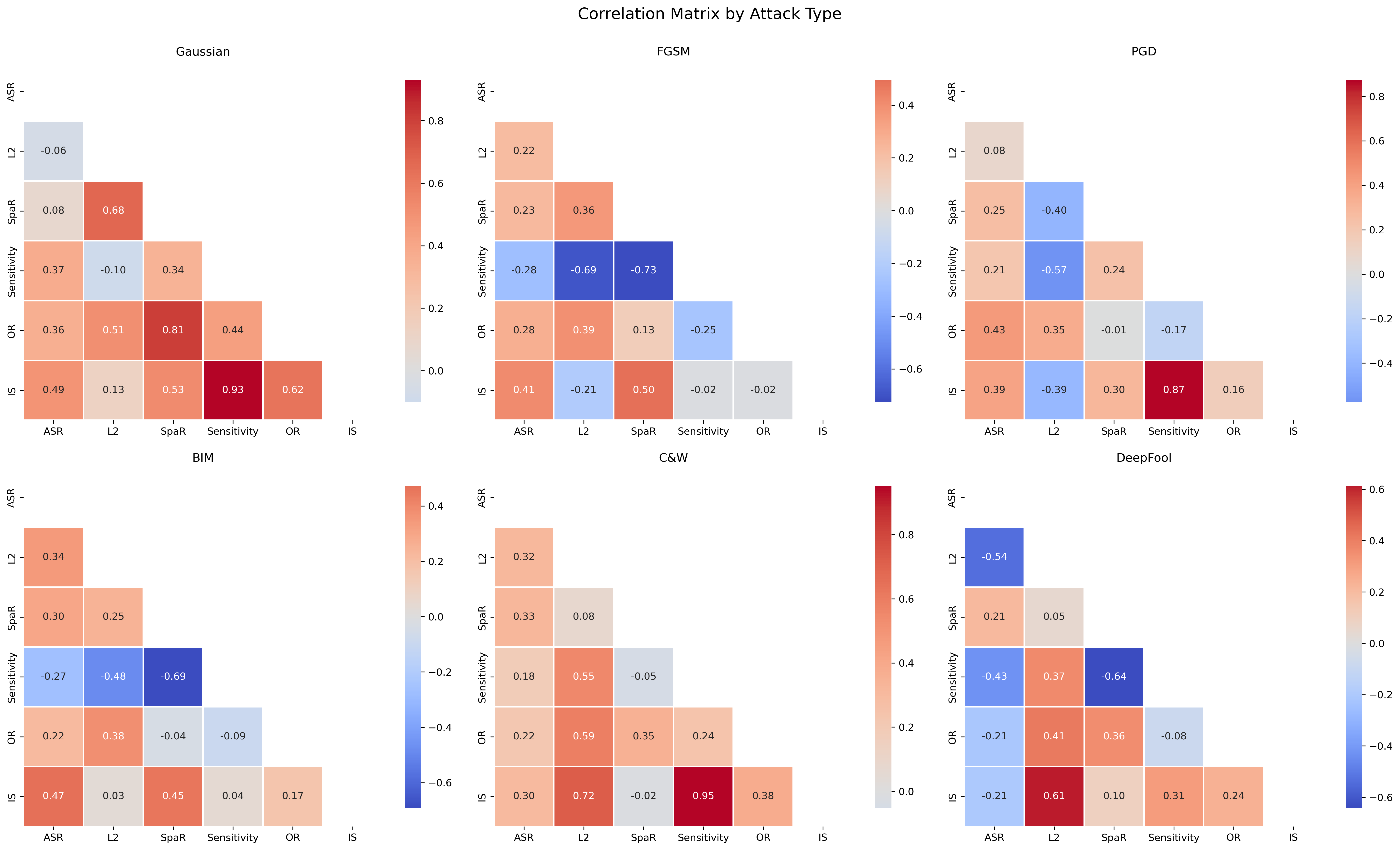}
    \caption{Correlation heatmap showing the relationship between Attack Success Rate and five imperceptibility metrics for each attack type. Red cells indicate positive correlations (trade-off), blue cells indicate negative correlations (efficient), and white cells indicate no correlation. }
    \label{fig:trade-off-2}
\end{figure}

\begin{figure}[htbp]
    \centering
    \includegraphics[width=\linewidth]{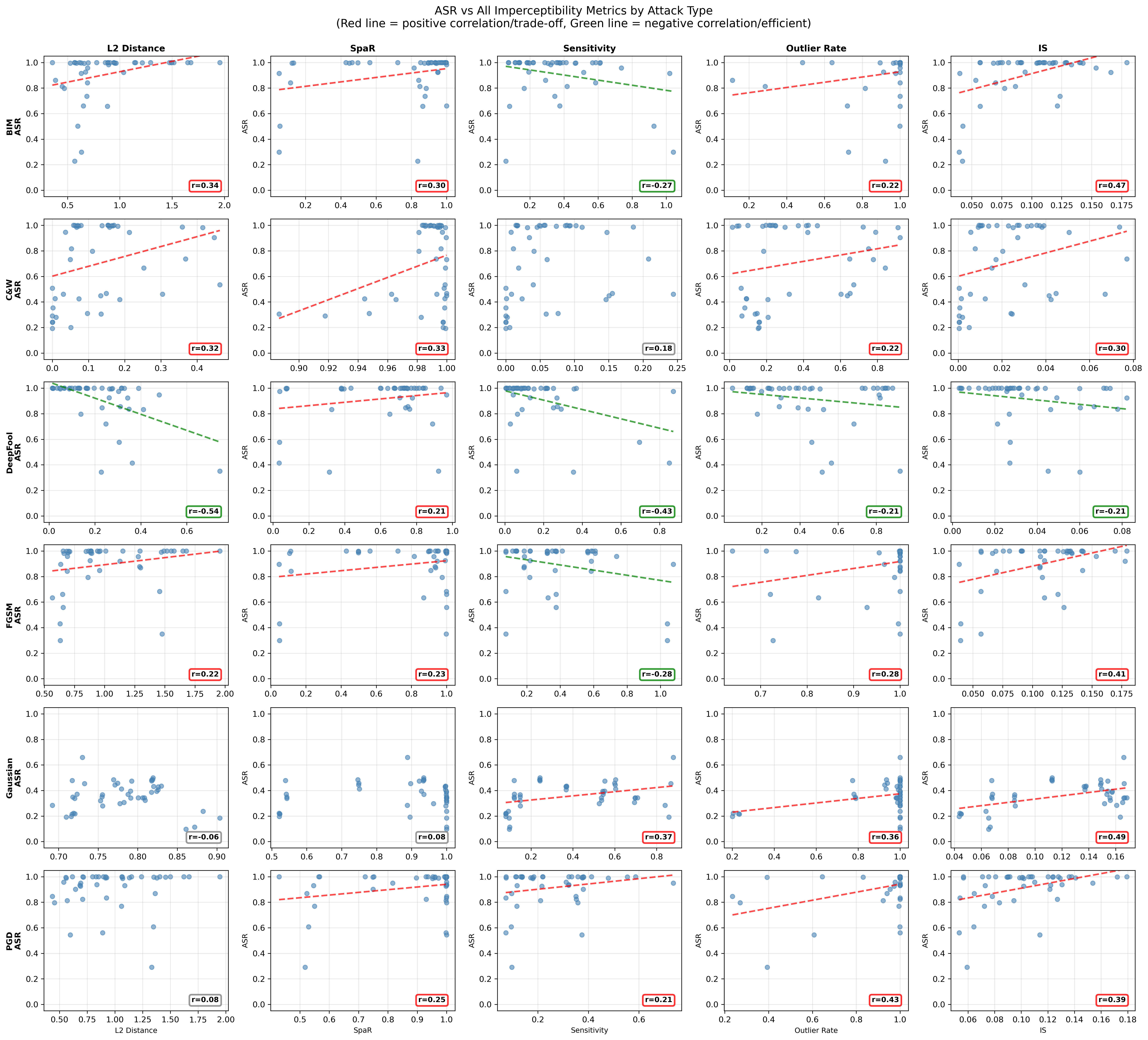}
    \caption{Scatter plots of ASR versus imperceptibility metrics by attack type. Trend lines show correlation direction: red (trade-off) or green (efficient). Each attack shows distinct patterns across different imperceptibility dimensions}
    \label{fig:trade-off-3}
\end{figure}
\newpage


\bibliographystyle{model5-names} 
\bibliography{bibliography}

\begin{thebibliography}{63}
\expandafter\ifx\csname natexlab\endcsname\relax\def\natexlab#1{#1}\fi
\providecommand{\url}[1]{\texttt{#1}}
\providecommand{\href}[2]{#2}
\providecommand{\path}[1]{#1}
\providecommand{\DOIprefix}{doi:}
\providecommand{\ArXivprefix}{arXiv:}
\providecommand{\URLprefix}{URL: }
\providecommand{\Pubmedprefix}{pmid:}
\providecommand{\doi}[1]{\href{http://dx.doi.org/#1}{\path{#1}}}
\providecommand{\Pubmed}[1]{\href{pmid:#1}{\path{#1}}}
\providecommand{\bibinfo}[2]{#2}
\ifx\xfnm\relax \def\xfnm[#1]{\unskip,\space#1}\fi
\bibitem[{Abdullah et~al.(2019)Abdullah, Garcia, Peeters, Traynor, Butler \& Wilson}]{abdullah2019practical}
\bibinfo{author}{Abdullah, H.}, \bibinfo{author}{Garcia, W.}, \bibinfo{author}{Peeters, C.}, \bibinfo{author}{Traynor, P.}, \bibinfo{author}{Butler, K.~R.}, \& \bibinfo{author}{Wilson, J.} (\bibinfo{year}{2019}).
\newblock \bibinfo{title}{Practical hidden voice attacks against speech and speaker recognition systems}.
\newblock {\it \bibinfo{journal}{arXiv preprint arXiv:1904.05734}\/}, .
\bibitem[{An et~al.(2019)An, Xiao, Stewart \& Sun}]{an2019longitudinal}
\bibinfo{author}{An, S.}, \bibinfo{author}{Xiao, C.}, \bibinfo{author}{Stewart, W.~F.}, \& \bibinfo{author}{Sun, J.} (\bibinfo{year}{2019}).
\newblock \bibinfo{title}{Longitudinal adversarial attack on electronic health records data}.
\newblock In {\it \bibinfo{booktitle}{The world wide web conference}\/} (pp. \bibinfo{pages}{2558--2564}).
\bibitem[{Asl et~al.(2024)Asl, Rafiei, Alohaly \& Takabi}]{asl2024semantic}
\bibinfo{author}{Asl, J.~R.}, \bibinfo{author}{Rafiei, M.~H.}, \bibinfo{author}{Alohaly, M.}, \& \bibinfo{author}{Takabi, D.} (\bibinfo{year}{2024}).
\newblock \bibinfo{title}{A semantic, syntactic, and context-aware natural language adversarial example generator}.
\newblock {\it \bibinfo{journal}{IEEE Transactions on Dependable and Secure Computing}\/},  {\it \bibinfo{volume}{21}\/}, \bibinfo{pages}{4754--4769}.
\bibitem[{Assion et~al.(2019)Assion, Schlicht, Gre{\ss}ner, G{\"{u}}nther, H{\"{u}}ger, Schmidt \& Rasheed}]{assion2019attackgenerator}
\bibinfo{author}{Assion, F.}, \bibinfo{author}{Schlicht, P.}, \bibinfo{author}{Gre{\ss}ner, F.}, \bibinfo{author}{G{\"{u}}nther, W.}, \bibinfo{author}{H{\"{u}}ger, F.}, \bibinfo{author}{Schmidt, N.~M.}, \& \bibinfo{author}{Rasheed, U.} (\bibinfo{year}{2019}).
\newblock \bibinfo{title}{The attack generator: {A} systematic approach towards constructing adversarial attacks}.
\newblock In {\it \bibinfo{booktitle}{{IEEE} Conference on Computer Vision and Pattern Recognition Workshops, {CVPR} Workshops 2019, Long Beach, CA, USA, June 16-20, 2019}\/} (pp. \bibinfo{pages}{1370--1379}).
\newblock \bibinfo{publisher}{Computer Vision Foundation / {IEEE}}.
\newblock \DOIprefix\doi{10.1109/CVPRW.2019.00177}.
\bibitem[{Aydin \& Temizel(2023)}]{aydin2023adversarial}
\bibinfo{author}{Aydin, A.}, \& \bibinfo{author}{Temizel, A.} (\bibinfo{year}{2023}).
\newblock \bibinfo{title}{Adversarial image generation by spatial transformation in perceptual colorspaces}.
\newblock {\it \bibinfo{journal}{Pattern Recognition Letters}\/},  {\it \bibinfo{volume}{174}\/}, \bibinfo{pages}{92--98}.
\bibitem[{Ballet et~al.(2019)Ballet, Renard, Aigrain, Laugel, Frossard \& Detyniecki}]{ballet2019imperceptible}
\bibinfo{author}{Ballet, V.}, \bibinfo{author}{Renard, X.}, \bibinfo{author}{Aigrain, J.}, \bibinfo{author}{Laugel, T.}, \bibinfo{author}{Frossard, P.}, \& \bibinfo{author}{Detyniecki, M.} (\bibinfo{year}{2019}).
\newblock \bibinfo{title}{Imperceptible adversarial attacks on tabular data}.
\newblock \href{http://arxiv.org/abs/1911.03274}{\tt arXiv:1911.03274}.
\bibitem[{Biggio et~al.(2013)Biggio, Corona, Maiorca, Nelson, {\v{S}}rndi{\'c}, Laskov, Giacinto \& Roli}]{biggio2013evasion}
\bibinfo{author}{Biggio, B.}, \bibinfo{author}{Corona, I.}, \bibinfo{author}{Maiorca, D.}, \bibinfo{author}{Nelson, B.}, \bibinfo{author}{{\v{S}}rndi{\'c}, N.}, \bibinfo{author}{Laskov, P.}, \bibinfo{author}{Giacinto, G.}, \& \bibinfo{author}{Roli, F.} (\bibinfo{year}{2013}).
\newblock \bibinfo{title}{Evasion attacks against machine learning at test time}.
\newblock In {\it \bibinfo{booktitle}{Joint European conference on machine learning and knowledge discovery in databases}\/} (pp. \bibinfo{pages}{387--402}).
\newblock \bibinfo{organization}{Springer}.
\bibitem[{Biggio \& Roli(2018)}]{biggio2018wild}
\bibinfo{author}{Biggio, B.}, \& \bibinfo{author}{Roli, F.} (\bibinfo{year}{2018}).
\newblock \bibinfo{title}{Wild patterns: Ten years after the rise of adversarial machine learning}.
\newblock In {\it \bibinfo{booktitle}{Proceedings of the 2018 ACM SIGSAC Conference on Computer and Communications Security}\/} (pp. \bibinfo{pages}{2154--2156}).
\bibitem[{Borisov et~al.(2022)Borisov, Leemann, Se{\ss}ler, Haug, Pawelczyk \& Kasneci}]{borisov2022deep}
\bibinfo{author}{Borisov, V.}, \bibinfo{author}{Leemann, T.}, \bibinfo{author}{Se{\ss}ler, K.}, \bibinfo{author}{Haug, J.}, \bibinfo{author}{Pawelczyk, M.}, \& \bibinfo{author}{Kasneci, G.} (\bibinfo{year}{2022}).
\newblock \bibinfo{title}{Deep neural networks and tabular data: A survey}.
\newblock {\it \bibinfo{journal}{IEEE transactions on neural networks and learning systems}\/},  {\it \bibinfo{volume}{35}\/}, \bibinfo{pages}{7499--7519}.
\bibitem[{Brendel et~al.(2017)Brendel, Rauber \& Bethge}]{brendel2017decision}
\bibinfo{author}{Brendel, W.}, \bibinfo{author}{Rauber, J.}, \& \bibinfo{author}{Bethge, M.} (\bibinfo{year}{2017}).
\newblock \bibinfo{title}{Decision-based adversarial attacks: Reliable attacks against black-box machine learning models}.
\newblock \href{http://arxiv.org/abs/1712.04248}{\tt arXiv:1712.04248}.
\bibitem[{Carlini \& Wagner(2017)}]{carlini2017towards}
\bibinfo{author}{Carlini, N.}, \& \bibinfo{author}{Wagner, D.~A.} (\bibinfo{year}{2017}).
\newblock \bibinfo{title}{Towards evaluating the robustness of neural networks}.
\newblock In {\it \bibinfo{booktitle}{2017 {IEEE} Symposium on Security and Privacy, {SP} 2017}\/} (pp. \bibinfo{pages}{39--57}).
\newblock \bibinfo{publisher}{{IEEE} Computer Society}.
\newblock \DOIprefix\doi{10.1109/SP.2017.49}.
\bibitem[{Cartella et~al.(2021)Cartella, Anuncia{\c{c}}{\~{a}}o, Funabiki, Yamaguchi, Akishita \& Elshocht}]{Cartella2021adversarial}
\bibinfo{author}{Cartella, F.}, \bibinfo{author}{Anuncia{\c{c}}{\~{a}}o, O.}, \bibinfo{author}{Funabiki, Y.}, \bibinfo{author}{Yamaguchi, D.}, \bibinfo{author}{Akishita, T.}, \& \bibinfo{author}{Elshocht, O.} (\bibinfo{year}{2021}).
\newblock \bibinfo{title}{Adversarial attacks for tabular data: Application to fraud detection and imbalanced data}.
\newblock In {\it \bibinfo{booktitle}{Workshop on Artificial Intelligence Safety 2021}\/}.
\newblock \URLprefix \url{https://ceur-ws.org/Vol-2808/Paper\_4.pdf}.
\bibitem[{Chen et~al.(2017)Chen, Zhang, Sharma, Yi \& Hsieh}]{chen2017zoo}
\bibinfo{author}{Chen, P.}, \bibinfo{author}{Zhang, H.}, \bibinfo{author}{Sharma, Y.}, \bibinfo{author}{Yi, J.}, \& \bibinfo{author}{Hsieh, C.} (\bibinfo{year}{2017}).
\newblock \bibinfo{title}{{ZOO:} zeroth order optimization based black-box attacks to deep neural networks without training substitute models}.
\newblock In {\it \bibinfo{booktitle}{Proceedings of the 10th {ACM} Workshop on Artificial Intelligence and Security, AISec@CCS 2017}\/} (pp. \bibinfo{pages}{15--26}).
\newblock \DOIprefix\doi{10.1145/3128572.3140448}.
\bibitem[{Chernikova \& Oprea(2022)}]{Chernikova2022FENCE}
\bibinfo{author}{Chernikova, A.}, \& \bibinfo{author}{Oprea, A.} (\bibinfo{year}{2022}).
\newblock \bibinfo{title}{{FENCE:} feasible evasion attacks on neural networks in constrained environments}.
\newblock {\it \bibinfo{journal}{{ACM} Transactions on Privacy and Security}\/},  {\it \bibinfo{volume}{25}\/}, \bibinfo{pages}{34:1--34:34}. \DOIprefix\doi{10.1145/3544746}.
\bibitem[{Cin{\`a} et~al.(2024)Cin{\`a}, Rony, Pintor, Demetrio, Demontis, Biggio, Ayed \& Roli}]{cina2024attackbench}
\bibinfo{author}{Cin{\`a}, A.~E.}, \bibinfo{author}{Rony, J.}, \bibinfo{author}{Pintor, M.}, \bibinfo{author}{Demetrio, L.}, \bibinfo{author}{Demontis, A.}, \bibinfo{author}{Biggio, B.}, \bibinfo{author}{Ayed, I.~B.}, \& \bibinfo{author}{Roli, F.} (\bibinfo{year}{2024}).
\newblock \bibinfo{title}{Attackbench: Evaluating gradient-based attacks for adversarial examples}.
\newblock \href{http://arxiv.org/abs/2404.19460}{\tt arXiv:2404.19460}.
\bibitem[{Cost \& Salzberg(1993)}]{cost1993weighted}
\bibinfo{author}{Cost, S.}, \& \bibinfo{author}{Salzberg, S.} (\bibinfo{year}{1993}).
\newblock \bibinfo{title}{A weighted nearest neighbor algorithm for learning with symbolic features}.
\newblock {\it \bibinfo{journal}{Machine Learning}\/},  {\it \bibinfo{volume}{10}\/}, \bibinfo{pages}{57--78}. \DOIprefix\doi{10.1023/A:1022664626993}.
\bibitem[{Croce et~al.(2021)Croce, Andriushchenko, Sehwag, Debenedetti, Flammarion, Chiang, Mittal \& Hein}]{croce2020robustbench}
\bibinfo{author}{Croce, F.}, \bibinfo{author}{Andriushchenko, M.}, \bibinfo{author}{Sehwag, V.}, \bibinfo{author}{Debenedetti, E.}, \bibinfo{author}{Flammarion, N.}, \bibinfo{author}{Chiang, M.}, \bibinfo{author}{Mittal, P.}, \& \bibinfo{author}{Hein, M.} (\bibinfo{year}{2021}).
\newblock \bibinfo{title}{Robustbench: A standardized adversarial robustness benchmark}.
\newblock In {\it \bibinfo{booktitle}{Proceedings of the Neural Information Processing Systems Track on Datasets and Benchmarks 1, NeurIPS Datasets and Benchmarks 2021}\/}.
\bibitem[{Dong et~al.(2020)Dong, Fu, Yang, Pang, Su, Xiao \& Zhu}]{dong2020benchmarking}
\bibinfo{author}{Dong, Y.}, \bibinfo{author}{Fu, Q.}, \bibinfo{author}{Yang, X.}, \bibinfo{author}{Pang, T.}, \bibinfo{author}{Su, H.}, \bibinfo{author}{Xiao, Z.}, \& \bibinfo{author}{Zhu, J.} (\bibinfo{year}{2020}).
\newblock \bibinfo{title}{Benchmarking adversarial robustness on image classification}.
\newblock In {\it \bibinfo{booktitle}{2020 {IEEE/CVF} Conference on Computer Vision and Pattern Recognition, {CVPR} 2020}\/} (pp. \bibinfo{pages}{321--331}).
\newblock \DOIprefix\doi{10.1109/CVPR42600.2020.00040}.
\bibitem[{Erdemir et~al.(2021)Erdemir, Bickford, Melis \& Aydore}]{erdemir2021adversarial}
\bibinfo{author}{Erdemir, E.}, \bibinfo{author}{Bickford, J.}, \bibinfo{author}{Melis, L.}, \& \bibinfo{author}{Aydore, S.} (\bibinfo{year}{2021}).
\newblock \bibinfo{title}{Adversarial robustness with non-uniform perturbations}.
\newblock In {\it \bibinfo{booktitle}{Advances in Neural Information Processing Systems 34: Annual Conference on Neural Information Processing Systems 2021, NeurIPS 2021}\/} (pp. \bibinfo{pages}{19147--19159}).
\bibitem[{Gao et~al.(2024)Gao, Gu, Vosoughi \& Mehnaz}]{gao2024semantic}
\bibinfo{author}{Gao, C.}, \bibinfo{author}{Gu, K.}, \bibinfo{author}{Vosoughi, S.}, \& \bibinfo{author}{Mehnaz, S.} (\bibinfo{year}{2024}).
\newblock \bibinfo{title}{Semantic-preserving adversarial example attack against bert}.
\newblock In {\it \bibinfo{booktitle}{Proceedings of the 4th Workshop on Trustworthy Natural Language Processing (TrustNLP 2024)}\/} (pp. \bibinfo{pages}{202--207}).
\bibitem[{Geirhos et~al.(2018)Geirhos, Rubisch, Michaelis, Bethge, Wichmann \& Brendel}]{geirhos2018imagenet}
\bibinfo{author}{Geirhos, R.}, \bibinfo{author}{Rubisch, P.}, \bibinfo{author}{Michaelis, C.}, \bibinfo{author}{Bethge, M.}, \bibinfo{author}{Wichmann, F.~A.}, \& \bibinfo{author}{Brendel, W.} (\bibinfo{year}{2018}).
\newblock \bibinfo{title}{Imagenet-trained cnns are biased towards texture; increasing shape bias improves accuracy and robustness}.
\newblock In {\it \bibinfo{booktitle}{International conference on learning representations}\/}.
\bibitem[{Goodfellow et~al.(2015)Goodfellow, Shlens \& Szegedy}]{goodfellow2015explaining}
\bibinfo{author}{Goodfellow, I.~J.}, \bibinfo{author}{Shlens, J.}, \& \bibinfo{author}{Szegedy, C.} (\bibinfo{year}{2015}).
\newblock \bibinfo{title}{Explaining and harnessing adversarial examples}.
\newblock In {\it \bibinfo{booktitle}{3rd International Conference on Learning Representations, {ICLR} 2015}\/}.
\bibitem[{Gorishniy et~al.(2021)Gorishniy, Rubachev, Khrulkov \& Babenko}]{gorishniy2021revisiting}
\bibinfo{author}{Gorishniy, Y.}, \bibinfo{author}{Rubachev, I.}, \bibinfo{author}{Khrulkov, V.}, \& \bibinfo{author}{Babenko, A.} (\bibinfo{year}{2021}).
\newblock \bibinfo{title}{Revisiting deep learning models for tabular data}.
\newblock In {\it \bibinfo{booktitle}{Advances in Neural Information Processing Systems 34: Annual Conference on Neural Information Processing Systems 2021, NeurIPS 2021}\/} (pp. \bibinfo{pages}{18932--18943}).
\bibitem[{Gower(1971)}]{gower1971general}
\bibinfo{author}{Gower, J.~C.} (\bibinfo{year}{1971}).
\newblock \bibinfo{title}{A general coefficient of similarity and some of its properties}.
\newblock {\it \bibinfo{journal}{Biometrics}\/},  (pp. \bibinfo{pages}{857--871}). \DOIprefix\doi{10.2307/2528823}.
\bibitem[{Gressel et~al.(2021)Gressel, Hegde, Sreekumar, Radhakrishnan, Harikumar, Achuthan et~al.}]{gressel2021feature}
\bibinfo{author}{Gressel, G.}, \bibinfo{author}{Hegde, N.}, \bibinfo{author}{Sreekumar, A.}, \bibinfo{author}{Radhakrishnan, R.}, \bibinfo{author}{Harikumar, K.}, \bibinfo{author}{Achuthan, K.} et~al. (\bibinfo{year}{2021}).
\newblock \bibinfo{title}{Feature importance guided attack: {A} model agnostic adversarial attack}.
\newblock \href{http://arxiv.org/abs/2106.14815}{\tt arXiv:2106.14815}.
\bibitem[{Gunasekaran et~al.(2023)}]{gunasekaran2023generating}
\bibinfo{author}{Gunasekaran, M.} et~al. (\bibinfo{year}{2023}).
\newblock \bibinfo{title}{Generating and defending against adversarial examples for loan eligibility prediction}.
\newblock In {\it \bibinfo{booktitle}{2023 International Conference on System, Computation, Automation and Networking (ICSCAN)}\/} (pp. \bibinfo{pages}{1--6}).
\newblock \bibinfo{organization}{IEEE}.
\bibitem[{He et~al.(2025)He, Ouyang, Alzubaidi, Barros \& Moreira}]{he2025investigating}
\bibinfo{author}{He, Z.}, \bibinfo{author}{Ouyang, C.}, \bibinfo{author}{Alzubaidi, L.}, \bibinfo{author}{Barros, A.}, \& \bibinfo{author}{Moreira, C.} (\bibinfo{year}{2025}).
\newblock \bibinfo{title}{Investigating imperceptibility of adversarial attacks on tabular data: An empirical analysis}.
\newblock {\it \bibinfo{journal}{Intelligent Systems with Applications}\/},  {\it \bibinfo{volume}{25}\/}, \bibinfo{pages}{200461}. \DOIprefix\doi{10.1016/J.ISWA.2024.200461}.
\bibitem[{Hingun et~al.(2023)Hingun, Sitawarin, Li \& Wagner}]{hingun2023reap}
\bibinfo{author}{Hingun, N.}, \bibinfo{author}{Sitawarin, C.}, \bibinfo{author}{Li, J.}, \& \bibinfo{author}{Wagner, D.} (\bibinfo{year}{2023}).
\newblock \bibinfo{title}{{REAP:} {A} large-scale realistic adversarial patch benchmark}.
\newblock In {\it \bibinfo{booktitle}{{IEEE/CVF} International Conference on Computer Vision, {ICCV} 2023}\/} (pp. \bibinfo{pages}{4640--4651}).
\newblock \bibinfo{publisher}{{IEEE}}.
\newblock \DOIprefix\doi{10.1109/ICCV51070.2023.00428}.
\bibitem[{Huang et~al.(2020)Huang, Khetan, Cvitkovic \& Karnin}]{huang2020tabtransformer}
\bibinfo{author}{Huang, X.}, \bibinfo{author}{Khetan, A.}, \bibinfo{author}{Cvitkovic, M.}, \& \bibinfo{author}{Karnin, Z.} (\bibinfo{year}{2020}).
\newblock \bibinfo{title}{Tabtransformer: Tabular data modeling using contextual embeddings}.
\newblock \href{http://arxiv.org/abs/2012.06678}{\tt arXiv:2012.06678}.
\bibitem[{Jin et~al.(2024)Jin, Zhang, Zhu \& Chen}]{jin2024benchmarking}
\bibinfo{author}{Jin, Z.}, \bibinfo{author}{Zhang, J.}, \bibinfo{author}{Zhu, Z.}, \& \bibinfo{author}{Chen, H.} (\bibinfo{year}{2024}).
\newblock \bibinfo{title}{Benchmarking transferable adversarial attacks}.
\newblock In {\it \bibinfo{booktitle}{Workshop on AI Systems with Confidential Computing (AISCC) 2024}\/}.
\newblock \DOIprefix\doi{10.14722/aiscc.2024.23017}.
\bibitem[{Khosravi et~al.(2023)Khosravi, Weston, Nugen, Mickley, Kremers, Wyles, Carter \& Taunton}]{khosravi2023demystifying}
\bibinfo{author}{Khosravi, B.}, \bibinfo{author}{Weston, A.~D.}, \bibinfo{author}{Nugen, F.}, \bibinfo{author}{Mickley, J.~P.}, \bibinfo{author}{Kremers, H.~M.}, \bibinfo{author}{Wyles, C.~C.}, \bibinfo{author}{Carter, R.~E.}, \& \bibinfo{author}{Taunton, M.~J.} (\bibinfo{year}{2023}).
\newblock \bibinfo{title}{Demystifying statistics and machine learning in analysis of structured tabular data}.
\newblock {\it \bibinfo{journal}{The Journal of arthroplasty}\/},  {\it \bibinfo{volume}{38}\/}, \bibinfo{pages}{1943--1947}.
\bibitem[{Kireev et~al.(2023)Kireev, Kulynych \& Troncoso}]{kireev2022adversarial}
\bibinfo{author}{Kireev, K.}, \bibinfo{author}{Kulynych, B.}, \& \bibinfo{author}{Troncoso, C.} (\bibinfo{year}{2023}).
\newblock \bibinfo{title}{Adversarial robustness for tabular data through cost and utility awareness}.
\newblock In {\it \bibinfo{booktitle}{30th Annual Network and Distributed System Security Symposium, {NDSS} 2023}\/}.
\newblock \DOIprefix\doi{10.14722/ndss.2023.24924}.
\bibitem[{Ko et~al.(2023)Ko, Kim \& Kwon}]{ko2023multi}
\bibinfo{author}{Ko, K.}, \bibinfo{author}{Kim, S.}, \& \bibinfo{author}{Kwon, H.} (\bibinfo{year}{2023}).
\newblock \bibinfo{title}{Multi-targeted audio adversarial example for use against speech recognition systems}.
\newblock {\it \bibinfo{journal}{Computers \& Security}\/},  {\it \bibinfo{volume}{128}\/}, \bibinfo{pages}{103168}.
\bibitem[{Kurakin et~al.(2017)Kurakin, Goodfellow \& Bengio}]{Kurakin2017adversarial}
\bibinfo{author}{Kurakin, A.}, \bibinfo{author}{Goodfellow, I.~J.}, \& \bibinfo{author}{Bengio, S.} (\bibinfo{year}{2017}).
\newblock \bibinfo{title}{Adversarial examples in the physical world}.
\newblock In {\it \bibinfo{booktitle}{5th International Conference on Learning Representations, {ICLR} 2017}\/}.
\bibitem[{Le \& Ho(2005)}]{le2005association}
\bibinfo{author}{Le, S.~Q.}, \& \bibinfo{author}{Ho, T.~B.} (\bibinfo{year}{2005}).
\newblock \bibinfo{title}{An association-based dissimilarity measure for categorical data}.
\newblock {\it \bibinfo{journal}{Pattern Recognition Letters}\/},  {\it \bibinfo{volume}{26}\/}, \bibinfo{pages}{2549--2557}. \DOIprefix\doi{10.1016/J.PATREC.2005.06.002}.
\bibitem[{Li et~al.(2021)Li, Lei, Gan \& Liu}]{li2021adversarial}
\bibinfo{author}{Li, L.}, \bibinfo{author}{Lei, J.}, \bibinfo{author}{Gan, Z.}, \& \bibinfo{author}{Liu, J.} (\bibinfo{year}{2021}).
\newblock \bibinfo{title}{Adversarial {VQA:} {A} new benchmark for evaluating the robustness of {VQA} models}.
\newblock In {\it \bibinfo{booktitle}{2021 {IEEE/CVF} International Conference on Computer Vision, {ICCV} 2021}\/} (pp. \bibinfo{pages}{2042--2051}).
\newblock \bibinfo{publisher}{{IEEE}}.
\newblock \DOIprefix\doi{10.1109/ICCV48922.2021.00205}.
\bibitem[{Lunghi et~al.(2023)Lunghi, Simitsis, Caelen \& Bontempi}]{lunghi2023adversarial}
\bibinfo{author}{Lunghi, D.}, \bibinfo{author}{Simitsis, A.}, \bibinfo{author}{Caelen, O.}, \& \bibinfo{author}{Bontempi, G.} (\bibinfo{year}{2023}).
\newblock \bibinfo{title}{Adversarial learning in real-world fraud detection: Challenges and perspectives}.
\newblock In {\it \bibinfo{booktitle}{Proceedings of the Second ACM Data Economy Workshop}\/} (pp. \bibinfo{pages}{27--33}).
\bibitem[{Madry et~al.(2018)Madry, Makelov, Schmidt, Tsipras \& Vladu}]{madry2017towards}
\bibinfo{author}{Madry, A.}, \bibinfo{author}{Makelov, A.}, \bibinfo{author}{Schmidt, L.}, \bibinfo{author}{Tsipras, D.}, \& \bibinfo{author}{Vladu, A.} (\bibinfo{year}{2018}).
\newblock \bibinfo{title}{Towards deep learning models resistant to adversarial attacks}.
\newblock In {\it \bibinfo{booktitle}{6th International Conference on Learning Representations, {ICLR} 2018}\/}.
\bibitem[{Mathov et~al.(2022)Mathov, Levy, Katzir, Shabtai \& Elovici}]{Mathov2022not}
\bibinfo{author}{Mathov, Y.}, \bibinfo{author}{Levy, E.}, \bibinfo{author}{Katzir, Z.}, \bibinfo{author}{Shabtai, A.}, \& \bibinfo{author}{Elovici, Y.} (\bibinfo{year}{2022}).
\newblock \bibinfo{title}{Not all datasets are born equal: On heterogeneous tabular data and adversarial examples}.
\newblock {\it \bibinfo{journal}{Knowledge-based Systems}\/},  {\it \bibinfo{volume}{242}\/}, \bibinfo{pages}{108377}. \DOIprefix\doi{10.1016/J.KNOSYS.2022.108377}.
\bibitem[{Moosavi{-}Dezfooli et~al.(2016)Moosavi{-}Dezfooli, Fawzi \& Frossard}]{Moosavi2016deepfool}
\bibinfo{author}{Moosavi{-}Dezfooli, S.}, \bibinfo{author}{Fawzi, A.}, \& \bibinfo{author}{Frossard, P.} (\bibinfo{year}{2016}).
\newblock \bibinfo{title}{Deepfool: {A} simple and accurate method to fool deep neural networks}.
\newblock In {\it \bibinfo{booktitle}{2016 {IEEE} Conference on Computer Vision and Pattern Recognition, {CVPR} 2016}\/} (pp. \bibinfo{pages}{2574--2582}).
\newblock \bibinfo{publisher}{{IEEE} Computer Society}.
\newblock \DOIprefix\doi{10.1109/CVPR.2016.282}.
\bibitem[{Mushava \& Murray(2024)}]{mushava2024flexible}
\bibinfo{author}{Mushava, J.}, \& \bibinfo{author}{Murray, M.} (\bibinfo{year}{2024}).
\newblock \bibinfo{title}{Flexible loss functions for binary classification in gradient-boosted decision trees: An application to credit scoring}.
\newblock {\it \bibinfo{journal}{Expert Systems with Applications}\/},  {\it \bibinfo{volume}{238}\/}, \bibinfo{pages}{121876}.
\bibitem[{Nandy et~al.(2023)Nandy, Chauhan, Saket \& Raghuveer}]{nandy2023non}
\bibinfo{author}{Nandy, J.}, \bibinfo{author}{Chauhan, J.}, \bibinfo{author}{Saket, R.}, \& \bibinfo{author}{Raghuveer, A.} (\bibinfo{year}{2023}).
\newblock \bibinfo{title}{Non-uniform adversarial perturbations for discrete tabular datasets}.
\newblock In {\it \bibinfo{booktitle}{Proceedings of the 32nd ACM International Conference on Information and Knowledge Management}\/} (pp. \bibinfo{pages}{1887--1896}).
\newblock \DOIprefix\doi{10.1145/3583780.3614992}.
\bibitem[{Noureddine et~al.(2023)Noureddine, Kheddar \& Maazouz}]{noureddine2023adversarial}
\bibinfo{author}{Noureddine, K.}, \bibinfo{author}{Kheddar, H.}, \& \bibinfo{author}{Maazouz, M.} (\bibinfo{year}{2023}).
\newblock \bibinfo{title}{Adversarial example detection techniques in speech recognition systems: A review}.
\newblock In {\it \bibinfo{booktitle}{2023 2nd International Conference on Electronics, Energy and Measurement (IC2EM)}\/} (pp. \bibinfo{pages}{1--7}).
\newblock \bibinfo{organization}{IEEE} volume~\bibinfo{volume}{1}.
\bibitem[{Papernot et~al.(2016)Papernot, McDaniel, Jha, Fredrikson, Celik \& Swami}]{papernot2016limitation}
\bibinfo{author}{Papernot, N.}, \bibinfo{author}{McDaniel, P.~D.}, \bibinfo{author}{Jha, S.}, \bibinfo{author}{Fredrikson, M.}, \bibinfo{author}{Celik, Z.~B.}, \& \bibinfo{author}{Swami, A.} (\bibinfo{year}{2016}).
\newblock \bibinfo{title}{The limitations of deep learning in adversarial settings}.
\newblock In {\it \bibinfo{booktitle}{{IEEE} European Symposium on Security and Privacy, EuroS{\&}P 2016}\/} (pp. \bibinfo{pages}{372--387}).
\newblock \bibinfo{publisher}{{IEEE}}.
\newblock \DOIprefix\doi{10.1109/EUROSP.2016.36}.
\bibitem[{Pelekis et~al.(2025)Pelekis, Koutroubas, Blika, Berdelis, Karakolis, Ntanos, Spiliotis \& Askounis}]{pelekis2025adversarial}
\bibinfo{author}{Pelekis, S.}, \bibinfo{author}{Koutroubas, T.}, \bibinfo{author}{Blika, A.}, \bibinfo{author}{Berdelis, A.}, \bibinfo{author}{Karakolis, E.}, \bibinfo{author}{Ntanos, C.}, \bibinfo{author}{Spiliotis, E.}, \& \bibinfo{author}{Askounis, D.} (\bibinfo{year}{2025}).
\newblock \bibinfo{title}{Adversarial machine learning: a review of methods, tools, and critical industry sectors.}
\newblock {\it \bibinfo{journal}{Artificial Intelligence Review}\/},  {\it \bibinfo{volume}{58}\/}.
\bibitem[{Qin et~al.(2019)Qin, Carlini, Cottrell, Goodfellow \& Raffel}]{qin2019imperceptible}
\bibinfo{author}{Qin, Y.}, \bibinfo{author}{Carlini, N.}, \bibinfo{author}{Cottrell, G.~W.}, \bibinfo{author}{Goodfellow, I.~J.}, \& \bibinfo{author}{Raffel, C.} (\bibinfo{year}{2019}).
\newblock \bibinfo{title}{Imperceptible, robust, and targeted adversarial examples for automatic speech recognition}.
\newblock In {\it \bibinfo{booktitle}{Proceedings of the 36th International Conference on Machine Learning, {ICML} 2019}\/} (pp. \bibinfo{pages}{5231--5240}).
\newblock volume~\bibinfo{volume}{97}.
\bibitem[{Rocamora et~al.(2024)Rocamora, Wu, Liu, Chrysos \& Cevher}]{rocamora2024revisiting}
\bibinfo{author}{Rocamora, E.~A.}, \bibinfo{author}{Wu, Y.}, \bibinfo{author}{Liu, F.}, \bibinfo{author}{Chrysos, G.~G.}, \& \bibinfo{author}{Cevher, V.} (\bibinfo{year}{2024}).
\newblock \bibinfo{title}{Revisiting character-level adversarial attacks for language models}.
\newblock {\it \bibinfo{journal}{arXiv preprint arXiv:2405.04346}\/}, .
\bibitem[{Sadeghi et~al.(2020)Sadeghi, Banerjee \& Gupta}]{sadeghi2020system}
\bibinfo{author}{Sadeghi, K.}, \bibinfo{author}{Banerjee, A.}, \& \bibinfo{author}{Gupta, S.~K.} (\bibinfo{year}{2020}).
\newblock \bibinfo{title}{A system-driven taxonomy of attacks and defenses in adversarial machine learning}.
\newblock {\it \bibinfo{journal}{IEEE transactions on emerging topics in computational intelligence}\/},  {\it \bibinfo{volume}{4}\/}, \bibinfo{pages}{450--467}.
\bibitem[{Shokri et~al.(2017)Shokri, Stronati, Song \& Shmatikov}]{shokri2017membership}
\bibinfo{author}{Shokri, R.}, \bibinfo{author}{Stronati, M.}, \bibinfo{author}{Song, C.}, \& \bibinfo{author}{Shmatikov, V.} (\bibinfo{year}{2017}).
\newblock \bibinfo{title}{Membership inference attacks against machine learning models}.
\newblock In {\it \bibinfo{booktitle}{2017 IEEE symposium on security and privacy (SP)}\/} (pp. \bibinfo{pages}{3--18}).
\newblock \bibinfo{organization}{IEEE}.
\bibitem[{Siddiqui et~al.(2020)Siddiqui, Dengel \& Ahmed}]{siddiqui2020benchmarking}
\bibinfo{author}{Siddiqui, S.~A.}, \bibinfo{author}{Dengel, A.}, \& \bibinfo{author}{Ahmed, S.} (\bibinfo{year}{2020}).
\newblock \bibinfo{title}{Benchmarking adversarial attacks and defenses for time-series data}.
\newblock In {\it \bibinfo{booktitle}{Neural Information Processing - 27th International Conference, {ICONIP} 2020}\/} (pp. \bibinfo{pages}{544--554}).
\newblock \bibinfo{organization}{Springer}.
\newblock \DOIprefix\doi{10.1007/978-3-030-63836-8_45}.
\bibitem[{Simonetto et~al.(2022)Simonetto, Dyrmishi, Ghamizi, Cordy \& Traon}]{simonetto2021unified}
\bibinfo{author}{Simonetto, T.}, \bibinfo{author}{Dyrmishi, S.}, \bibinfo{author}{Ghamizi, S.}, \bibinfo{author}{Cordy, M.}, \& \bibinfo{author}{Traon, Y.~L.} (\bibinfo{year}{2022}).
\newblock \bibinfo{title}{A unified framework for adversarial attack and defense in constrained feature space}.
\newblock In {\it \bibinfo{booktitle}{Proceedings of the Thirty-First International Joint Conference on Artificial Intelligence, {IJCAI} 2022}\/}.
\newblock \DOIprefix\doi{10.24963/IJCAI.2022/183}.
\bibitem[{Simonetto et~al.(2024{\natexlab{a}})Simonetto, Ghamizi \& Cordy}]{simonetto2024constrained}
\bibinfo{author}{Simonetto, T.}, \bibinfo{author}{Ghamizi, S.}, \& \bibinfo{author}{Cordy, M.} (\bibinfo{year}{2024}{\natexlab{a}}).
\newblock \bibinfo{title}{Constrained adaptive attack: Effective adversarial attack against deep neural networks for tabular data}.
\newblock {\it \bibinfo{journal}{Advances in Neural Information Processing Systems}\/},  {\it \bibinfo{volume}{37}\/}, \bibinfo{pages}{27817--27849}.
\bibitem[{Simonetto et~al.(2024{\natexlab{b}})Simonetto, Ghamizi \& Cordy}]{simonetto2024tabularbench}
\bibinfo{author}{Simonetto, T.}, \bibinfo{author}{Ghamizi, S.}, \& \bibinfo{author}{Cordy, M.} (\bibinfo{year}{2024}{\natexlab{b}}).
\newblock \bibinfo{title}{Tabularbench: Benchmarking adversarial robustness for tabular deep learning in real-world use-cases}.
\newblock {\it \bibinfo{journal}{Advances in Neural Information Processing Systems}\/},  {\it \bibinfo{volume}{37}\/}, \bibinfo{pages}{78394--78430}.
\bibitem[{Szegedy et~al.(2014)Szegedy, Zaremba, Sutskever, Bruna, Erhan, Goodfellow \& Fergus}]{szegedy2014intriguing}
\bibinfo{author}{Szegedy, C.}, \bibinfo{author}{Zaremba, W.}, \bibinfo{author}{Sutskever, I.}, \bibinfo{author}{Bruna, J.}, \bibinfo{author}{Erhan, D.}, \bibinfo{author}{Goodfellow, I.~J.}, \& \bibinfo{author}{Fergus, R.} (\bibinfo{year}{2014}).
\newblock \bibinfo{title}{Intriguing properties of neural networks}.
\newblock In {\it \bibinfo{booktitle}{2nd International Conference on Learning Representations, {ICLR} 2014}\/}.
\bibitem[{Thiyagalingam et~al.(2022)Thiyagalingam, Shankar, Fox \& Hey}]{thiyagalingam2022scientific}
\bibinfo{author}{Thiyagalingam, J.}, \bibinfo{author}{Shankar, M.}, \bibinfo{author}{Fox, G.}, \& \bibinfo{author}{Hey, T.} (\bibinfo{year}{2022}).
\newblock \bibinfo{title}{Scientific machine learning benchmarks}.
\newblock {\it \bibinfo{journal}{Nature Reviews Physics}\/},  {\it \bibinfo{volume}{4}\/}, \bibinfo{pages}{413--420}. \DOIprefix\doi{10.1038/s42254-022-00441-7}.
\bibitem[{Wang et~al.(2025)Wang, Li \& He}]{wang2025unified}
\bibinfo{author}{Wang, J.}, \bibinfo{author}{Li, F.}, \& \bibinfo{author}{He, L.} (\bibinfo{year}{2025}).
\newblock \bibinfo{title}{A unified framework for adversarial patch attacks against visual 3d object detection in autonomous driving}.
\newblock {\it \bibinfo{journal}{IEEE Transactions on Circuits and Systems for Video Technology}\/}, .
\bibitem[{Weng et~al.(2024)Weng, Luo, Lin \& Li}]{weng2024comparative}
\bibinfo{author}{Weng, J.}, \bibinfo{author}{Luo, Z.}, \bibinfo{author}{Lin, D.}, \& \bibinfo{author}{Li, S.} (\bibinfo{year}{2024}).
\newblock \bibinfo{title}{Comparative evaluation of recent universal adversarial perturbations in image classification}.
\newblock {\it \bibinfo{journal}{Computers \& Security}\/},  {\it \bibinfo{volume}{136}\/}, \bibinfo{pages}{103576}.
\bibitem[{Yang et~al.(2023)Yang, Huang, Cao, Ma, Zhang \& Li}]{yang2023quantifying}
\bibinfo{author}{Yang, Y.}, \bibinfo{author}{Huang, P.}, \bibinfo{author}{Cao, J.}, \bibinfo{author}{Ma, F.}, \bibinfo{author}{Zhang, J.}, \& \bibinfo{author}{Li, J.} (\bibinfo{year}{2023}).
\newblock \bibinfo{title}{Quantifying robustness to adversarial word substitutions}.
\newblock In {\it \bibinfo{booktitle}{Joint European Conference on Machine Learning and Knowledge Discovery in Databases}\/} (pp. \bibinfo{pages}{95--112}).
\newblock \bibinfo{organization}{Springer}.
\bibitem[{Yi et~al.(2023)Yi, Cao, Pu, Wu, Chen, Khan, Francis \& Li}]{yi2023fraud}
\bibinfo{author}{Yi, Z.}, \bibinfo{author}{Cao, X.}, \bibinfo{author}{Pu, X.}, \bibinfo{author}{Wu, Y.}, \bibinfo{author}{Chen, Z.}, \bibinfo{author}{Khan, A.~T.}, \bibinfo{author}{Francis, A.}, \& \bibinfo{author}{Li, S.} (\bibinfo{year}{2023}).
\newblock \bibinfo{title}{Fraud detection in capital markets: A novel machine learning approach}.
\newblock {\it \bibinfo{journal}{Expert Systems with Applications}\/},  {\it \bibinfo{volume}{231}\/}, \bibinfo{pages}{120760}.
\bibitem[{Zhang et~al.(2020)Zhang, Sheng, Alhazmi \& Li}]{zhang2020adversarial}
\bibinfo{author}{Zhang, W.~E.}, \bibinfo{author}{Sheng, Q.~Z.}, \bibinfo{author}{Alhazmi, A.}, \& \bibinfo{author}{Li, C.} (\bibinfo{year}{2020}).
\newblock \bibinfo{title}{Adversarial attacks on deep-learning models in natural language processing: {A} survey}.
\newblock {\it \bibinfo{journal}{ACM Transactions on Intelligent Systems and Technology}\/},  {\it \bibinfo{volume}{11}\/}, \bibinfo{pages}{1--41}. \DOIprefix\doi{10.1145/3374217}.
\bibitem[{Zheng et~al.(2023)Zheng, Yan, Zhu, Chen \& Wu}]{zheng2023blackboxbench}
\bibinfo{author}{Zheng, M.}, \bibinfo{author}{Yan, X.}, \bibinfo{author}{Zhu, Z.}, \bibinfo{author}{Chen, H.}, \& \bibinfo{author}{Wu, B.} (\bibinfo{year}{2023}).
\newblock \bibinfo{title}{Blackboxbench: A comprehensive benchmark of black-box adversarial attacks}.
\newblock \href{http://arxiv.org/abs/2312.16979}{\tt arXiv:2312.16979}.
\bibitem[{Zheng et~al.(2021)Zheng, Zou, Dong, Cen, Yin, Xu, Yang \& Tang}]{zheng2021graph}
\bibinfo{author}{Zheng, Q.}, \bibinfo{author}{Zou, X.}, \bibinfo{author}{Dong, Y.}, \bibinfo{author}{Cen, Y.}, \bibinfo{author}{Yin, D.}, \bibinfo{author}{Xu, J.}, \bibinfo{author}{Yang, Y.}, \& \bibinfo{author}{Tang, J.} (\bibinfo{year}{2021}).
\newblock \bibinfo{title}{Graph robustness benchmark: Benchmarking the adversarial robustness of graph machine learning}.
\newblock In {\it \bibinfo{booktitle}{Proceedings of the Neural Information Processing Systems Track on Datasets and Benchmarks 1, NeurIPS Datasets and Benchmarks 2021}\/}.
\bibitem[{Zhou et~al.(2022)Zhou, Zaidi, Zhang \& Li}]{zhou2022Discretization}
\bibinfo{author}{Zhou, J.}, \bibinfo{author}{Zaidi, N.~A.}, \bibinfo{author}{Zhang, Y.}, \& \bibinfo{author}{Li, G.} (\bibinfo{year}{2022}).
\newblock \bibinfo{title}{Discretization inspired defence algorithm against adversarial attacks on tabular data}.
\newblock In {\it \bibinfo{booktitle}{Advances in Knowledge Discovery and Data Mining - 26th Pacific-Asia Conference, {PAKDD} 2022}\/} (pp. \bibinfo{pages}{367--379}).
\newblock volume \bibinfo{volume}{13281} of {\it \bibinfo{series}{Lecture Notes in Computer Science}\/}.
\newblock \DOIprefix\doi{10.1007/978-3-031-05936-0_29}.

\end{thebibliography}



\end{document}